%% file: neuro_template.tex
\documentclass[preprint,12pt]{elsarticle}
\pdfoutput=1 

\usepackage{amssymb}
\usepackage{graphicx}
\usepackage{amsmath}
\usepackage{amsfonts}
\usepackage{color}
\usepackage[english]{babel}
\usepackage{microtype}
\usepackage{booktabs}

\journal{Neurocomputing}

\newcommand{\sign}[1]{\mathrm{sign}\left({#1}\right)}
\newcommand{\givenbeta}[1]{\left\{{#1}\right\}}
\usepackage{algorithmic}
\usepackage{algorithm}
\usepackage{url}

\begin{document}

\begin{frontmatter}

\title{Mixture of von Mises-Fisher distribution
with sparse prototypes}

\author[inst2]{Fabrice Rossi}

\affiliation[inst2]{organization={Universite Paris Dauphine-PSL - CEREMADE UMR 7534},
            city={Paris},
            country={France}}

\author[inst1]{Florian Barbaro}

\affiliation[inst1]{organization={Universite Paris 1 Pantheon-Sorbonne - Laboratoire SAMM EA 4543},
            city={Paris},
            country={France}}

\begin{abstract}
Mixtures of von Mises-Fisher distributions can be used to cluster data on the
unit hypersphere. This is particularly adapted for high-dimensional
directional data such as texts. We propose in this article to estimate a von
Mises mixture using a $l_1$ penalized likelihood. This leads to sparse
prototypes that improve clustering interpretability. We introduce an expectation-maximisation (EM) algorithm for this estimation and explore the trade-off between the sparsity term and the likelihood one with a path following algorithm. 
The model's behaviour is studied on simulated data and, we show the advantages of the approach on real data benchmark. 
We also introduce a new data set on financial reports and exhibit the benefits of our method for exploratory analysis. 
\end{abstract}


\end{frontmatter}

\section{Introduction}
\label{intro}
High dimensional data are difficult to study as many classical machine
learning techniques are impaired by the so called \emph{curse of
  dimensionality} \cite{Bellman+2015, clarke2008properties}. One of the manifestation of this curse
is the tendency of distances to concentrate: pairwise distances between
observations have both a large mean and a small variance (see
 \cite{10.1007/3-540-49257-7_15, 4216305}). This shows also that a
 multivariate Gaussian distribution is mostly concentrated on a central
 sphere. 

 As a consequence, the classical Gaussian mixture model is generally not
 adapted to high-dimen\-sional data and numerous variants have been proposed
 to cluster such data, see
 e.g. \cite{bouveyron2007high,JMLR:v8:pan07a,zhao19:_regul_gauss_mixtur_model_high_dimen_clust}
 and in particular the survey \cite{BOUVEYRON201452}.  One of the main
 strategy to adapt Gaussian mixtures to high dimensional settings is to reduce
 in some way the relevant dimensions of the components of the mixture. For
 instance in \cite{bouveyron2007high}, the authors propose a method in which
 each component of the Gaussian mixture is associated to a specific
 low-dimensional projection. In this sense, it can be seen as a generalization
 of the principle of principal component analysis mixture
 \cite{TippingBishop1999MixPCA}.

This strategy can be applied in a more direct way for a particular case of
high-dimensional data, the so-called \emph{directional data}
\cite{mardia2009directional} for which the correlation between two vectors is
more informative than the norm of their difference (i.e. the Euclidean
distance). This type of data appears naturally in the classical vector
representation of texts, as well as microarray analysis and recommender
systems. In addition of the need for a specific similarity measure, those data
have frequently more variables than the number of observations. This
constrains strongly the type of Gaussian mixture than can be considered as
e.g. the covariance matrix of the data is degenerate. For those data
Gaussian-type models are doubly non adapted: they suffer from the adverse
effects of high dimensionality and are based on a non adapted underlying
metric. 

A natural way to handle directional data is to carry out a normalisation that
places them on the unit hyper-sphere. Notice that the concentration phenomenon
recalled above has already a tendency to push all observations on such a
hyper-sphere. This gives to the directional model a broader application domain
in high dimensional spaces. Then one can use clustering techniques
that address specifically the fact the data are spherical, such as spherical
k-means \cite{Dhillon2001}. In particular, the von Mises-Fisher distribution
can be used as the building block for mixture models for directional data. 

The von Mises-Fisher (vMF) distribution is a probability distribution on the
unit hypersphere which is close to the wrapped version of the normal
distribution but is also simpler and more tractable. It uses two parameters: a
directional mean and a concentration parameter $\kappa$ which play similar
roles as the mean and the precision (inverse of the variance) in the Gaussian
distribution. Its density is given by
\begin{equation*}
f(\boldsymbol{x}| \boldsymbol{\mu},\kappa)=c_d(\kappa)\exp^{\kappa\boldsymbol{\mu}^T\boldsymbol{x}},
\end{equation*}
where $c_d(\kappa)$ is the normalizing constant. Interestingly the inner
product $\boldsymbol{\mu}^T\boldsymbol{x}$ can be seen as a form of projection
to a one dimensional subspace, emphasizing the link between this approach and
the ones developed to adapt Gaussian mixtures to high dimensional data.

Early application of the
von Mises-Fisher (vMF) distribution were limited to low-dimensional data due
to the difficulty of estimating the $\kappa$ concentration parameter which
involves inverting ratios of Bessel functions (see
e.g. \cite{mclachlan2004finite}). However Banerjee et al. introduced in
\cite{10.5555/1046920.1088718} a new estimation technique for the
concentration parameter and showed that it was adapted for high dimensional
spherical data. It was shown in \cite{10.5555/1046920.1088718,10.5555/1086342.1086348} that
mixtures of vMF distribution are particularly adapted for directional data
clustering. This early work has led to the development of numerous applications
of vMF distribution such as the spherical topic model
\cite{reisinger:icml10}, inspired by Latent Dirichlet Allocation, and Bayesian
variations of spherical mixture models in \cite{pmlr-v32-gopal14}. 

In order to improve further mixture of vMF distributions,
\cite{pmlr-v51-salah16} introduced structure and sparsity in the directional means. The
approach is inspired by co-clustering and enforces a diagonal structure on the
matrix of directional means (after a proper reordering). In the case of text data
analysis, this amounts to finding clusters of texts that are characterized by
a specific vocabulary. An improvement of the algorithm was introduced in
\cite{salah2017model}: a \emph{conscience mechanism} prevents the method from
generating highly skewed cluster size distributions. 

In the present article, we aim also at producing sparse directional means but
we follow a different strategy. In particular, we consider that the
co-clustering structure is too strict in some applications where some of the
clusters should be able to share vocabulary (using again text clustering as
the typical application of directional data clustering). Following
\cite{JMLR:v8:pan07a}, we propose to use a $l_1$ penalty for a mixture of von
Mises-Fisher distributions to enforce the sparsity in the directional means
and thus improve the understanding of classification results for
high-dimensional data. Our solution is based on a modification of the
Expectation-Maximisation (EM) algorithm \cite{DempsterEtAlEM1977} proposed by
\cite{10.5555/1046920.1088718}. Moreover, we propose an efficient methodology
for tuning the penalty parameter that handles the trade off between the
likelihood and the sparsity of the solution. It combines a path following
strategy with the use of model selection criteria to select such a trade off. As in
\cite{pmlr-v51-salah16,salah2017model}, reordering the columns of the matrix
of directional means, enables us to display those means in an organized
fashion, emphasizing common aspects (e.g. vocabulary) and exclusive ones.

The rest of the paper is organized as follows. In Section \ref{sec:MvMF} we
recall the mixture of von Mises-Fisher distributions model from
\cite{10.5555/1046920.1088718}. In Section \ref{sec:sparseMvMF} we describe
our $l_1$ regularized variant together with the modified EM algorithm and the
path following strategy adapted for selecting the regularization trade-off. In
Section \ref{sec:exp} we analyze the behavior of the proposed model in
details, using artificial data. Section \ref{sec:exper-real-world} is
dedicated to a comparison of the proposed model with reference models on both
simulated data and a real world benchmark. Finally, Section \ref{sec:exper-real-world:WFC}
proposes an application of our model on a recent text database about 8-K
reports. 

\section{Mixture of von Mises-Fisher distribution}\label{sec:MvMF}
We present briefly in this section the mixture of von Mises-Fisher
distributions model from \cite{10.5555/1046920.1088718}. This generative model
provides a distribution on $\mathbb{S}^{d-1}$, the $(d-1)$ dimensional unit
sphere embedded in  $\mathbb{R}^{d}$, that is
\begin{equation*}
\mathbb{S}^{d-1}=\left\{\boldsymbol{x}\in \mathbb{R}^{d} | \left\lVert \boldsymbol{x}\right\rVert_2=1\right\},
\end{equation*}
where $\left\lVert.\right\rVert_2$ denotes the $l_2$ (Euclidean) norm in
$\mathbb{R}^{d}$. 

\subsection{The von Mises-Fisher (vMF) distribution}
The von Mises-Fisher distribution is defined on $\mathbb{S}^{d-1}$
($d\geq 2$) by the following probability density function
\begin{equation}
f(\boldsymbol{x}| \boldsymbol{\mu},\kappa)=c_d(\kappa)\exp^{\kappa\boldsymbol{\mu}^T\boldsymbol{x}},\label{vmf}
\end{equation}
where $\boldsymbol{\mu}\in \mathbb{S}^{d-1}$ is the directional mean of the
distribution and $\kappa\geq 0$ its concentration parameter. The normalization
term $c_d(\kappa)$ is given by
\begin{equation}
 c_d(\kappa)=\frac{\kappa^{d/2-1}}{(2\pi)^{d/2}I_{d/2-1}(\kappa)},\label{constvmf}
\end{equation}
where $I_{r}$ denotes the modified Bessel function of the first kind and order
$r$.

\subsection{Maximum likelihood estimates}\label{sec:maxim-likel-estim}
As shown in e.g. \cite{10.5555/1046920.1088718}, the maximum likelihood
estimates (MLE) of the directional mean of a vMF from a sample of $N$ independent
identically distributed observations $\boldsymbol{X}=(\boldsymbol{x}_i)_{1\leq
  i\leq N}$ is straightforward as we have
\begin{equation}
  \label{eq:mu:MLE}
\widehat{\boldsymbol{\mu}}=\frac{\sum_{i=1}^n\boldsymbol{x}_i}{\left\lVert \sum_{i=1}^n\boldsymbol{x}_i\right\rVert_2}.
\end{equation}
However, the estimation of $\kappa$ is only indirect. One can show indeed that
$\widehat{\kappa}$ is the solution of the following equation
\begin{equation}
  \label{eq:kappa:equation}
\frac{I_{d/2}(\widehat{\kappa})}{I_{d/2-1}(\widehat{\kappa})}=\frac{1}{n}\left\lVert \sum_{i=1}^n\boldsymbol{x}_i\right\rVert_2,
\end{equation}
which has no closed form solution. We follow the strategy of
\cite{10.5555/1046920.1088718} which estimates $\kappa$ via
\begin{equation}
  \label{eq:kappa:estimation}
\widetilde{\kappa}=\frac{\bar{r}d-\bar{r}^3}{1-\bar{r}^2},
\end{equation}
with
\begin{equation}
  \label{eq:bar:r}
\bar{r}=\frac{1}{n}\left\lVert \sum_{i=1}^n\boldsymbol{x}_i\right\rVert_2.
\end{equation}
Notice that we use this approach as it provides a good trade-off between
complexity and accuracy, but more advanced numerical schemes can be used, see
for instance \cite{hornik2014movmf} for a discussion about them. 

\subsection{Mixture of vMF}
To model multimodal distributions on the sphere, we use a mixture of $K$ vMF
distributions whose probability density function is given by
\begin{equation}
f(\boldsymbol{x}| \boldsymbol{\Theta})=\sum_{k=1}^K\alpha_k f_k(\boldsymbol{x}| \theta_k),\label{kvmf}
\end{equation}
where each $f_k$ is a vMF density function $\theta_k=(\boldsymbol{\mu}_k,
\kappa_k)$ and where $\boldsymbol{\Theta}$ gathers the $K$ directional means
$(\boldsymbol{\mu}_k)_{1\leq k\leq K}$, the $K$ concentration parameters
$(\kappa_k)_{1\leq k\leq K}$ and the mixture proportions $(\alpha_k)_{1\leq
  k\leq K}$ with $\alpha_k\geq 0$ and $\sum_{k=1}^k\alpha_k=1$.

The parameters $\boldsymbol{\Theta}$ can be estimated from a data set by
maximum likelihood using the EM algorithm, as show in
\cite{10.5555/1046920.1088718}. We derive a variation of the algorithm adapted
to our proposed regularized estimation in Section \ref{sec:sparseMvMF}. 

\section{Mixture of sparse vMF}\label{sec:sparseMvMF}
Following \cite{JMLR:v8:pan07a}, we propose to replace the standard maximum
likelihood estimate (MLE) of $\boldsymbol{\Theta}$ by a $l_1$ regularized
MLE. This induces sparsity in the directional means and, consequently, ease
the interpretation of the results. We derive the EM algorithm (Algorithm
\ref{code:EM}) and the path following strategy in the present section. We
also discuss information criteria for model selection. 

\subsection{A penalized likelihood for sparse directional means}
\subsubsection{Mixture representation}\label{sec:mixt-repr}
We use the classical represent of a mixture via latent variables. We assume
that the full data set consists of $N$ independent and identically
distributed pairs $(\boldsymbol{x}_i, z_i)_{1\leq i\leq N}=(\boldsymbol{X},\boldsymbol{Z})$. The $(z_i)_{i\leq
  i\leq N}$ are the latent unobserved variables while the
$(\boldsymbol{x}_i)_{1\leq i\leq N}$ are observed. Each $z_i$ follows a
categorical distribution over $\{1,\ldots, K\}$ with parameter
$\boldsymbol{\alpha}=(\alpha_k)_{1\leq k\leq K}$,
i.e. $\mathbb{P}(z_i=k|\boldsymbol{\alpha})=\alpha_k$.

Then the conditional density of $\boldsymbol{x}_i$ given $z_i=k$ is $f_k$, the
$k$-th component of the vMF mixture, i.e.
\begin{equation}
  \label{eq:x:given:z}
p(  \boldsymbol{x}_i|z_i=k,\boldsymbol{\Theta})=f_k(\boldsymbol{x}_i|\theta_k)=c_d(\kappa_k)\exp^{\kappa_k\boldsymbol{\mu}_k^T\boldsymbol{x}_i}.
\end{equation}
Obviously, this leads to the marginal distribution of $p(
\boldsymbol{x}_i|\boldsymbol{\Theta})$ given by equation \eqref{kvmf} and the
log-likelihood of the observed data is therefore
\begin{equation}
L(\boldsymbol{\Theta}|\boldsymbol{X})=\sum_{i=1}^n \ln\left(\sum_{k=1}^K\alpha_kf_k(\boldsymbol{x}_i|\theta_k)\right).\label{vraisemblance}
\end{equation}
To ease the derivation of the EM algorithm we introduce the classical one hot encoding
representation of the hidden variables: $z_i$ is represented by the binary
vector $\boldsymbol{z}_i$ with $\sum_{k=1}^Kz_{ik}=1$ and such that
$z_i=k\Leftrightarrow z_{ij}=0$ for $j\neq k$ and $z_{ik}=1$. Then the
log-likelihood of the complete data is given by
\begin{equation}
L(\boldsymbol{\Theta}|\boldsymbol{X},\boldsymbol{Z})=\sum_{i=1}^n\sum_{k=1}^K
z_{ik} \left(\ln \alpha_k + \ln f_k(\boldsymbol{x}_i|\theta_k)\right).\label{vraisemblancecompl} 
\end{equation}

\subsubsection{Penalized likelihood}
We propose to penalize the log-likelihood by the $l_1$ norms of the directional
means allowing thus to increase their sparsity. More precisely, we
estimate $\boldsymbol{\Theta}$ by maximizing the following penalized log-likelihood 
\begin{equation}
L_p(\boldsymbol{\Theta}|\boldsymbol{X})=L(\boldsymbol{\Theta}|\boldsymbol{X})-\beta 
\sum_{k=1}^K\left\lVert\boldsymbol{\mu_k}\right\rVert_1,
\label{lassologvraigvmf}
\end{equation}
where $\beta$  regulates the trade-off between likelihood and sparsity, and
where $\left\lVert.\right\rVert_1$ denotes the $l_1$ norm. As we will use the
complete log-likelihood in the EM algorithm, we introduce its penalized
version as follows
\begin{equation}
L_{p}(\boldsymbol{\Theta}|\boldsymbol{X},\boldsymbol{Z})=L(\boldsymbol{\Theta}|\boldsymbol{X},\boldsymbol{Z})-\beta 
\sum_{k=1}^K\left\lVert\boldsymbol{\mu_k}\right\rVert_1.\label{vraisemblancecomplpen}
\end{equation}

\subsection{EM algorithm}
We derive in this section the proposed EM algorithm. As proposed originally in
\cite{DempsterEtAlEM1977}, the Expectation-Maximization algorithm estimates
the parameters of a model from incomplete data by maximizing
the (penalized) log-likelihood via an alternating scheme (see the generic
Algorithm \ref{code:generic:EM}). In the Expectation
phase (E), one computes the expectation of the \emph{complete} log-likelihood
with respect to the latent unobserved variables. The distribution used for the
expectation is the posterior distribution of the latent variables given the
observed data and the current estimate of the parameters. In the Maximization
phase (M), the expectation computed in the E phase is maximized with respect
to the parameters, providing a new estimate. This two phase process is
repeated until convergence of the log-likelihood. 

\begin{algorithm}[htb]
  \caption{Generic EM algorithm}\label{code:generic:EM}
  \begin{algorithmic}
    \STATE{Initialise $\boldsymbol{\Theta^{(0)}}$ randomly}
    \STATE{$m\leftarrow 0$}
    \REPEAT
    \STATE{\emph{E phase}}
    \STATE{Compute $q^{(m)}(\boldsymbol{Z})=\mathbb{P}(\boldsymbol{Z}|\boldsymbol{X},\boldsymbol{\Theta^{(m)}})$    }    
    \STATE{Compute
      $Q(\boldsymbol{\Theta}|\boldsymbol{\Theta}^{(m)})=\mathbb{E}_{\mathcal{Z}\sim
        q^{(m)}}\left(
        L(\boldsymbol{\Theta}|\boldsymbol{X},\boldsymbol{Z})\right)$}
    \STATE{\emph{M phase}}
    \STATE{Compute $\boldsymbol{\Theta}^{(m+1)}=\arg\max_{\boldsymbol{\Theta}}Q(\boldsymbol{\Theta}|\boldsymbol{\Theta}^{(m)})$}
    \STATE{$m\leftarrow m+1$}
    \UNTIL{convergence of $L(\boldsymbol{\Theta}^{(m+1)}|\boldsymbol{X})$}
  \end{algorithmic}
\end{algorithm}

\subsubsection{E phase}
We follow both \cite{JMLR:v8:pan07a} and \cite{10.5555/1046920.1088718} to
derive the EM algorithm for our penalized estimator. In the expectation step
of the EM, we compute the expectation of $\ln
L_{p}(\boldsymbol{\Theta}|\boldsymbol{X},\boldsymbol{Z})$ with respect to a
distribution over the latent variables $\boldsymbol{Z}$. Obviously
\begin{equation}
  \label{eq:em:Ecalculation}
\mathbb{E}_{\boldsymbol{Z}\sim
  q}\left(L_{p}(\boldsymbol{\Theta}|\boldsymbol{X},\boldsymbol{Z})\right)=
\mathbb{E}_{\boldsymbol{Z}\sim q}\left(L(\boldsymbol{\Theta}|\boldsymbol{X},\boldsymbol{Z})\right)-\beta 
\sum_{k=1}^K\left\lVert\boldsymbol{\mu_k}\right\rVert_1,
\end{equation}
for any distribution $q$ as the penalty term does not depend on
$\boldsymbol{Z}$. Then the E phase for the penalized likelihood estimation
almost identical to the one derived in \cite{JMLR:v8:pan07a} without
penalization.

We need first to compute $q^{(m)}(\boldsymbol{Z})$. By independence of the
pairs $(\boldsymbol{x}_i, z_i)_{1\leq i\leq
  N}$, we have
\begin{equation}
q^{(m)}(\boldsymbol{Z})=\prod_{i=1}^N
\mathbb{P}(z_i|\boldsymbol{x}_i,\boldsymbol{\Theta^{(m)}}).
\end{equation}
Then, using assumptions from Section \ref{sec:mixt-repr}, we have
\begin{equation}
  \label{eq:pre:stepE_t}
\mathbb{P}(z_i=k|\boldsymbol{x}_i,\boldsymbol{\Theta^{(m)}})=\frac{\alpha_k^{(m)}f_k(\boldsymbol{x}_i,\theta_k^{(m)})}{\sum_{l=1}^K \alpha_l^{(m)} f_l(\boldsymbol{x}_i,\theta^{(m)}_l)}.
\end{equation}
Moreover, using the linearity of the expectation and equation \eqref{vraisemblancecompl}, we have
\begin{align}\notag
Q(\boldsymbol{\Theta}|\boldsymbol{\Theta}^{(m)})&=\mathbb{E}_{\mathcal{Z}\sim q^{(m)}}\left(                                              L(\boldsymbol{\Theta}|\boldsymbol{X},\boldsymbol{Z})\right),\\
   \notag
  &=\sum_{i=1}^n\sum_{k=1}^K\mathbb{E}_{\mathcal{Z}\sim q^{(m)}}(z_{ik}) \left(\ln \alpha_k
    + \ln f_k(\boldsymbol{x}_i|\theta_k)\right),\\\notag
 &=\sum_{i=1}^n\sum_{k=1}^K\mathbb{P}(z_i=k|\boldsymbol{x}_i,\boldsymbol{\Theta^{(m)}}) \left(\ln \alpha_k
   + \ln f_k(\boldsymbol{x}_i|\theta_k)\right),\\\label{eq:vraisestimation:nopen}
&  =\sum_{i=1}^n\sum_{k=1}^K \tau^{(m)}_{ik} \left(\ln \alpha_k
+ \ln f_k(\boldsymbol{x}_i|\theta_k)\right), 
\end{align}
where we have introduced the notation
\begin{equation}\label{eq:tau_ik}
\tau^{(m)}_{ik}=\mathbb{P}(z_i=k|\boldsymbol{x}_i,\boldsymbol{\Theta^{(m)}}).
\end{equation}
Finally, we have
\begin{equation}\label{eq:vraisestimation}
Q_p(\boldsymbol{\Theta}|\boldsymbol{\Theta}^{(m)})=\sum_{i=1}^n\sum_{k=1}^K \tau^{(m)}_{ik} \left(\ln \alpha_k
+ \ln f_k(\boldsymbol{x}_i|\theta_k)\right)-\beta 
\sum_{k=1}^K\left\lVert\boldsymbol{\mu_k}\right\rVert_1.
\end{equation}

\subsubsection{M phase}
In the M phase, we maximize
$Q_P(\boldsymbol{\Theta}|\boldsymbol{\Theta}^{(m)})$ with respect to
$\boldsymbol{\Theta}$. To do so, we introduce the following Lagrangian function
\begin{equation}
\mathcal{L}(\boldsymbol{\Theta},\zeta,\boldsymbol{\lambda}|\boldsymbol{\Theta}^{(m)})=Q_p(\boldsymbol{\Theta}|\boldsymbol{\Theta}^{(m)})+\zeta\left(\sum_{k=1}^K \alpha_k-1\right)+\sum_{k=1}^K\lambda_k(1-\left\lVert\boldsymbol{\mu_k}\right\rVert_2^2),\label{eq:lagrangien}
\end{equation}
in which the multipliers enforce the equality constraints. We look for
stationary points of the Lagrangian by setting the partial derivatives with
respect to the parameters to zero. 

A straightforward derivation shows that the partial derivatives of
$\mathcal{L}$ with respect to the $\alpha_k$ are equal to zero if and only if
\begin{equation}
\forall k,\  \alpha_k=\frac{1}{n}\sum_{i=1}^n \tau^{(m)}_{ik}.\label{eq:lagralphaf}
\end{equation}
This is the standard M phase update obtained in
\cite{10.5555/1046920.1088718}, an obvious fact considering that the
penalization term does not apply to the $\alpha_k$. 

The case of the other parameters is more complicated. A derivation provided in
\ref{appendix:em:derivation} shows that for $\boldsymbol{\Theta}$ is a
stationary point of the Lagrangian if for all $k$, $\kappa_k$ and
$\boldsymbol{\mu}_k$ are such that 
\begin{equation}\label{eq:kappa:l1}
\frac{I_{d/2}(\kappa_k)}{I_{d/2-1}(\kappa_k)}=\boldsymbol{\mu}_k^T\frac{\sum_{i=1}^n\tau^{(m)}_{ik}
  \boldsymbol{x}_i}{\sum_{i=1}^n\tau^{(m)}_{ik}},
\end{equation}
and
\begin{equation}\label{eq:mugene}
\mu_{kj}=\frac{\sign{r^{(m)}_{kj}}}{2\lambda_k}\max(\kappa_k\lvert
r^{(m)}_{kj}\rvert-\beta,0),
\end{equation}
where
\begin{equation}\label{eq:rvect}
\boldsymbol{r}^{(m)}_{k}=\sum_{i}\tau^{(m)}_{ik}\boldsymbol{x}_{i},  
\end{equation}
and
\begin{equation}
\lambda_k=\frac{1}{2} \sqrt{\sum_{j=1}^d (\max(\kappa_k\lvert  r^{(m)}_{kj}\rvert-\beta,0))^2} .\label{lambdaverif2}
\end{equation}
Unfortunately, equation \eqref{eq:kappa:l1}, the equations \eqref{eq:mugene} for
all $j$ and equation \eqref{lambdaverif2} are coupled, and no close form
formula can be used to compute directly a solution. 

In the particular case where $\beta=0$ (i.e. no regularization), $\mu_{kj}$
simplifies to $\frac{\kappa_kr^{(m)}_{kj}}{2\lambda_k}$, which 
implies $\lambda_k=\frac{1}{2}\sqrt{\sum_{j=1}^d
  \kappa^2_k(r^{(m)}_{kj})^2}$. In turns this simplifies to
\begin{equation*}
\boldsymbol{\mu}_k= \frac{ \sum_{i}\tau^{(m)}_{ik}\boldsymbol{x}_{i}}{\left\lVert\sum_{i}\tau^{(m)}_{ik}\boldsymbol{x}_{i}\right\rVert_2},
\end{equation*}
and thus $\boldsymbol{\mu}_k$ does not depend on $\kappa_k$. This is used in
\cite{10.5555/1046920.1088718} to obtain closed form equations for the M
phase.

However in our case where $\beta>0$, we cannot leverage such uncoupling of the
equations. Therefore we propose to solve the M phase approximately, using a
fixed point strategy. Using the current estimate of $\kappa_k$, we compute an
updated estimation of $\boldsymbol{\mu}_k$ using equations \eqref{eq:mugene}
and \eqref{lambdaverif2}. Then we update $\kappa_k$ using the estimator recalled in Section
\ref{sec:maxim-likel-estim}, i.e. 
\begin{equation}
  \label{eq:kappa:estim}
\kappa_k=\frac{d\rho_k-\rho_k^3}{1-\rho_k^2},
\end{equation}
with
\begin{equation}
 \rho_k= \frac{\boldsymbol{\mu}_k^T \boldsymbol{r}^{(m)}_k}{\sum_{i}\tau^{(m)}_{ik}}.
\end{equation}
As pointed out in Section \ref{sec:maxim-likel-estim}, more advanced numerical
schemes can be used to estimate $\kappa_k$. They can be plugged in the EM
algorithm without any difficulty as they simply solve equation
\eqref{eq:kappa:l1}. 

We iterate those two updates until convergence. Notice that to enforce
consistency of this strategy with the closed form equations from
\cite{10.5555/1046920.1088718} in the case where $\beta=0$, we must update
$\boldsymbol{\mu}_k$ and then $\kappa_k$. The reverse sequence does not
generate consistent updates. 

The final EM algorithm is summarised in Algorithm
\ref{code:EM}. Implementation details are discussed in
\ref{appendix:implementation}. 

\begin{algorithm}[H]
  \caption{EM for penalized likelihood estimation}\label{code:EM}
  \begin{algorithmic}
    \REQUIRE{$\beta\geq 0$ (the regularisation parameter)}
    \REQUIRE{$\boldsymbol{\Theta}_{init}$ (an optional initialisation value for $\boldsymbol{\Theta^{(0)}}$)}
    \STATE{Initialise $\boldsymbol{\Theta^{(0)}}$ to
      $\boldsymbol{\Theta}_{init}$ or randomly (see Algorithm \ref{code:EM:init})}
    \STATE{$m\leftarrow 0$}
    \REPEAT
    \STATE{
\begin{align*}
      \tau^{(m)}_{ik}&\leftarrow\dfrac{\alpha_k^{(m)}f_k(\boldsymbol{x}_i,\theta_k^{(m)})}{\sum_{l=1}^K
        \alpha_l^{(m)} f_l(\boldsymbol{x}_i,\theta^{(m)}_l)} 
&\boldsymbol{r}^{(m)}_{k}&\leftarrow\sum_{i=1}^n\tau^{(m)}_{ik}\boldsymbol{x}_{i}                     
    \end{align*}
    }    
    \STATE{
\begin{align*}
 \alpha^{(m+1)}_k&\leftarrow\frac{1}{n}\sum_{i=1}^n \tau^{(m)}_{ik}      
\end{align*}
}
\STATE{
  \begin{align*}
    \kappa^{(m+1)}_k&\leftarrow \kappa^{(m)}_k
  \end{align*}
  }
  \REPEAT
\STATE{\small
  \begin{equation*}
\mu_{kj}^{(m+1)}\leftarrow\frac{\sign{r^{(m)}_{kj}}}{\sqrt{\sum_{j=1}^d (\max(\kappa^{(m+1)}_k\lvert  r^{(m)}_{kj}\rvert-\beta,0))^2}}\max(\kappa^{(m+1)}_k\lvert
r^{(m)}_{kj}\rvert-\beta,0)
  \end{equation*}
  }
\STATE{
 \begin{align*}
 \rho_k&\leftarrow \frac{{\boldsymbol{\mu}^{(m+1)}_k}^T
         \boldsymbol{r}^{(m)}_k}{\sum_{i=1}^n\tau^{(m)}_{ik}}
&\kappa^{(m+1)}_k&\leftarrow\frac{d\rho_k-\rho_k^3}{1-\rho_k^2}   
\end{align*}
}
  \UNTIL{convergence of $\kappa^{(m+1)}_k$ and $\boldsymbol{\mu_k}^{(m+1)}$}
  \STATE{$m\leftarrow m+1$}
    \UNTIL{convergence of $L(\boldsymbol{\Theta}^{(m+1)}|\boldsymbol{X})$}
  \end{algorithmic}
\end{algorithm}

\subsubsection{Shared $\kappa$}
As shown e.g. in \cite{hornik2014movmf}, in high dimensional settings, the
components of mixtures of vMF tend to overspecialize to subsets of the data as
their concentration parameters become very large. The problem can be reduced
by using a single $\kappa$ parameter shared among all the components. In this
case, the collection of $K$ equations \eqref{eq:kappa:l1} are replaced by the
single equation
\begin{equation}
  \label{eq:kappa:shared}
\frac{I_{d/2}(\kappa)}{I_{d/2-1}(\kappa)}=\frac{1}{N}\sum_{k=1}^K\boldsymbol{\mu}_k^T\left(\sum_{i=1}^n\tau^{(m)}_{ik}
  \boldsymbol{x}_i\right).
\end{equation}
Then equation \eqref{eq:mugene} is replaced by
\begin{equation}
  \label{eq:mugene:shared}
\mu_{kj}=\frac{\sign{r^{(m)}_{kj}}}{2\lambda_k}\max(\kappa\lvert
r^{(m)}_{kj}\rvert-\beta,0),
\end{equation}
and equation \eqref{lambdaverif2} by
\begin{equation}
\lambda_k=\frac{1}{2} \sqrt{\sum_{j=1}^d (\max(\kappa\lvert  r^{(m)}_{kj}\rvert-\beta,0))^2} .\label{lambdaverif2:shared}.
\end{equation}
The rest of Algorithm \ref{code:EM} remains unchanged. 

\subsection{Path following strategy}
Algorithm \ref{code:EM} can be applied for any fixed value of $\beta$. A
possible strategy for exploring the effect of $\beta$ would be to apply the
algorithm from scratch for different values, for instance regularly spaced on a
grid. To reduce the computational burden, improve convergence and provide
consistency between the models, we propose on the contrary to adopt a path
following strategy.

The key idea is to start with a non sparse solution for $\beta=0$ and to increase
progressively the value of $\beta$, restarting each time Algorithm
\ref{code:EM} from the previous solution. In addition, meaningful increments to $\beta$
can be computed from equation \eqref{eq:mugene}: we can indeed look for a
minimal increase of $\beta$ that is guaranteed to increase the sparsity of the
directional means (at least during the first iteration of Algorithm
\ref{code:EM}).

Let us denote $\boldsymbol{\Theta}\givenbeta{\beta}$ the parameter estimated by applying Algorithm
\ref{code:EM} until convergence for a given value of $\beta$. For instance
$\mu_{kj}\givenbeta{0}$ is the $j$-coordinate of directional mean of the $k$ component
when $\beta=0$. By a natural extension $\boldsymbol{r}_{k}\givenbeta{\beta}$ is the result of
applying equation \eqref{eq:rvect} to $\boldsymbol{\Theta}\givenbeta{\beta}$ (using
equations \eqref{eq:tau_ik} and \eqref{eq:pre:stepE_t}). 

To illustrate the calculation of meaningful increments to
$\beta$, let us first consider the initial solution obtained with
$\beta_0=0$ and  let us define $\beta_1$ as follows
\begin{equation}
\beta_1=\min_{1\leq k\leq K, 1\leq j\leq d, \kappa_k\givenbeta{0}\lvert  r\givenbeta{0}_{kj}\rvert>0}\kappa_k\givenbeta{0}\lvert  r_{kj}\givenbeta{0}\rvert.
\end{equation}
Let us consider $0<\beta<\beta_1$ and the first iteration of Algorithm
\ref{code:EM} initialized with
$\boldsymbol{\Theta}^{(0)}=\boldsymbol{\Theta}\givenbeta{0}$. The E phase does not
depend on $\beta$ and none of the quantities computed in this phase change
from $\boldsymbol{\Theta}\givenbeta{0}$ (as Algorithm \ref{code:EM} converged). This is
also the case for the first part of the M phase and the $\kappa_k$ and the
$\boldsymbol{\mu}_k$ remain unchanged
(e.g. $\kappa_k^{(1)}=\kappa_k\givenbeta{0}$). Then consider the update to
$\mu^{(1)}_{kj}$. According to equation \eqref{eq:mugene}, $\mu_{kj}\givenbeta{0}=0$ can
only be a consequence of $r_{kj}\givenbeta{0}=0$. Then for any value of $\beta>0$,
$\mu^{(1)}_{kj}=0$. On the contrary, if $|r_{kj}\givenbeta{0}|>0$, then
$|\mu^{(1)}_{kj}|>0$ for any $\beta<\beta_1$ as a consequence of the
definition of $\beta_1$ and of equation \eqref{eq:mugene}. Obviously
$|\mu^{(1)}_{kj}|<|\mu_{kj}\givenbeta{0}|$ because of the shrinkage effect induced by
$\beta>0$ in equation \eqref{eq:mugene}, but unless $\beta\geq \beta_1$, the
directional mean sparsity will not increase during this first step of the
algorithm. The full effects of setting $\beta$ to a non zero value cannot be
predicted from this simple analysis, and the sparsity might increase because
of the modification of the $\kappa_k$ and of the $\tau_{ij}$ induced by the
shrinkage. Nevertheless, setting $\beta$ to $\beta_1$ is the smallest increase
from $\beta_0$ that is \emph{guaranteed} to increase the sparsity of the
solution during the \emph{first step} of the algorithm. 

A similar reasoning shows that we can guarantee an increase in sparsity (in
the first step of the algorithm) when starting with
$\boldsymbol{\Theta}\givenbeta{\beta_{p-1}}$ by choosing $\beta_p$ given by
\begin{equation}
  \label{cheminbeta}
\beta_p=\beta_{p-1}+\min_{h,j,\kappa_k\givenbeta{\beta_{p-1}}|r_{kj}\givenbeta{\beta_{p-1}}|-\beta_{p-1}>0}\kappa_k\givenbeta{\beta_{p-1}}|r_{kj}\givenbeta{\beta_{p-1}}|-\beta_{p-1}.
\end{equation}
In practice, we propose to start with $\beta_0=0$ and to iterate updates based
on equation \eqref{cheminbeta} to generate a series of solutions. 
To avoid taking too many steps on this path, we set values smaller than the 
chosen numerical precision threshold to zero after updating $\beta$. The final
path following algorithm is given in Algorithm \ref{code:path}. In this
summary, $EM(\beta)$ is a call to Algorithm \ref{code:EM} with a random
initialisation for $\boldsymbol{\Theta^{(0)}}$, while
$EM(\beta,\boldsymbol{\Theta})$ uses $\boldsymbol{\Theta}$ as the initial
value of $\boldsymbol{\Theta^{(0)}}$. 

In practice, the number of steps taken by the algorithm can be as high as the
number of dimensions multiply by $K$, when coordinates are set to zero almost
one by one. In order to reduce the computational burden, one can enforce
minimal (relative) increase of $\beta$ between two steps. It is also possible
to limit the path to $P$ steps (as in Algorithm \ref{code:path}) or to keep
exploring it until the maximal sparsity is reached (only one non zero
parameter per directional mean). Those heuristics will be used in the
experiments. 

\begin{algorithm}[htb]
  \caption{Path following}\label{code:path}
  \begin{algorithmic}
    \REQUIRE{$P>0$ (the number of $\beta$ to explore on the path)}
    \REQUIRE{$\epsilon>0$ (the numerical precision below which directional
      means coordinates are set to 0)}
    \STATE{$\beta_0\leftarrow 0$}
    \STATE{$\boldsymbol{\Theta}\givenbeta{0}\leftarrow EM(\beta_0)$}
    \FOR{$p=1$ \TO $P-1$}
    \STATE{\begin{equation*}
        \beta_p\leftarrow\beta_{p-1}+\min_{h,j,\kappa_k\givenbeta{\beta_{p-1}}|r_{kj}\givenbeta{\beta_{p-1}}|-\beta_{p-1}>0}\kappa_k\givenbeta{\beta_{p-1}}\bigl|r_{kj}\givenbeta{\beta_{p-1}}\bigr|-\beta_{p-1}
      \end{equation*}
    }
    \STATE{
\begin{equation*}      
      \boldsymbol{\Theta}\givenbeta{\beta_p}\leftarrow
      EM(\beta_p,\boldsymbol{\Theta}\givenbeta{\beta_{p-1}})
    \end{equation*}
  }
  \IF{$|\mu_{kj}|<\epsilon$}
  \STATE{$\mu_{kj}\leftarrow 0$}
  \ENDIF
    \ENDFOR
  \end{algorithmic}
\end{algorithm}

\subsection{Model selection}\label{sec:model-selection}
Following \cite{BouberimaEtAl2010,salah:tel-01835699} we propose to use 
information criteria for model selection, especially in order to set the value of $\beta$. Former studies
\cite{BouberimaEtAl2010,salah:tel-01835699} have been somewhat inconclusive
concerning the ability of the Akaike
Information Criterion \cite{Akaike1998} (AIC), the Bayesian Information Criterion
\cite{schwarz1978estimating} (BIC) and their variants to select systematically
an appropriate number of components. For mixtures of vMF, the AIC tends to
overfit by selecting too many components, while the BIC tends to underfit
unless the number of observations is sufficient large (several times the
number of dimensions). For the co-clustering variant of mixtures of vMF
proposed in \cite{pmlr-v51-salah16}, AIC seems to be the most appropriate
solution considering the small number of free parameters of this model (see
\cite{salah:tel-01835699}).

The limitations of the AIC and of the BIC in high dimensional settings is well
known, and several variations have been proposed to address the problem in the
context of supervised learning (mainly linear regression). Variants include
the Risk Inflation Criterion (RIC, \cite{10.1214/aos/1176325766}) and its
specific extension to high dimensional settings the RICc \cite{sam.10088}, as
well as the extended BIC (EBIC \cite{10.1093/biomet/asn034,10.2307/24310025}). Other variants
can be found in e.g. \cite{BouberimaEtAl2010}. 

The general formula for those criteria is given by
\begin{equation}
  \label{eq:all:IC}
IC(\boldsymbol{\Theta}\givenbeta{\beta})= \phi(n, d)\times C(\boldsymbol{\Theta}\givenbeta{\beta}) -2 \times \log L(\boldsymbol{\Theta}\givenbeta{\beta}|\boldsymbol{X}),
\end{equation}
where $C(\boldsymbol{\Theta}\givenbeta{\beta})$ is the number of free
parameters in the model and $\phi(n, d)$ is a criterion dependent
coefficient that may depend on the number of observations $n$ and their
dimension $d$. Table \ref{tab:IC:coefficients} gives the definition of the
coefficient function for a selection of the criteria considered in the present
paper. 

\begin{table}[htbp]
  \centering
  \begin{tabular}{lc}\toprule
    Criterion   & $\phi(n, d)$\\\midrule
    AIC \cite{Akaike1998} & 2 \\
    BIC \cite{schwarz1978estimating} & $\log n$\\
    RIC \cite{10.1214/aos/1176325766} & $2\log d $\\
    RICc \cite{sam.10088} & $2(\log d  + \log\log d)$\\
    EBIC \cite{10.1093/biomet/asn034} & $\log n + 2\gamma\log d$\\\bottomrule
  \end{tabular}
  \caption{Coefficients for the different criteria: $n$ is the number of
    observations and $d$ their dimension. The parameter $\gamma$ of the EBIC
    is set to $0.5$ as recommended in \cite{10.1093/biomet/asn034}.}
  \label{tab:IC:coefficients}
\end{table}

When $\beta=0$, $C(\boldsymbol{\Theta}\givenbeta{0})$ is easy to compute.
When the $\boldsymbol{\kappa}$ are unconstrained, they contribute $K$ free
parameters (and a single parameter when a common $\kappa$ is used). The
$\boldsymbol{\alpha}$ sum to one, and thus contribute $K-1$ free
parameters. When $\beta=0$, the directional means are simply constrained by
their unitary norm and thus contribute $K(d-1)$ free
parameters\footnote{Notice that \cite{BouberimaEtAl2010,salah:tel-01835699}
  overlook the unitary norm constraint and consider $Kd$ parameters.}.

Unfortunately, estimating the number of free parameters for the directional
means under regularisation is not obvious. It
has been shown in \cite{Zou_2007} that in the case of lasso regression, a
consistent estimator of the degree of freedom of the model is given by
counting the number of non-zero terms in the regression. However, the authors
emphasize that this result does not generalize to other settings, such as for
instance elastic net. As a consequence, we propose to use as the number of
free parameters for a given directional mean $\boldsymbol{\mu}$
\begin{equation}
C_{dm}(\boldsymbol{\mu}_k)=\max\left(1,\sum_{j=1}^d\mathbb{I}_{\mu_{kj}\neq 0}-1\right),
\end{equation}
in which $\mathbb{I}$ is the characteristic function. In the particular case
where only a single coordinate is non zero because of a strong regularisation,
the unitary norm constraint reduces the set of possible values for this
coordinate to $\{-1, 1\}$. We still consider this as a free parameter and thus
we set $C(\boldsymbol{\mu})$ to one in this particular case (hence the $\max$
operator in the definition). Then the number of free parameters is given by
\begin{equation}
  \label{eq:effeciveparam}
C(\boldsymbol{\Theta}\givenbeta{\beta})=(2K-1)+\sum_{k=1}^KC_{dm}(\boldsymbol{\mu}_{k}\givenbeta{\beta}).
\end{equation}
In practice, we propose to use the BIC or the AIC to select the optimal
$\beta$ on the regularisation path. We propose to use other criteria as guides
for selecting interesting configurations in terms of the number of components
in the mixture. Because of the inherent difficulty in estimating a model in
the high dimension low number of observations case, we cannot recommend to
focus on a single criterion. 

\subsection{Exploratory use}\label{sec:exploratory-use}
Once a the parameters of the model have been estimating, they can be used for
two exploratory tasks.

Firstly, As is classical in mixture models, the $\tau^{(m)}_{ik}$ from
Equation \eqref{eq:tau_ik} can be used to define
a hard/crisp clustering of the observations into $K$ clusters. The cluster
index of observation $\boldsymbol{x}_i$, $c^{(m)}_i$, is given by
\begin{equation}
c^{(m)}_i=\arg\max_{1\leq k\leq K }  \tau^{(m)}_{ik}.
\end{equation}
They can also be used directly to detect ambiguous assignments.

Secondly, the directional means themselves can provide interesting insights on
the data. As we consider high dimensional data, a direct analysis is
difficult and we propose to rely on a graphical representation, as used in
e.g. \cite{pmlr-v51-salah16,salah2017model}. The key idea is to represent the
directional means (or the full data set) as an image in which the grey level
of a pixel encodes the value of a coordinate: the $j$-pixel of the $i$-th row
of the image represents $\mu_{ij}$ (or $x_{ij}$). This type of pixel-oriented
visualisation \cite{Keim2000} must use some form of ordering to provide
insights on the data.

The coordinates are ordered with the help of the sparsity pattern of the
directional means. We introduce first a binary version of the directional
means given by
\begin{equation}
  \label{eq:binary:proto}
b_{kj}=\mathbb{I}_{\mu_{kj}\neq 0},
\end{equation}
and the counts of non zero coordinates
\begin{equation}
  \label{eq:column:order:quantity}
n_j=\sum_{k=1}^Kb_{kj}.
\end{equation}
Then we use lexical ordering defined as follows. Dimension $j$ is smaller than
dimension $j'$, $j\prec j'$, if
\begin{enumerate}
\item $n_j>n_{j'}$: we start with dimensions that are non zero for all
  directional means;
\item or when $n_j=n_{j'}$:
  \begin{enumerate}
  \item if $\exists k$ $\forall l<k$ $b_{lj}=b_{lj'}$ and $b_{kj}>b_{kj'}$
  \item or $\forall k b_{kj}=b_{kj'}$ and 
  \begin{equation}
\sum_{k=1}^K\left|\mu_{kj}\right|>\sum_{k=1}^K\left|\mu_{kj'}\right|.
  \end{equation}
  \end{enumerate}
\end{enumerate}
Inside a block of dimensions with the same $n_j$,
dimensions are grouped based on common non zero pattern (i.e. on identical
$b_{kj}$) and then on the intensity of the non zero coordinates. This ordering
is somewhat arbitrary but it leads to readable pixel representations. In
particular, it emphasizes common non zero values (i.e. common vocabulary in the
case of text data) as well as exclusive dimension (i.e. words used only by some
texts). To further emphasize the different groups of dimensions, we chose for
each group of pixels with the same $n_j$ a different hue. 

Rows are ordered in decreasing size of the corresponding clusters,
i.e. according to the $\alpha_k$. Both ordering can be applied to the data
set. In this case, we use an arbitrary ordering of the observations inside
their cluster.

\subsection{Summary and proposed methodology}\label{sec:summary}
In summary, we propose to build a sparse mixture of vMF as follows:
\begin{enumerate}
\item select a set of candidate values for $K$ the number of components
  $\mathcal{K}$;
\item for each $K\in\mathcal{K}$
  \begin{enumerate}
  \item run algorithm \ref{code:path} to obtain a collection of regularisation
    values and their associated parameters
    $\boldsymbol{\Theta}_K\givenbeta{\beta}$;
  \item keep the dense solution $\boldsymbol{\Theta}_K\givenbeta{0}$ and the
    best sparse solution $\boldsymbol{\Theta}_K\givenbeta{\beta_K^*}$
    according to the AIC/BIC;
  \end{enumerate}
\item select the best $K$, $K^*$, according to an information criterion applied to the
  dense model $\boldsymbol{\Theta}_K\givenbeta{0}$;
\item the final model is described by $\boldsymbol{\Theta}_{K^*}\givenbeta{\beta_{K^*}^*}$.
\end{enumerate}
In Sections \ref{sec:exp} and \ref{sec:exper-real-world}, we study this
procedure and compare it with variations.

\section{Analysis of  the proposed methodology}\label{sec:exp}
We present in this Section experiments that illustrate the behavior of our
methodology on simulated data. Banerjee et al. already demonstrated in
\cite{10.5555/1046920.1088718} the interest of the mixture of von Mises-Fisher
distribution compared to other clustering solutions for directional
data. Therefore the main focuses of our evaluation are the behavior of the
path following algorithm, the effects of the regularisation approach and the
relevance of the information criteria for model selection. Our goal is to
justify the choices that lead to the procedure proposed in Section
\ref{sec:summary}. 

The section is structured as follows. The data generation procedure is
described in Section \ref{sec:simul-data-gener}. Section
\ref{sec:path-foll-strat} discusses the behavior of the path following
strategy on a simple example and shows that this strategy is preferable to
alternative solutions such as a grid based search. Section
\ref{sec:simulation-study} analyses in details the behavior of the proposed
model using a medium scale simulation study. 

\subsection{Simulated data generation}\label{sec:simul-data-gener}
To study the behavior of the model, we use simulated data sets that are
generated by mixtures of von Mises-Fisher distributions. We generate the
parameters of the distributions in a semi-random way that enables us to
control the separation between the components. The general procedure for a
mixture of $K$ components in dimension $d$ is the following one:
\begin{itemize}
\item we sample $20\times K$ random vectors uniformly on the unit hypersphere
  $\mathbb{S}^{d-1}$;
\item we extract from those vectors $K$ maximally separated vectors by
  minimizing their pairwise inner products in a greedy way: those are the
  directional means of the mixture $(\boldsymbol{\mu}_k)_{1 \leq k\leq K}$;
\item in most of the simulations, we sparsify the directional means by setting
  to zero a randomly selected subset of their coordinates. We make sure to
  keep non zero directional means and to have them all distinct;
\item we chose $\boldsymbol{\kappa}$ in such a way to ensure a given degree of
  overlapping between the components. The overlapping is measured as the error
  rate of crisp assignments obtained by the model using the true parameters
  compared to the ground truth. For a dimension $d=100$, we use a base
  $\kappa=17.34$ to obtain $2.5\%$ of overlapping, and $\kappa=15.09$ to
  obtain $5\%$ of overlapping.
\item for each component, $\kappa_k$ is sampled from the Gaussian distribution
  $\mathcal{N}(\mu=\kappa,\sigma=0.025\times\kappa)$;
\item the final concentration of each component of the mixture is adjusted
  for intrinsic separability. This consists in using
$\kappa'_k$ defined by
\begin{equation}
\kappa'_k=\frac{2\kappa_k}{1-\max_{l\neq k}\boldsymbol{\mu}_k^T\boldsymbol{\mu}_l}.\label{eq:kappa:scaling}
\end{equation}
\end{itemize}
For simplicity, we use systematically a balanced mixture with
$\alpha_k=\frac{1}{K}$. 

\subsection{Path following strategy illustration}\label{sec:path-foll-strat}
We illustrate in this section the behavior of the path following algorithm
(Algorithm \ref{code:path}) on a simple example. We use $K^*=4$
components in dimension $d=10$, with a separation of $5\%$ ($\kappa=5.37$). We
generate $N=500$ observations, which makes the estimation relatively easy
considering the low dimension of the data (we do not introduce sparsity in
the directional means). We run our path following algorithm from the best
configuration (in terms of likelihood) among 10 random initial configurations.

\begin{figure}[htbp]
  \centering
\includegraphics{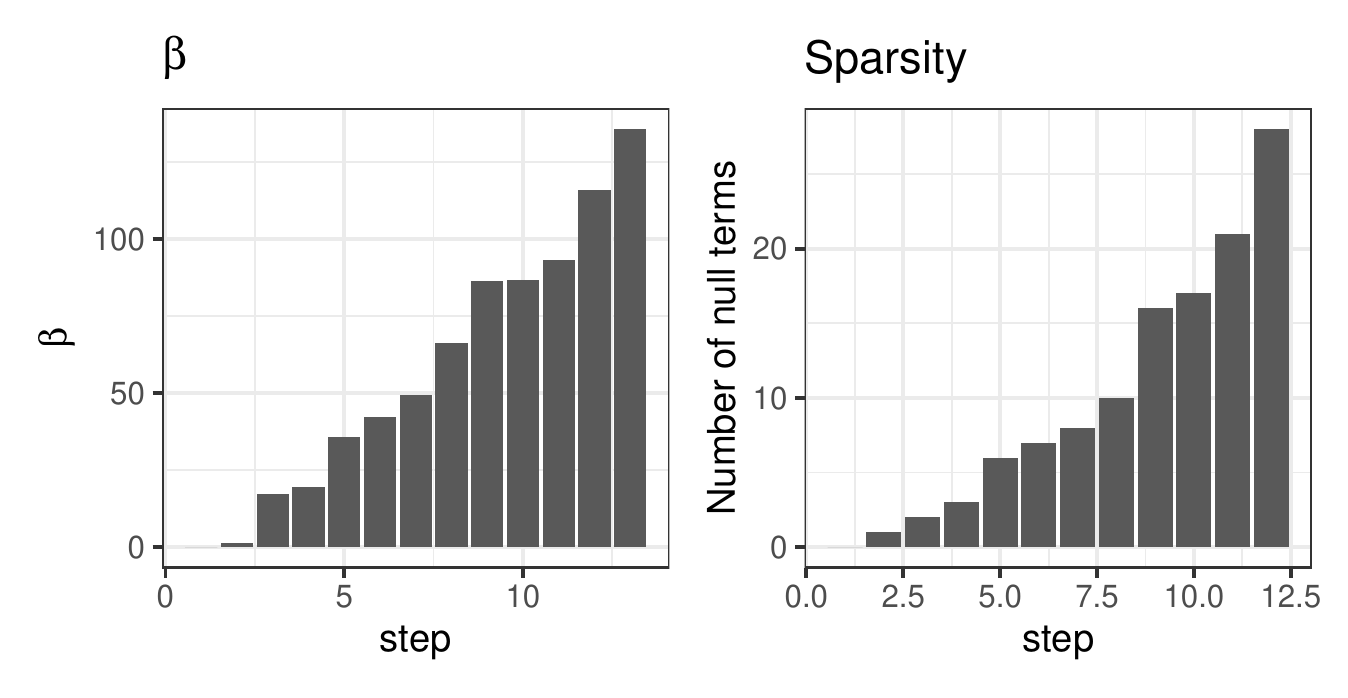}  
  \caption{Evolution of $\beta$ and of the sparsity of the solution during the
    path following algorithm. }
  \label{fig:path:beta}
\end{figure}

\begin{figure}[htbp]
  \centering
\includegraphics{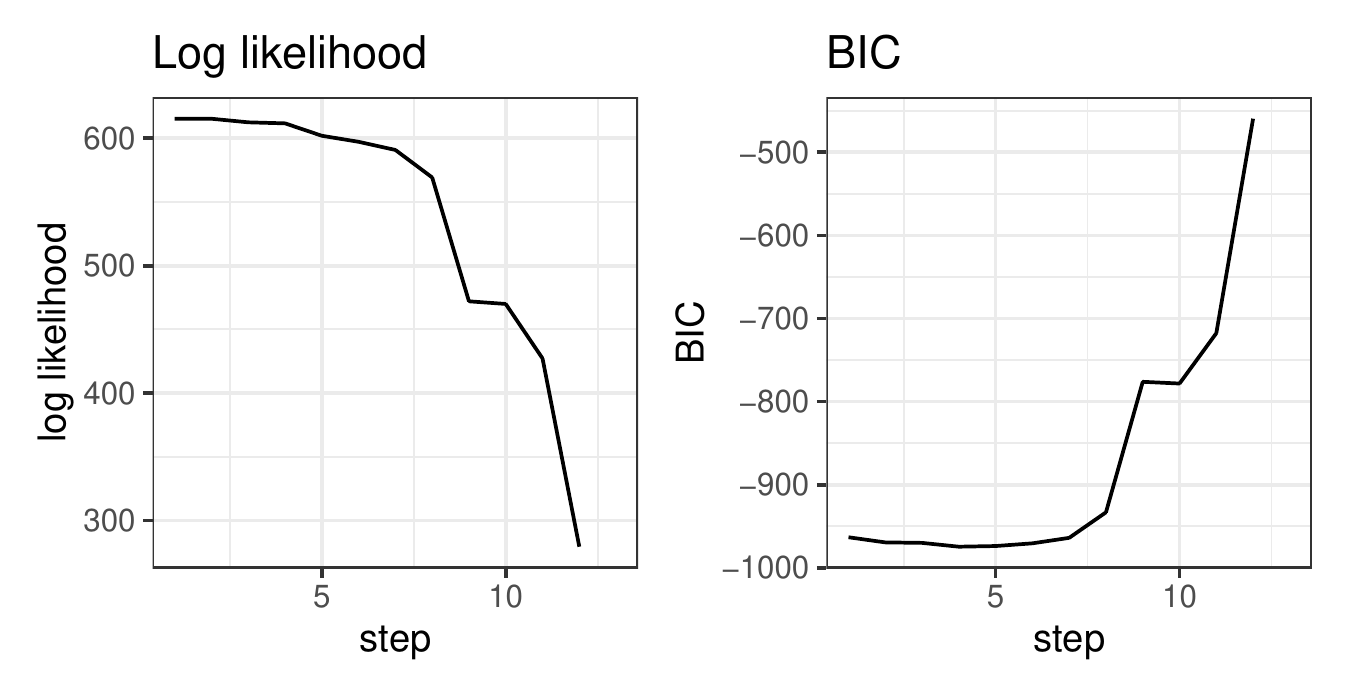}  
  \caption{Evolution of the log likelihood and of the BIC of the solution during the
    path following algorithm. }
  \label{fig:path:BIC}
\end{figure}

Figures \ref{fig:path:beta} and \ref{fig:path:BIC} show the behavior of the
algorithm. In this particular example, the path contains 13 steps. During the
final step, the EM algorithm did not converge to a configuration with 4 components,
as expected when the sparsity becomes too important. While no sparsity was
enforced during the generation of the data set, it was nevertheless worthwhile 
to set some of the components to zero as it lead to a small decrease of the
BIC (around step 5). 

To evaluate the interest of the path following algorithm on this simple
example, we compare four different approaches:
\begin{enumerate}
\item our proposed path following Algorithm \ref{code:path};
\item directly applying the EM Algorithm \ref{code:EM} using the $\beta$s
  computed by the path, restarting each time the algorithm from the dense
  initial configuration ($\beta=0$) used by the path following algorithm;
\item directly applying the EM Algorithm \ref{code:EM} using the $\beta$s
  computed by the path, starting from 10 random initial configurations for each
  $\beta$;
  \item directly applying the EM Algorithm \ref{code:EM} using a regular grid
    of 50 values for $\beta$ between 0 and the maximum value obtained by the
    path following algorithm, restarting each time the algorithm from the dense
  initial configuration ($\beta=0$). 
\end{enumerate}

Solution 2 generates exactly the same estimates as the ones obtained by the
path following algorithm but in a longer running time (25\%
more iterations of the EM algorithm). 

\begin{figure}[htbp]
  \centering
\includegraphics{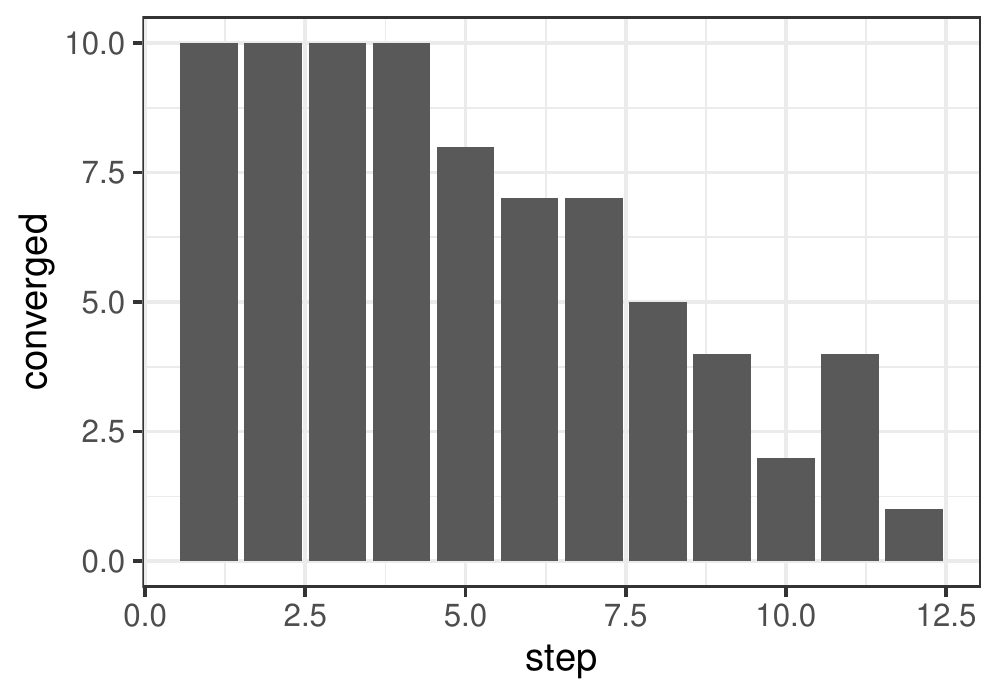}  
  \caption{Number of converging EM runs (among 10) in solution 3 as a function of $\beta$
    (represented here by the step in the path following algorithm). }
  \label{fig:path:converged}
\end{figure}

Solution 3 generates also identical results as the ones obtained by the
path following algorithm. However, we used obviously roughly ten times more
computational resources and in addition a large number of the initial
configurations did not allow the EM algorithm to converge for larger values
of $\beta$, as seen on Figure \ref{fig:path:converged}. 

\begin{figure}[htbp]
  \centering
\includegraphics{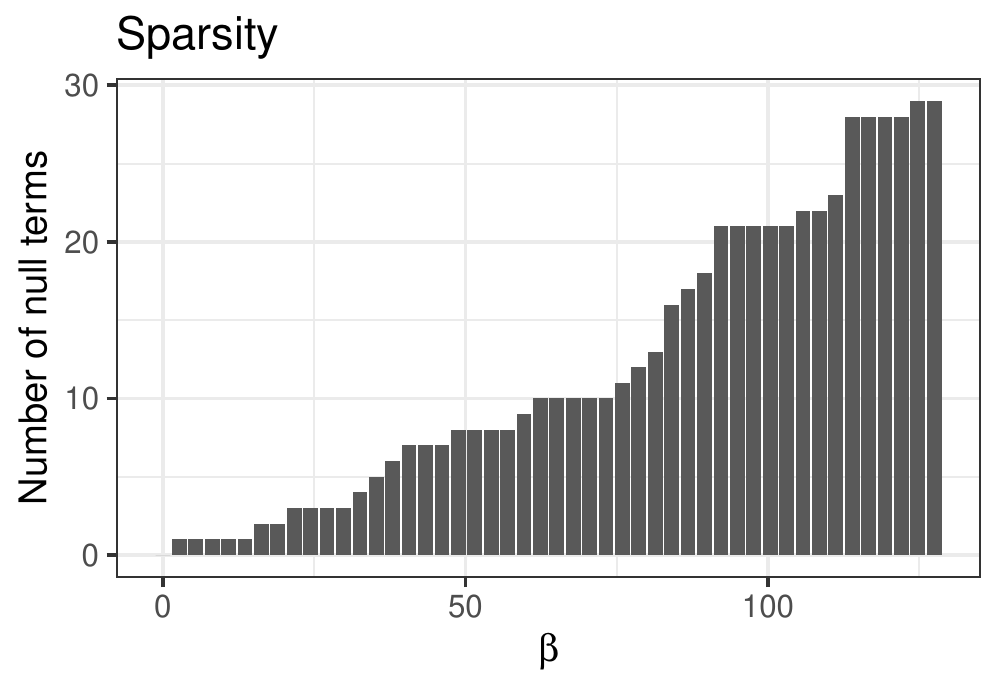}  
  \caption{Sparsity of the solution as a function of $\beta$. }
  \label{fig:grid:beta}
\end{figure}

\begin{figure}[htbp]
  \centering
\includegraphics{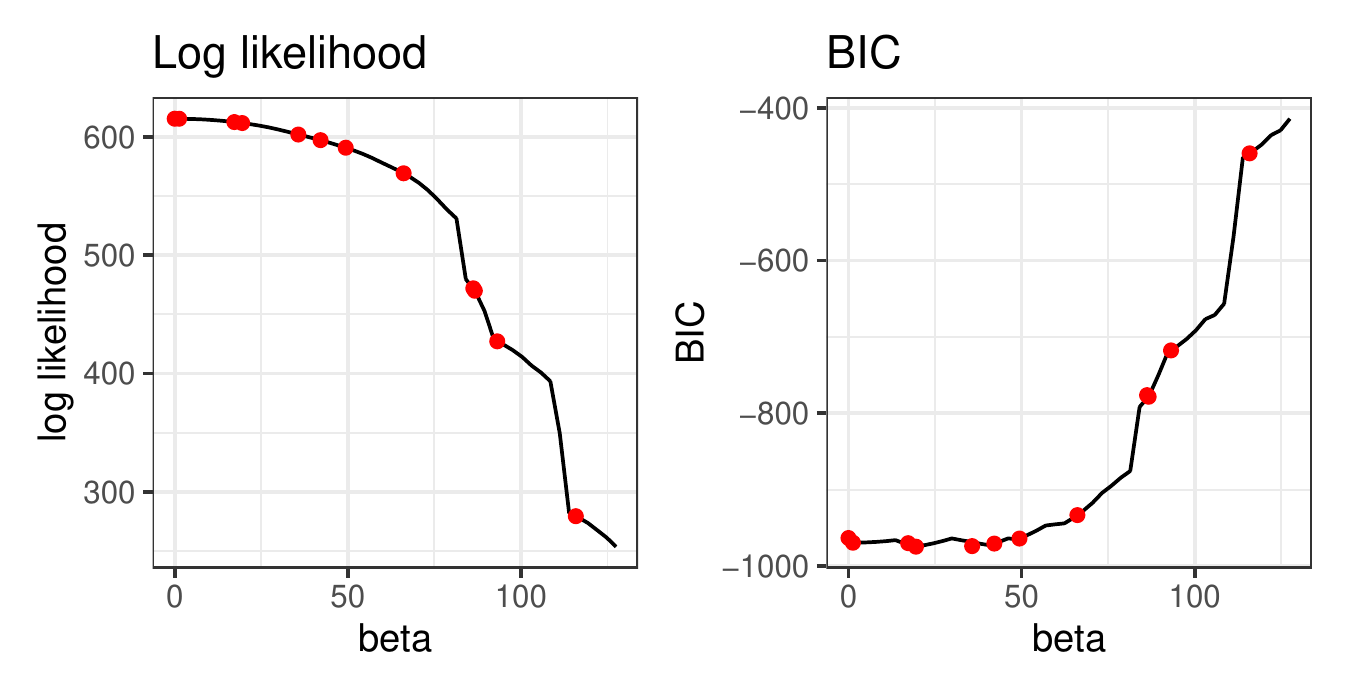}  
  \caption{Log likelihood and of the BIC of the solution as a function of
    $\beta$. The red dots are the configurations obtained by the path
    following algorithm.}
  \label{fig:grid:BIC}
\end{figure}

Notice finally that the values of $\beta$ are quite unpredictable. Without the
path following strategy, we would have had to study the effect of $\beta$s
sampled from an arbitrary grid of values, as tested in solution 4. Results are
presented on Figures \ref{fig:grid:beta} and \ref{fig:grid:BIC}. They show an
identical behavior of the grid based search and of the path following
algorithm in terms of likelihood and BIC. Some sparsity levels might be missed
during the path following (compare Figure \ref{fig:grid:beta} and Figure
\ref{fig:path:beta}), but this is easily fixable by testing some additional
values for $\beta$ inside intervals where the jump in sparsity is large.

In summary, the path following algorithm provides efficiently a good sampling
of the values of $\beta$ that have a significant effect on the sparsity of the
solution. If finer grain analysis is needed, one can sample the intervals
between values on the path that show a large modification in the sparsity of the solution. 

\subsection{Simulation study}\label{sec:simulation-study}
In this section, we study in a more systematic way the behavior of the
proposed methodology. Our goal is to evaluate the computational burden of
testing several $\beta$s via the path following strategy (Section
\ref{sec:path-characteristics}), to confirm and complement previous results about model
selection with information criteria (Section \ref{sec:over-model-select}), to
study to what extent those criteria can be used to select an optimal $\beta$
(Section \ref{sec:select-path-spars}) and finally to assess the difficulty of
recovering a planted sparse structure (Section \ref{sec:sparse-direct-mean}).

The study is based on the $d=100$ dimensional case, with $K^*=4$ components
and for two degrees of overlapping between the components (2.5 \% and 5 \%),
three level of sparsity in the directional means ( 5 \%, 10 \% and 15 \%) and
two data size (200 and 1000 observations, respectively). Notice that while the
directional means are sparse, this is not the case of the observations
themselves unless the $\boldsymbol{\kappa}$s are set to significantly larger
values than the ones we use. We report here only the results obtained for
component specific values of $\boldsymbol{\kappa}$ as the ones obtained with a
shared $\kappa$ do not depart significantly from them in this setting.

We report statistics obtained by generating 100 data sets for each of the
configurations under consideration. In each run, the model is obtained by
running the EM algorithm from ten random initial configurations (see
\ref{appendix:implementation}) and by keeping the best final configuration according to
the (penalized) likelihood. The path following algorithm is started from this
best configuration and is parameterised to ensure
a minimum relative increase of $10^{-3}$ between two consecutive values of
$\beta$.

\subsubsection{Path characteristics and computational burden}\label{sec:path-characteristics}
\begin{figure}[htbp]
  \centering
\includegraphics{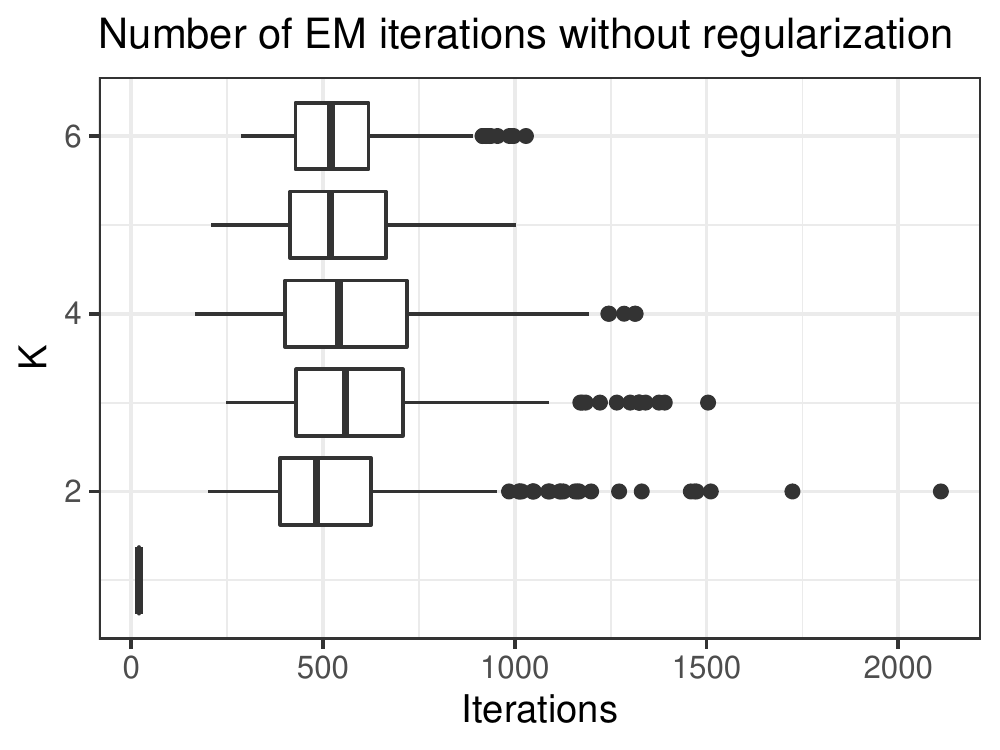}  
  \caption{Distributions of the number of EM iterations needed to obtain the
    first model with $\beta=0$ over 600 data sets with $d=100$ and
  $N=200$, as a function of $K$, the number of components. The figure
  aggregates results for all values of the separation and the sparsity.}
  \label{fig:path:steps:base:iter:200}
\end{figure}

\begin{figure}[htbp]
  \centering
\includegraphics{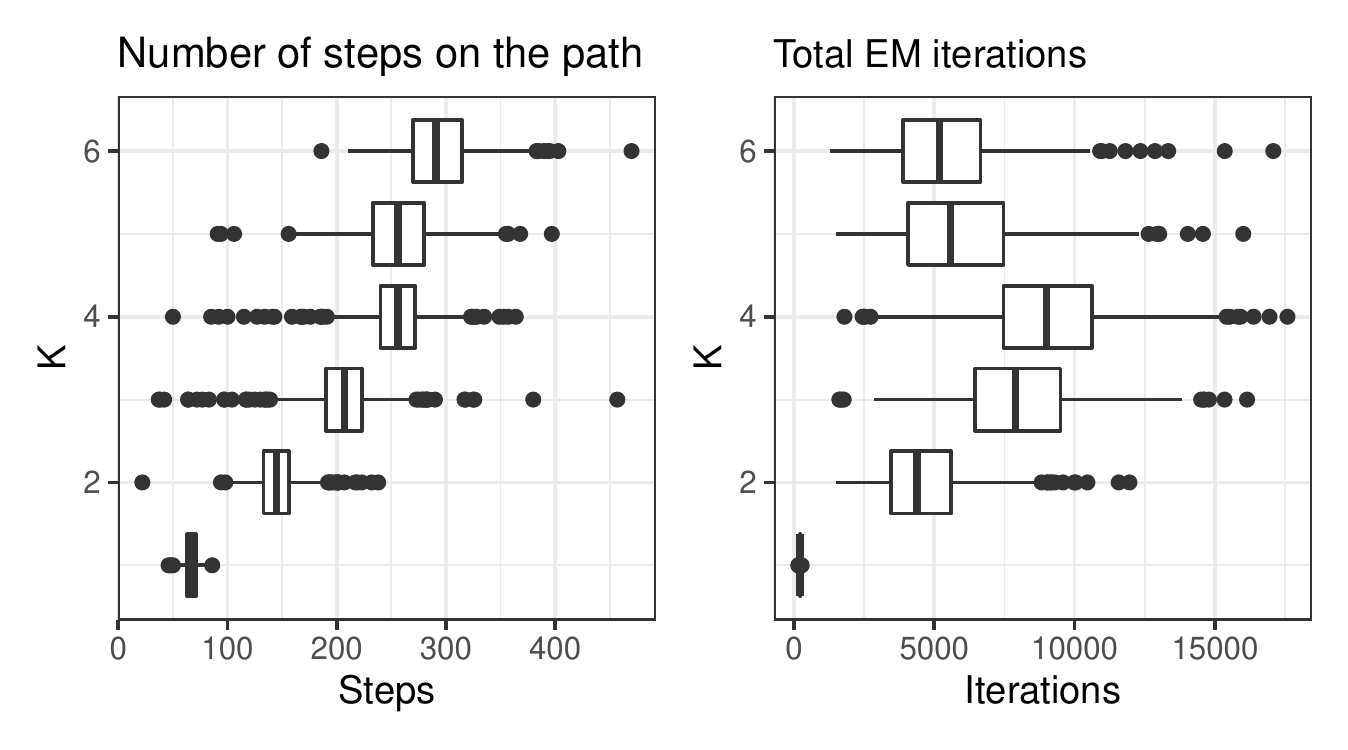}  
  \caption{Distributions of the number of steps (a.k.a. values of $\beta$) and
  of the total number of EM iterations over 600 data sets with $d=100$ and
  $N=200$, as a function of $K$, the number of components. The figures
  aggregate results for all values of the separation and the sparsity.}
  \label{fig:path:steps:iter:200}
\end{figure}

The behavior of the path following algorithm is summarized by Figure
\ref{fig:path:steps:iter:200} which shows the distribution of the number of
steps taken on the path as well as the distribution of the total number of
iterations of the EM algorithm. Compared to the dense case (i.e. to the
initialisation of the algorithm) represented on Figure
\ref{fig:path:steps:base:iter:200}, following the path increases significantly
the computational burden. However, the increase is far less important that
what could be expected from the number of different values of $\beta$
considered during the path exploration. Indeed, the median number of EM
iterations needed to obtain an initial dense configuration is larger than $500$
(for $K\geq 2$), while it is smaller than $10000$ for the subsequent path
exploration. This 20 times ratio, is significantly smaller than the median
number of steps (at least $150$ for $K\geq 2$). In other words, restarting
from the previous configuration when $\beta$ is increased is very efficient:
in general the new stable configuration is obtained using a small number of
iterations of the EM, significantly less than the ones needed to obtain the
first dense model. 

The results shown here for $N=200$ observations are representative of the
results obtained with more observations. The number of iterations tend to grow
for larger $K$ when $n$ increases, but that does not change significantly the
number of steps on the path or the ratio between the number of EM iterations in
the dense case and on the path. 

In summary, the simulation confirms the results obtained in Section
\ref{sec:path-foll-strat}: the computational burden of estimating several
models for different values of $\beta$ is large but the path following
strategy helps mitigating this cost.

\subsubsection{Model selection: number of
  components}\label{sec:over-model-select}
As explained in Section \ref{sec:model-selection}, previous studies on
mixtures of vMF have been somewhat inconclusive about the ability of
information criteria to the select the number of components of the mixture. We
confirm the complex behavior of the two main criteria (AIC and BIC) in this
section. 

For each of the 100 replications, we apply the proposed methodology : we keep
the original dense model as a reference. Then we select along the $\beta$ path
the best model according to each of the information criterion presented in
Section \ref{sec:model-selection}. Finally, we report the number of components
selected in this two cases (dense versus sparse) by minimizing the information
criteria. Notice that in the dense case, we have a single model evaluated by
multiple criteria, while in the sparse case, each criterion selects a
different model on the path.

\begin{figure}[htbp]
  \centering
\includegraphics{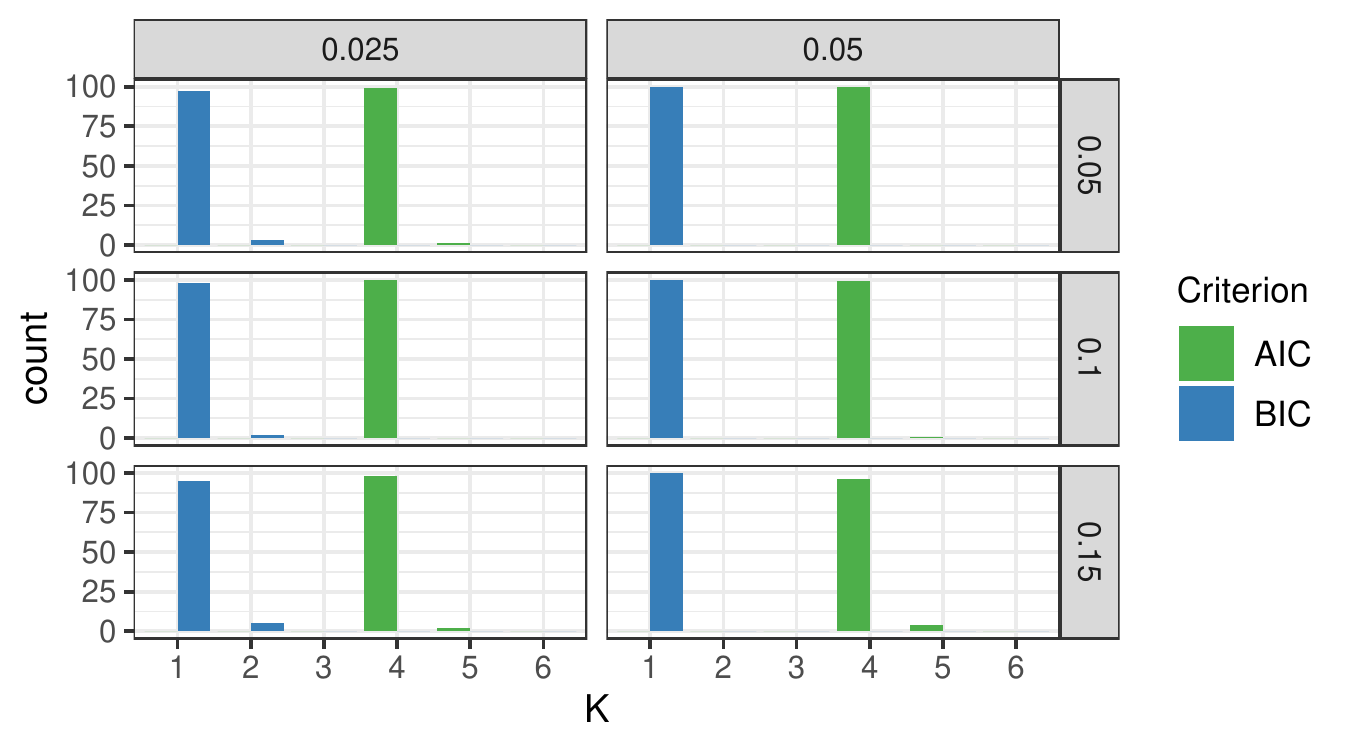}  
  \caption{\textbf{Dense case}: number of times each $K$ is selected as the best configuration by
    AIC or BIC for $\mathbf{N=200}$ observations and $\beta=0$, across overlapping
    values (in column) and directional mean sparsity (in row).}
  \label{fig:mix:sparse:100:200:dense}
\end{figure}

\begin{figure}[htbp]
  \centering
\includegraphics{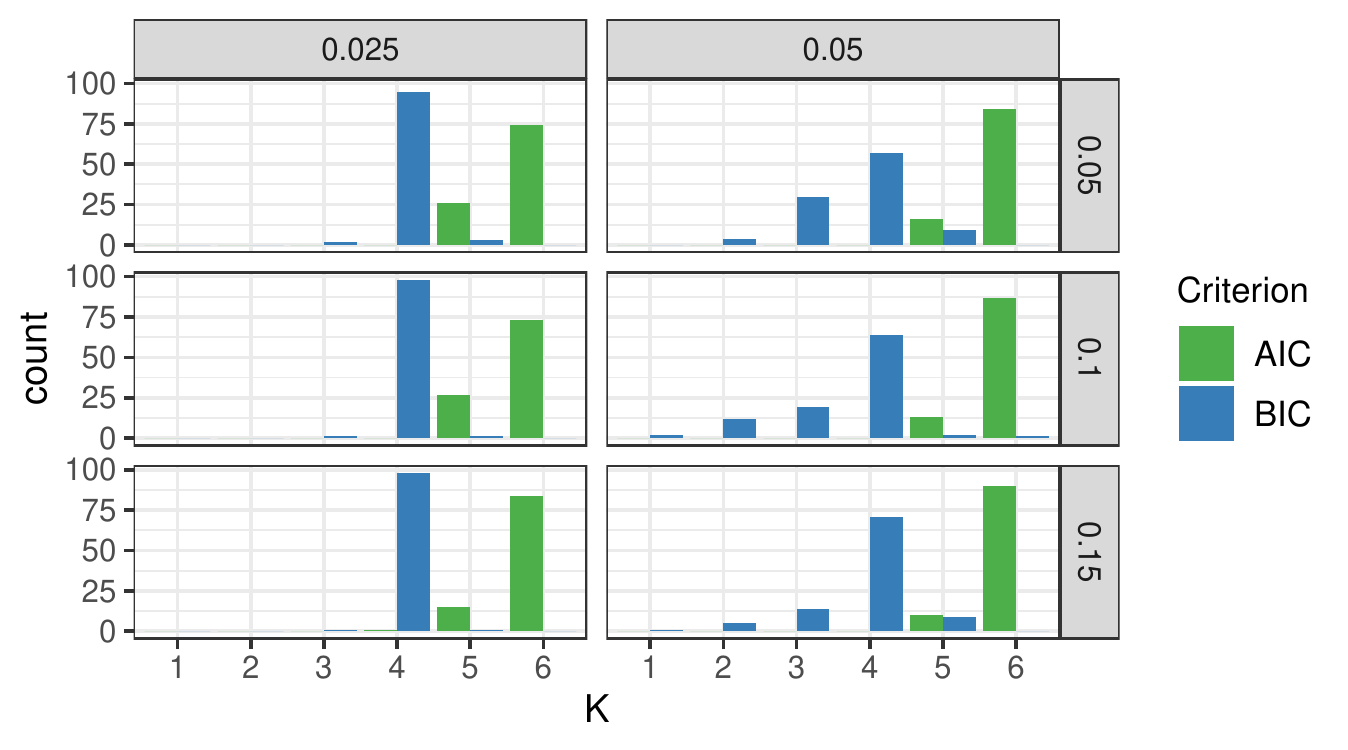}  
\caption{\textbf{Sparse case}: number of times each $K$ is selected as the best configuration by AIC
  or BIC for $\mathbf{N=200}$ observations for the optimal $\beta$ selected by each
  criterion, across overlapping values (in column) and directional mean
  sparsity (in row).}
  \label{fig:mix:sparse:100:200:sparse}
\end{figure}

Figures \ref{fig:mix:sparse:100:200:dense} and
\ref{fig:mix:sparse:100:200:sparse} show the results of this 
approach in the dense case and in the sparse one (for AIC and BIC), with
$N=200$ observations. As the sparsification reduces the number of effective
parameters without reducing too much the likelihood, it favors models with
more components. In this setting, this proves beneficial for the BIC but
drives already the AIC in its overfitting regime. 

\begin{figure}[htbp]
  \centering
\includegraphics{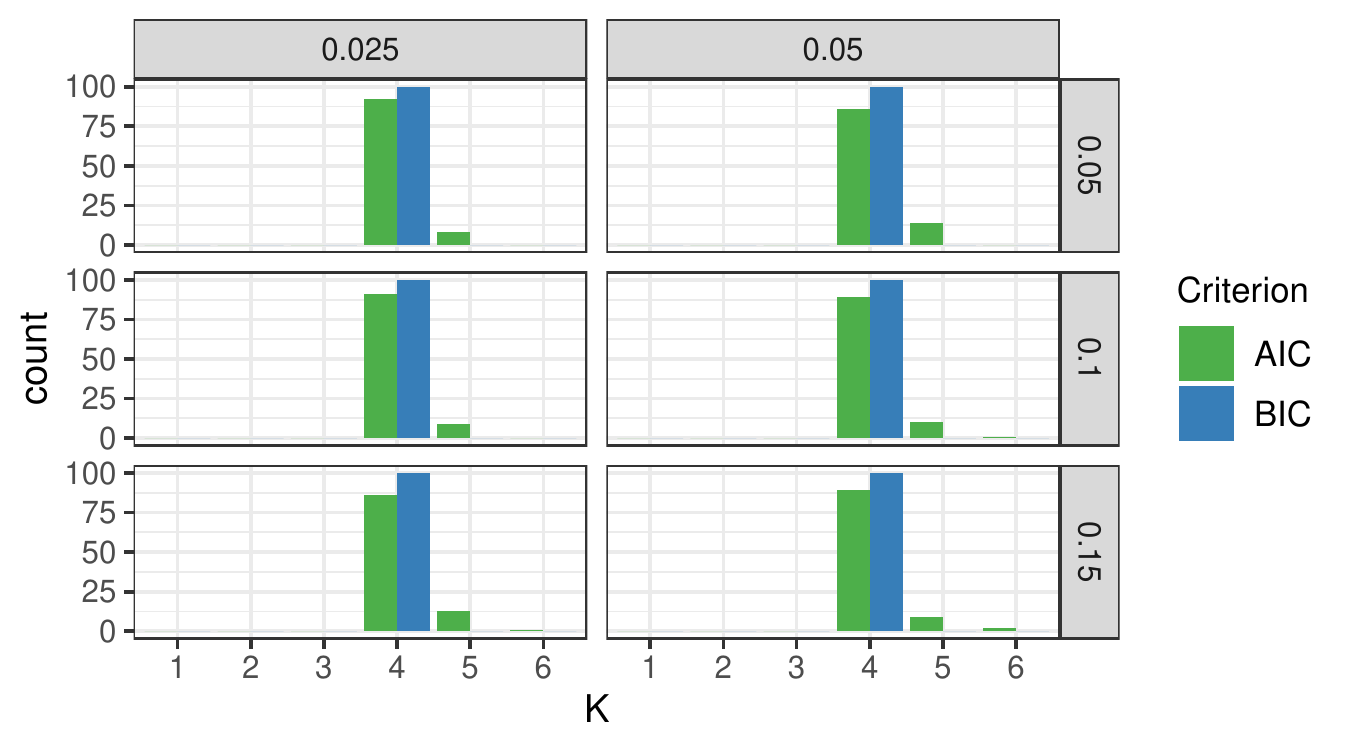}  
  \caption{\textbf{Dense case}: number of times each $K$ is selected as the best configuration by
    AIC or BIC for $\mathbf{N=1000}$ observations and $\beta=0$, across overlapping
    values (in column) and directional mean sparsity (in row).}
  \label{fig:mix:sparse:100:1000:dense}
\end{figure}

\begin{figure}[htbp]
  \centering
\includegraphics{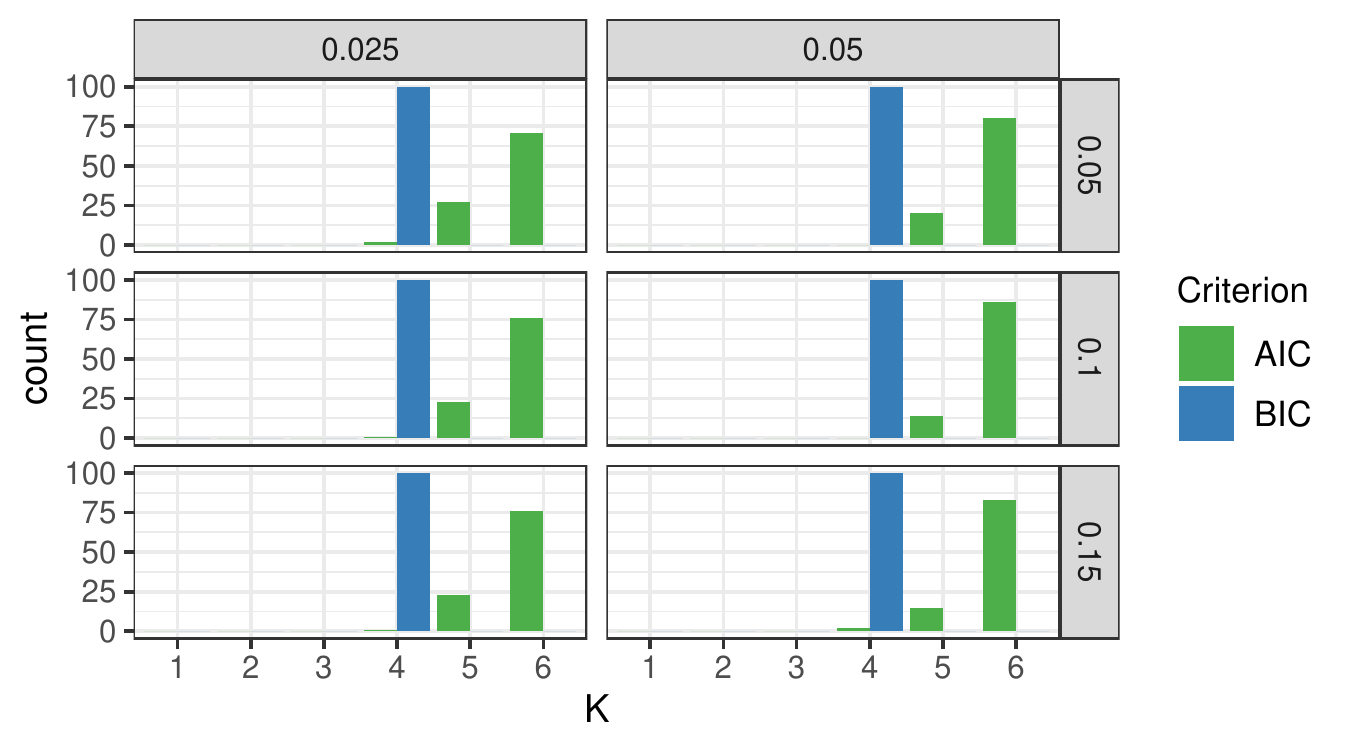}  
\caption{\textbf{Sparse case}: number of times each $K$ is selected as the best configuration by AIC
  or BIC for $\mathbf{N=1000}$ observations for the optimal $\beta$ selected by each
  criterion, across overlapping values (in column) and directional mean
  sparsity (in row).}
  \label{fig:mix:sparse:100:1000:sparse}
\end{figure}

Unfortunately, this overfitting behavior of AIC manifests even more in the
simpler case with $N=1000$ observations (see Figures \ref{fig:mix:sparse:100:1000:dense} and
\ref{fig:mix:sparse:100:1000:sparse}), while BIC on the contrary is able to
recover the true number of components, with and without sparsity enforcement. 

The simulation study tends to favor the BIC, but this is probably an effect of
the reasonable ratio between the dimension $d=100$ and the number of
observations $N=200$ and $N=1000$. Experiments in Section
\ref{sec:exper-real-world} and \ref{sec:exper-real-world:WFC} will show
examples of a less appropriate behavior of the BIC in more adverse setting,
when $d$ is large compared to $n$. This confirms previous results summarized
in Section \ref{sec:model-selection}, which tend to show that information
criteria can only be use to guide the exploration of the data for this type of
mixture models.

We have not included in this section the results obtained for other
information criteria recalled in Section \ref{sec:model-selection}. On
simulated data, they perform uniformly worse than the AIC and the BIC in the
small number of observations regime ($N=200$ for $d=100$) and
roughly identically to the BIC in the large number of observations case
($N=1000$). We investigate their practical relevance on real world data in Section
\ref{sec:exper-real-world} and \ref{sec:exper-real-world:WFC}. 

\subsubsection{Selection on the path}\label{sec:select-path-spars}
\begin{figure}[htbp]
  \centering
\includegraphics{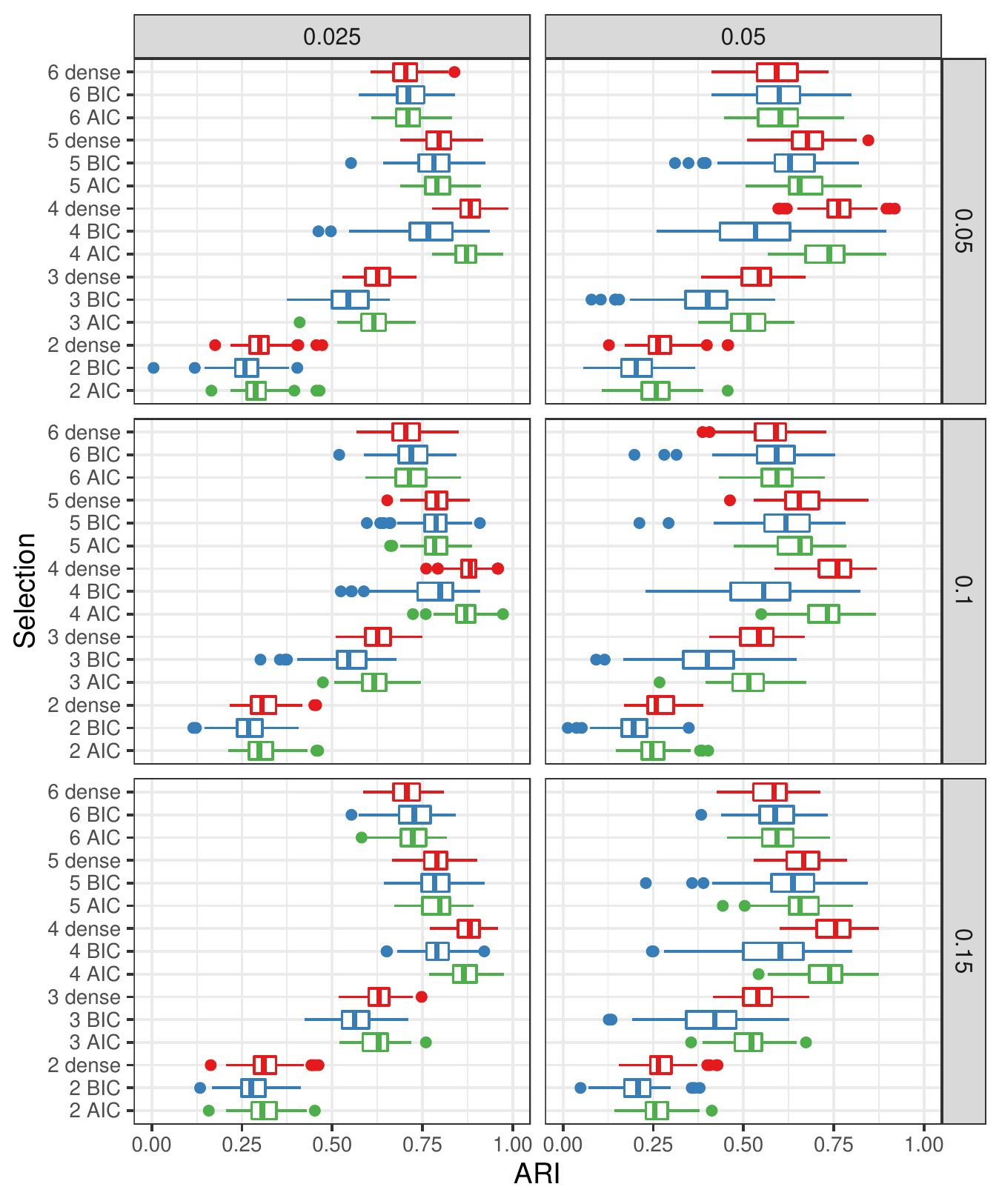}  
\caption{Adjusted rand index distribution for the optimal dense model (in
  red) and for the optimal sparse models according to the AIC (green) and BIC (blue), as a function
  of $K$, the number of components, for $\mathbf{N=200}$. Panels are
  organised based on overlapping (vertically) and on sparsity (horizontally).}
  \label{fig:path:sparse:ARIP1:200}
\end{figure}

We study now the effect of selecting the best $\beta$ with the BIC or the
AIC. We use as performance metric the adjusted rand index
(ARI) between the ground truth and the crisp assignments produced by the
different models. Figure \ref{fig:path:sparse:ARIP1:200} shows the results for
$N=200$ observations. In this case, the BIC tends to over sparsify the
directional means compared to the ARI, especially when the $K=4$, the true
number of components. 

\begin{figure}[htbp]
  \centering
\includegraphics{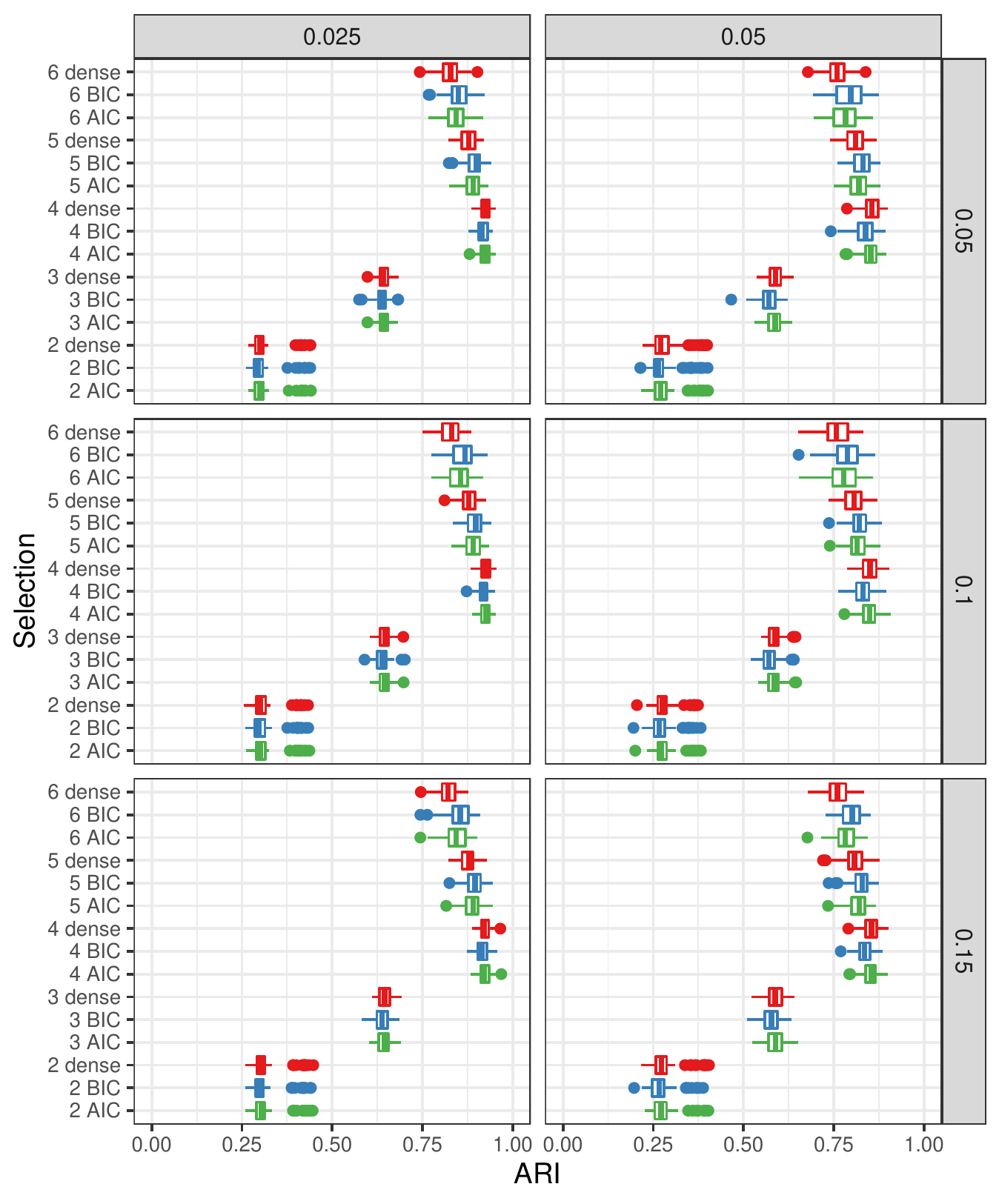}  
\caption{Adjusted rand index distribution for the optimal dense model (in
  red) and for the optimal sparse models according to the AIC (green) and BIC (blue), as a function
  of $K$, the number of components, for $\mathbf{N=1000}$. Panels are
  organised based on overlapping (vertically) and on sparsity (horizontally).}
  \label{fig:path:sparse:ARIP1:1000}
\end{figure}

The phenomenon is linked to the difficultly of the estimation, as shown on
Figure \ref{fig:path:sparse:ARIP1:1000} with $N=1000$ observations. When we
have more observations, when the true directional means are sparser or when
the components overlap less, the ARI drop between BIC and AIC is less
pronounced. 

\begin{figure}[htbp]
  \centering
\includegraphics{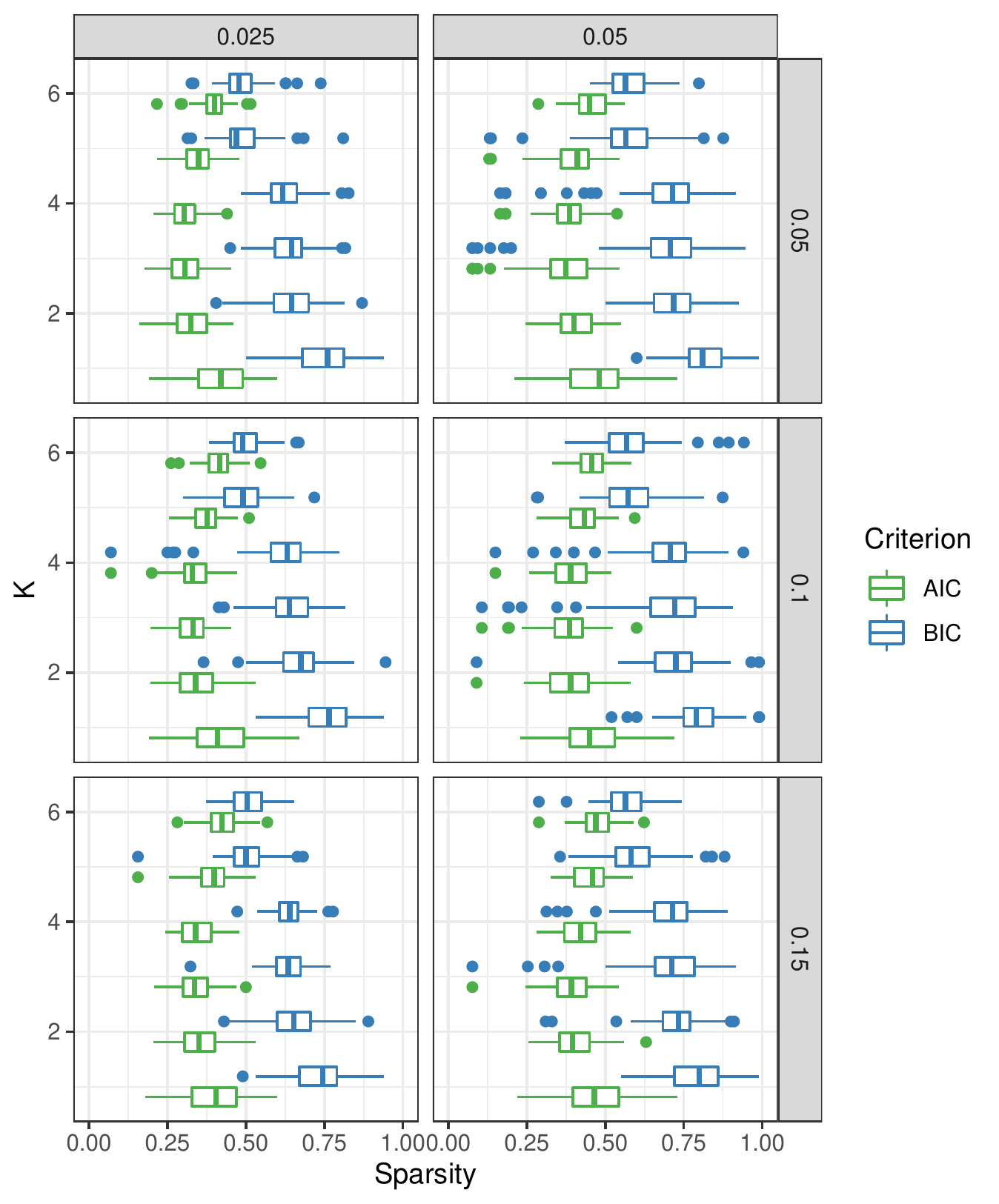}  
\caption{Sparsity achieved by the models selected by AIC and BIC, as a function
  of $K$, the number of components, for $\mathbf{N=200}$. Panels are
  organised based on overlapping (vertically) and on sparsity (horizontally).}
  \label{fig:path:sparse:sparsity:P1:200}
\end{figure}

\begin{figure}[htbp]
  \centering
\includegraphics{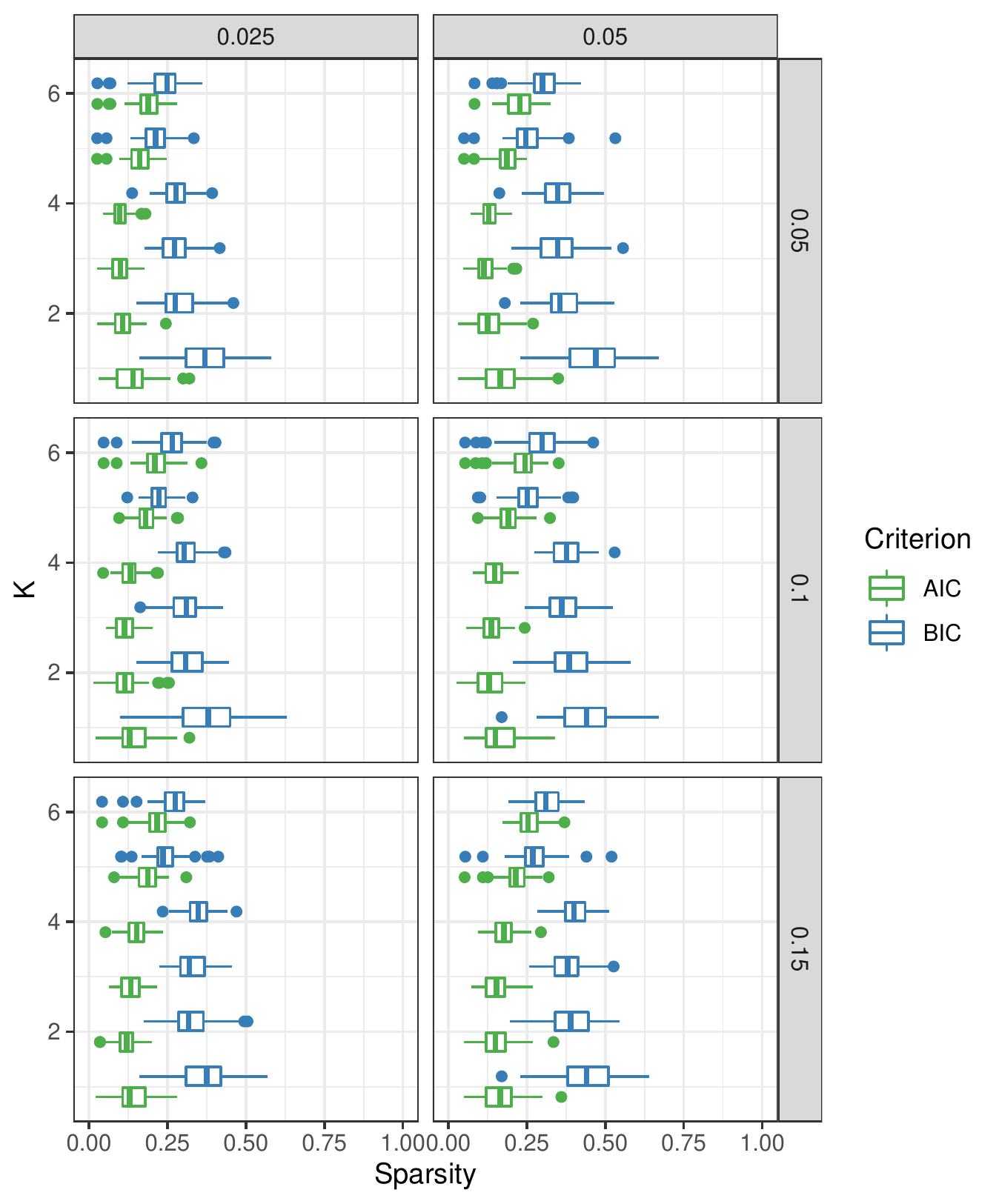}  
\caption{Sparsity achieved by the models selected by AIC and BIC, as a function
  of $K$, the number of components, for $\mathbf{N=1000}$. Panels are
  organised based on overlapping (vertically) and on sparsity (horizontally).}
  \label{fig:path:sparse:sparsity:P1:1000}
\end{figure}

It is also linked to the sparsity achievable given the number of observations,
as illustrated by Figures \ref{fig:path:sparse:sparsity:P1:200} and
\ref{fig:path:sparse:sparsity:P1:1000}. Indeed with more observations,
estimations of the directional mean components are tighter and the non zero
ones need a larger value of $\beta$ to be removed. The compromise between
sparsity and likelihood is more pronounced toward dense models. 

\subsubsection{Sparse directional mean recovery}\label{sec:sparse-direct-mean}

\begin{figure}[htbp]
  \centering
\includegraphics{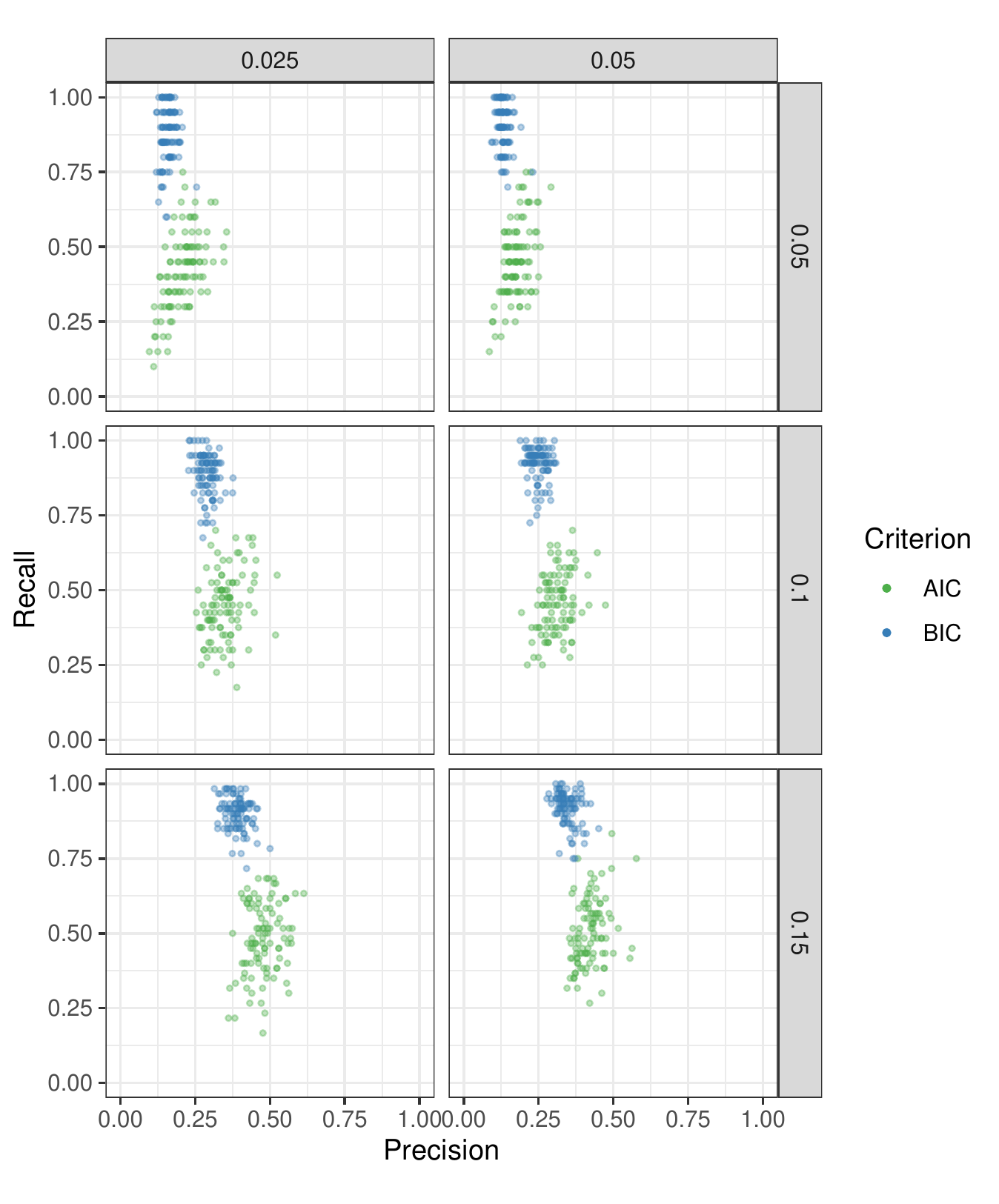}  
\caption{Precision and recall of the zero components of the direction means in
  the optimal sparse models according to the AIC and BIC for $K=K^*=4$ for
  $N=\mathbf{1000}$ and $d=100$. Panels are organised based on overlapping
  (vertically) and on sparsity (horizontally).}
  \label{fig:path:sparse:PR:1000}
\end{figure}

Finally, Figure \ref{fig:path:sparse:PR:1000} shows the precision and recall
of the optimal AIC and BIC models for 100 data sets with $N=1000$ and
$d=100$. They are measured by comparing the classification of the coordinates
of the directional means into two classes (zero and non zero components) with
the true classification induced by sparsifying the directional components
during the artificial data generation (notice that this makes sense only when
$K=K^*$). The low value of the precision confirms the tendency of both
criteria to select too sparse representations. On a sufficiently large data
set, the BIC as a significantly better recall than the AIC, but with
significant loss in precision. Results for smaller data sets tend to be worse
in precision and roughly equivalent in recall.

\subsection{Conclusion}
In summary, the path following strategy is an efficient way of exploring the
sparsification of the solutions. In the low number of observations regime,
the use of regularisation enables to select an optimal number of
components using the BIC. However in this regime, it also tends to select too
sparse directional means compared to the true parameters. Due to the large
number of parameters and the high dimension of the data under consideration,
this is not surprising, but the experiments show that care should be exercised
when using this type of model (regularized or not). The information criteria
offer only some general hints for the selection of the best models. In an
data exploration point of view, this means that one should consider a
collection of models obtained by applying the proposed procedure with
different choice of information criterion. The sparsity of directional means
is also to be consider with caution.

\section{Comparison with reference models}\label{sec:exper-real-world}
In this section, we compare our model to two reference models designed for
directional data, the spherical k-means algorithm \cite{1183895} and a model
based co-clustering algorithm, dbmovMFs, proposed in \cite{pmlr-v51-salah16}.

We describe briefly the reference models in Section
\ref{sec:reference-models}. Section \ref{sec:simulated-data} compares the
models on the artificial data introduced in Section
\ref{sec:simulation-study}, while Section \ref{sec:comp-science-techn}
compares them on the popular benchmark CSTR.

\subsection{Reference models}\label{sec:reference-models}
\subsubsection{Spherical k-means (Sk-means)}
The spherical k-means algorithm (Sk-means), originally proposed in
\cite{1183895}, is a simple adaptation of the k-means algorithm to the cosine
dissimilarity. Let us a consider a collection of $N$ observations
$\boldsymbol{X}=(\boldsymbol{x}_i)_{1\leq i\leq N}$ on the hypersphere
$\mathbb{S}^{d-1}$. Given a number of clusters $K$, Sk-means tries to find
a set of $K$ prototypes $(\boldsymbol{\mu}_k)_{1\leq k\leq K}$ in
$\mathbb{S}^{d-1}$ and a clustering/membership $\boldsymbol{Z}=(z_i)_{1\leq i\leq N}$, that assigns $\boldsymbol{x}_i$ to  cluster
$z_i\in\{1,\ldots, K\}$ such that the coherence
\begin{equation}
  \label{eq:sk}
 \mathcal{Q}( (\boldsymbol{\mu}_k)_{1\leq k\leq K}, (z_i)_{1\leq i\leq
   N})=\sum_{i=1}^N \boldsymbol{\mu}_{k_i}^T\boldsymbol{x}_i
\end{equation}
is maximal.

Several methods have been proposed to maximize the coherence (see
e.g. \cite{JSSv050i10}). The original method proposed in \cite{1183895} is
Lloyd-Forgy style fixed-point algorithm which iterates between determining
optimal memberships for fixed prototypes, and computing optimal prototypes for
fixed memberships. In particular the prototypes are the normalized average of
the points assigned to their cluster. We used this method in the following
experiences (as implemented in the R package \texttt{skmeans}
\cite{JSSv050i10}). Apart from the number of clusters $K$, the spherical
k-means algorithm has no meta-parameter.

\subsubsection{Diagonal Block vMF mixture model (dbmovMFs)}
The diagonal block vMF mixture model (dbmovMFs) was proposed in
\cite{pmlr-v51-salah16}. It can be seen as a constrained version of the
classical mixture of vMF distribution. The key idea is to enforce on the
directional means a block structure that mimics the one used in co-clustering
algorithms. Technically, this is done by introducing a crisp clustering on the
dimensions/columns, represented by a crisp assignment matrix
$\boldsymbol{W}=(w_{jk})_{1\leq j\leq d, 1\leq k\leq K}$, where $w_{jk}=1$ if
dimension $j$ is assigned to cluster $k$ and 0 if not (notice that there
are as many column clusters as there are components in the mixture). 

The directional means are strongly constrained to a diagonal structure, that
is
\begin{equation}
\mu_{kj}=w_{jk}\mu_k,
\end{equation}
where $\mu_k$ is real number. Thus $\boldsymbol{\mu}_k$ has a zero coordinate
on all the dimensions that are not assigned to dimension cluster $k$, and a
fixed value $\mu_k$ on dimensions that are in this cluster. As a consequence,
the complete data likelihood as the following form
\begin{equation}
\prod_{i=1}^N\prod_{k=1}^K \left(\alpha_k c_d(\kappa_k)\times \prod_{j=1}^d(\exp^{\kappa_k\mu_k\boldsymbol{x}_{ij}})^{w_{jk}} \right)^{z_{ik}}.
\label{completelikelihooddbmovMFs}
\end{equation}
This complete data likelihood is used as the basis of a EM algorithm described
in \cite{pmlr-v51-salah16}. The algorithm has some common aspect to the one
proposed in \cite{10.5555/1046920.1088718} but also include a specific phase
of column cluster update. We use the authors
implementation\footnote{\url{https://github.com/dbmovMFs/DirecCoclus/}}. Notice
that the authors proposed several variants of the EM algorithm, but also
showed in \cite{pmlr-v51-salah16} that the best results are obtained by the
classical EM. Therefore we use it in all our experiments. Apart from the
number of components $K$, dbmovMFs has no meta-parameter. 

\subsection{Simulated data}\label{sec:simulated-data}
We compare in this section our model to Sk-means and dbmovMFs on the simulated
data used in Section \ref{sec:simulation-study}. For each configuration (data
size, sparsity and separation), we run Sk-means and dbmovMFs in a similar way
as we applied our model: both algorithms are initialized randomly 10 times and
the best model is kept according to its specific quality metric (largest
coherence for the Sk-means and largest likelihood for dbmovMFs). The random
initialisation is similar to the one described in algorithm
\ref{code:EM:init} (random directional means selected from the data set
followed by an initial crisp clustering).

\begin{figure}[htbp]
  \centering
\includegraphics{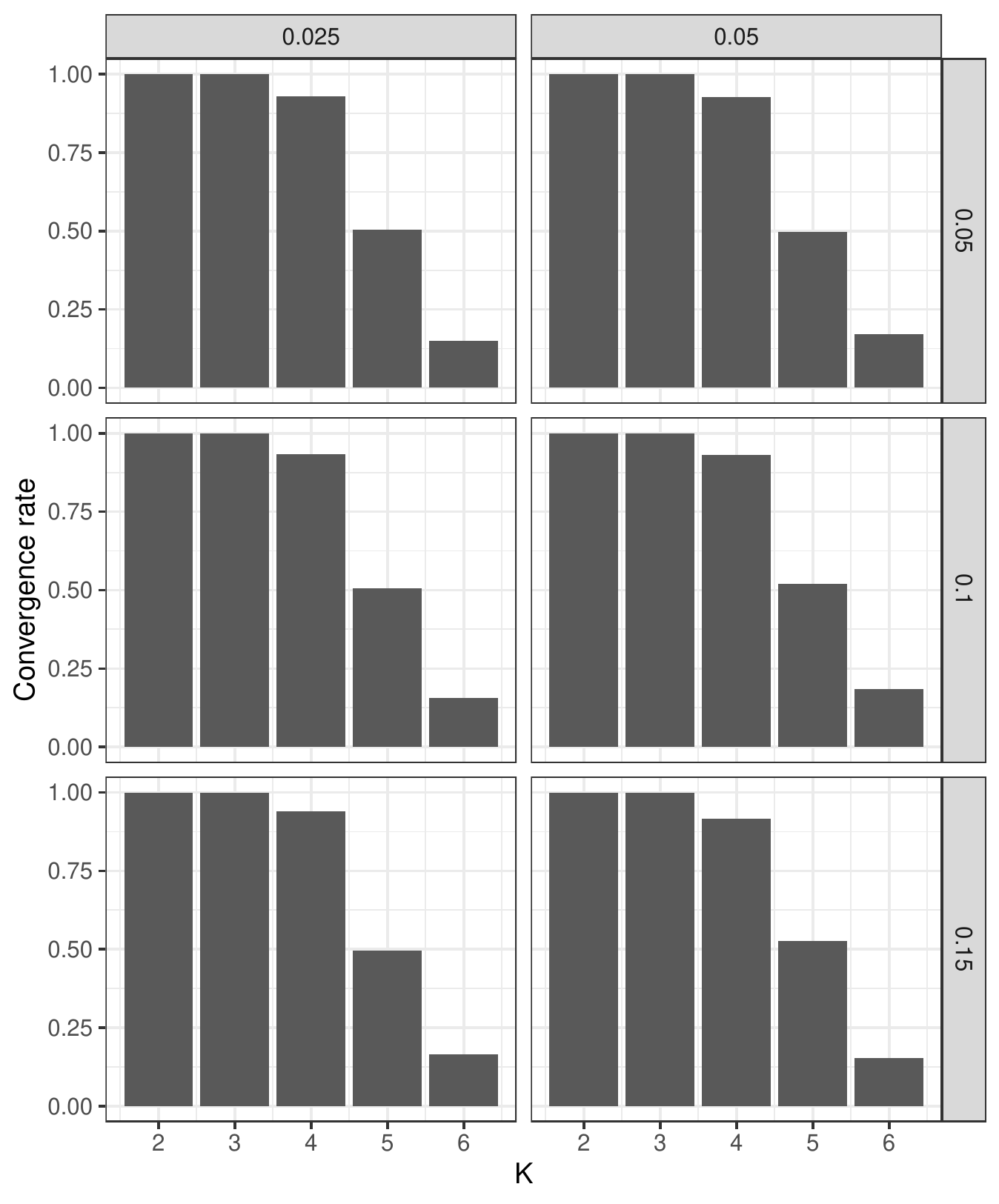}
\caption{Convergence rates for dbmovMFs for $N=\mathbf{200}$ and
  $d=100$. Panels are organised based on overlapping (vertically) and on
  sparsity (horizontally).}
  \label{fig:simul:dbmovfs:rate}
\end{figure}

DbmovMFs performs extremely poorly on the simulated data, mainly because the
sparsity constraints associated to the diagonal block structure are
too restrictive. In fact, the EM algorithm fails to converge for a significant
part of the initialisation, especially for higher values of $K$: some of the
components of the mixture become empty. Notice that this never happens for
Sk-means or for our algorithm. Figure \ref{fig:simul:dbmovfs:rate} illustrates
the phenomenon by displaying for each $K$ and each setting, the ratio between
the number of converging runs of dbmovMFs and the total of attempted runs. The
results are reported for $N=200$ observations but they are even worse for
$N=1000$. 

\begin{figure}[htbp]
  \centering
\includegraphics{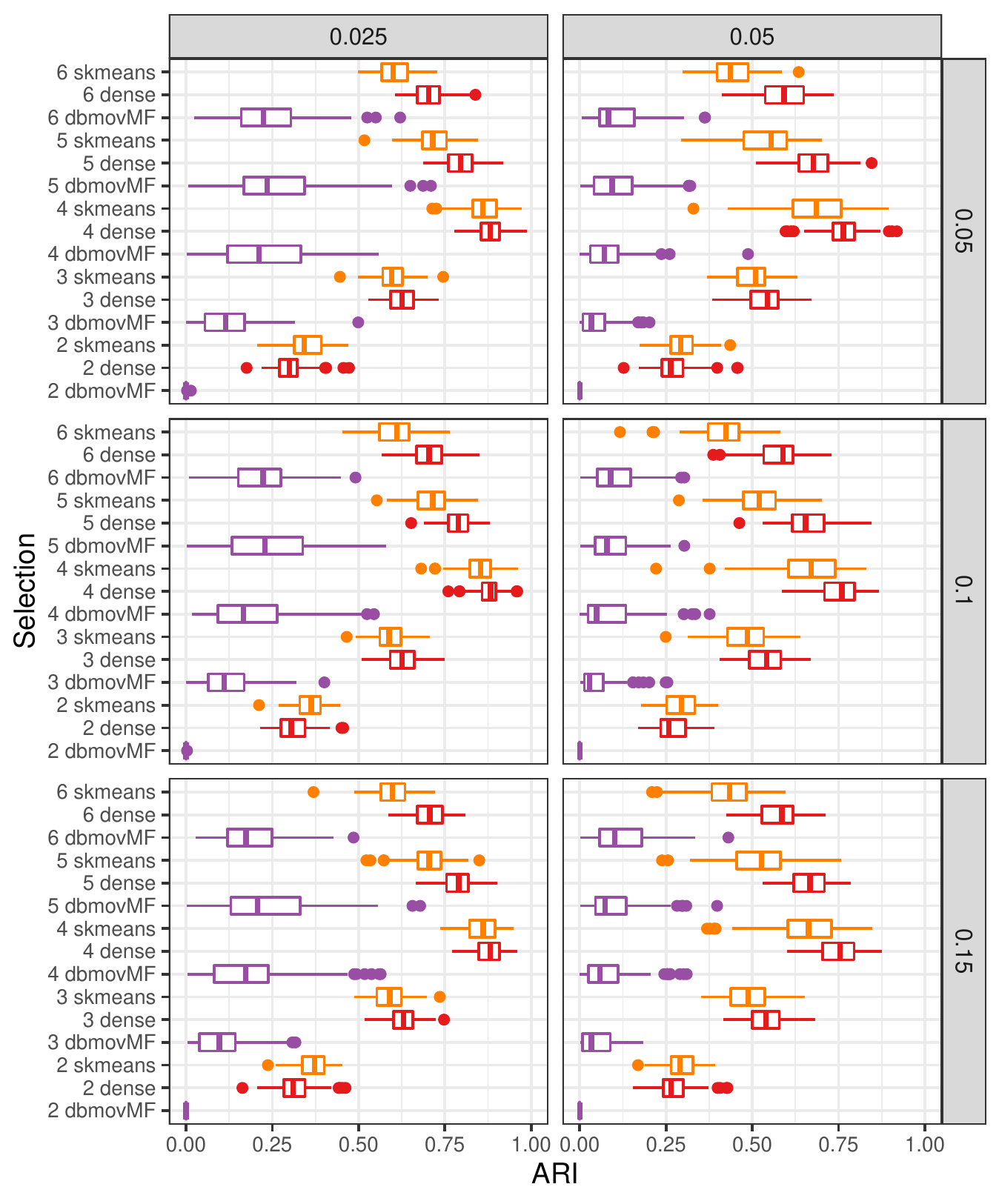}
\caption{Adjusted rand index distribution for the optimal dense model (in
  red), for the spherical k-means algorithm (in orange) and for dbmovMFs (in
  purple), as a function of $K$, the number of components, for
  $\mathbf{N=200}$. Panels are organised based on overlapping (vertically) and
  on sparsity (horizontally).}
  \label{fig:path:sparse:ARI:reference:200}
\end{figure}

\begin{figure}[htbp]
  \centering
\includegraphics{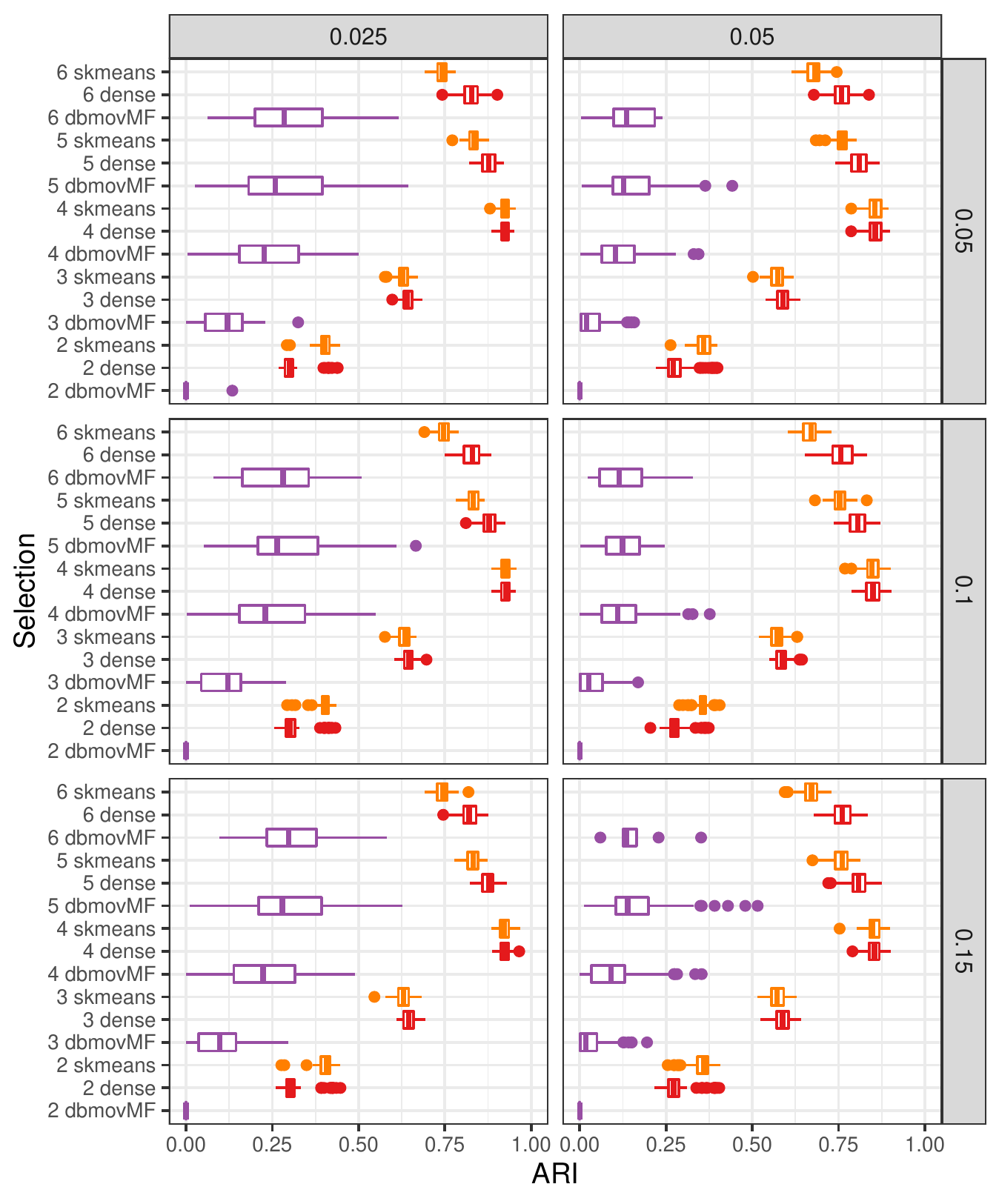}
\caption{Adjusted rand index distribution for the optimal dense model (in
  red), for the spherical k-means algorithm (in orange) and for dbmovMFs (in
  purple), as a function of $K$, the number of components, for
  $\mathbf{N=1000}$. Panels are organised based on overlapping (vertically) and
  on sparsity (horizontally).}
  \label{fig:path:sparse:ARI:reference:1000}
\end{figure}

In terms of recovering the ground truth as measured by the ARI, both Sk-means
and dbmovMFs performances are generally below than the dense solution obtained
by our methodology, as shown on Figures
\ref{fig:path:sparse:ARI:reference:200} and
\ref{fig:path:sparse:ARI:reference:1000}. DbmovMFs performs extremely poorly
and is unable to recover the planted structure. Spherical k-means results are
identical to those of our approach for $N=1000$ and $K=4$. In other
configurations (a smaller data set or a mispecification of the number of
clusters) that are always inferior, excepted in the particular case of $K=2$. 

\begin{figure}[htbp]
  \centering
\includegraphics{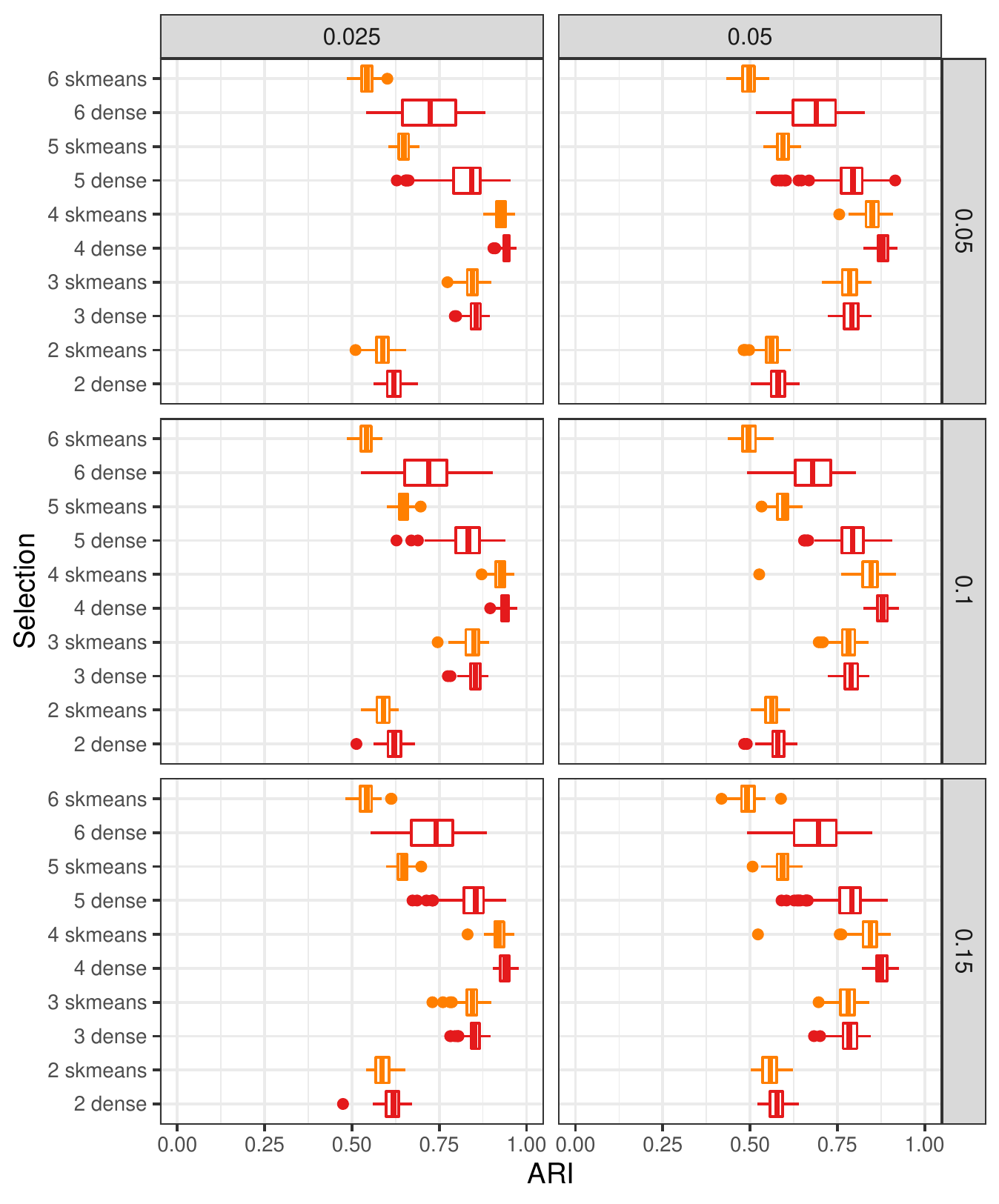}
\caption{Unbalanced clusters: adjusted rand index distribution for the optimal dense model (in
  red) and for the spherical k-means algorithm (in orange), as a function of $K$, the number of components, for
  $\mathbf{N=1000}$. Panels are organised based on overlapping (vertically) and
  on sparsity (horizontally).}
  \label{fig:path:sparse:ARI:reference:unbalanced:1000}
\end{figure}

Notice that the setting is very favorable for Sk-means as the clusters are
balanced and use quite similar concentration values $\kappa_k$. As pointed out
in \cite{pmlr-v51-salah16}, the performances of Sk-means tend to deteriorate
when the true clusters are unbalanced. We have confirmed this behavior by
generating another collection of artificial data exactly as in Section
\ref{sec:simulation-study} but with
\begin{equation*}
  \boldsymbol{\alpha}=\left(\frac{1}{2},
  \frac{1}{4}, \frac{1}{8}, \frac{1}{8}\right).
\end{equation*}
Results are provided in Figure
\ref{fig:path:sparse:ARI:reference:unbalanced:1000} the case of $N=1000$
observations. The proposed model recovers the true clustering uniformly better
than the Sk-means (the results for $N=200$, omitted, show a larger separation
between the methods). 

In conclusion, experiments on artificial data show, as expected, that the
patterns generated by a mixture of vMF distributions are difficult to recover
for Sk-means and nearly impossible to recover for dbmovMFs. The spherical
k-means works reasonably well when the data set contains enough observations
and when the clusters are balanced, but is outperformed by our approach most
of the time. The block structure imposed by dbmovMFs is too strong for it to
deal with data with limited sparsity.

\subsection{Computer Science Technical Reports (CSTR)}\label{sec:comp-science-techn}
The CSTR data set, proposed in
\cite{10.1145/1081870.1081894}\footnote{Available for instance here at this
  URL \url{https://github.com/dbmovMFs/DirecCoclus/tree/master/Data}}, is a
good example of a rather high dimensional but small data set with $N=475$
examples in dimension $d=1000$. It has been produced from a selection of 475
abstracts of technical reports\footnote{Reports can be downloaded from the
  department web site
  \url{https://www.cs.rochester.edu/research/technical_reports.html}}
published by the Department of Computer Science 
at the University of Rochester between 1991 and 2002. The reports are
represented on an undisclosed dictionary of 1000 words, with a binary encoding
(a word is present or not in an abstract). Based on the research areas
developed by the CS department at the time of the collection, the abstracts
are grouped in $K=4$ classes (Natural Language Processing, Robotics/Vision,
Systems, and Theory).

We use this real world data set to compare our approach to Sk-means and
dbmovMFs. 

\subsubsection{Experimental protocol}
While CSTR has been used frequently as a benchmark, some care must be
exercised in doing so. Indeed the classes of the CSTR data set are not
clusters as shown by a simple experiment: using as the initial partition the
true classes, an application of the standard spherical k-means algorithm
\cite{JSSv050i10} leads to a different partition after convergence. The
adjusted rand index (ARI) between the two partitions is of $0.835$. As shown
on the confusion matrix between the two partitions (see Table
\ref{table:cstr:confusion}), two of the classes are somewhat difficult to
recover from a clustering point of view.

\begin{table}[htbp]
  \centering
\begin{tabular}{rrrrr}
  \toprule
 & 1 & 2 & 3 & 4 \\ 
  \midrule
1 &  71 &  26 &   3 &   1 \\ 
  2 &   0 &  70 &   1 &   0 \\ 
  3 &   0 &   1 & 176 &   1 \\ 
  4 &   0 &   2 &   5 & 118 \\ 
   \bottomrule
\end{tabular}
  \caption{Confusion matrix between the classes of the CSTR data set (in row)
    and the classes obtained by the spherical k-means (in column).}\label{table:cstr:confusion}
\end{table}

The behavior of the mixture of vMF distributions on CSTR is similar to the one
of the spherical k-means. Using the same initialisation, we obtain after
convergence an ARI of $0.818$ with component specific $\kappa$s and of $0.837$ with a common
$\kappa$. The confusions matrices (omitted) are almost identical to the
spherical k-means one. As a consequence, an ARI around $0.84$ should be
considered as the maximum a method can reach on this data set. Higher results
could be only a matter of chance or obtained with a notion of cluster that is
more aligned with the ground truth classes.

Another difficulty is the small size of the data set compared to its number of
features. This increases the variance of the estimates provided by any
algorithm and as a consequence, the final clustering obtained by different
methods from a random initialization tend to be much more dependent on this
initial configuration than in the case of a simpler data set (as e.g. in the
artificial data experiments reported above). To provide meaningful results, we
proceed as follows. For each algorithm, we use a common set of 50 random
initial configurations (obtained with algorithm \ref{code:EM:init}): this
ensure that the algorithms are used under exactly the same testing
conditions. After convergence of a given algorithm, we keep the best
configuration in terms of the quality criterion of this algorithm (e.g. the
likelihood for mixture models) and report the ARI of the corresponding
clustering. We repeat this procedure 50 times (thus considering 250 random
initial configurations) to assess the variability of the results.

\subsubsection{Results for the dense models: ARI}
\begin{figure}[htbp]
  \centering
\includegraphics{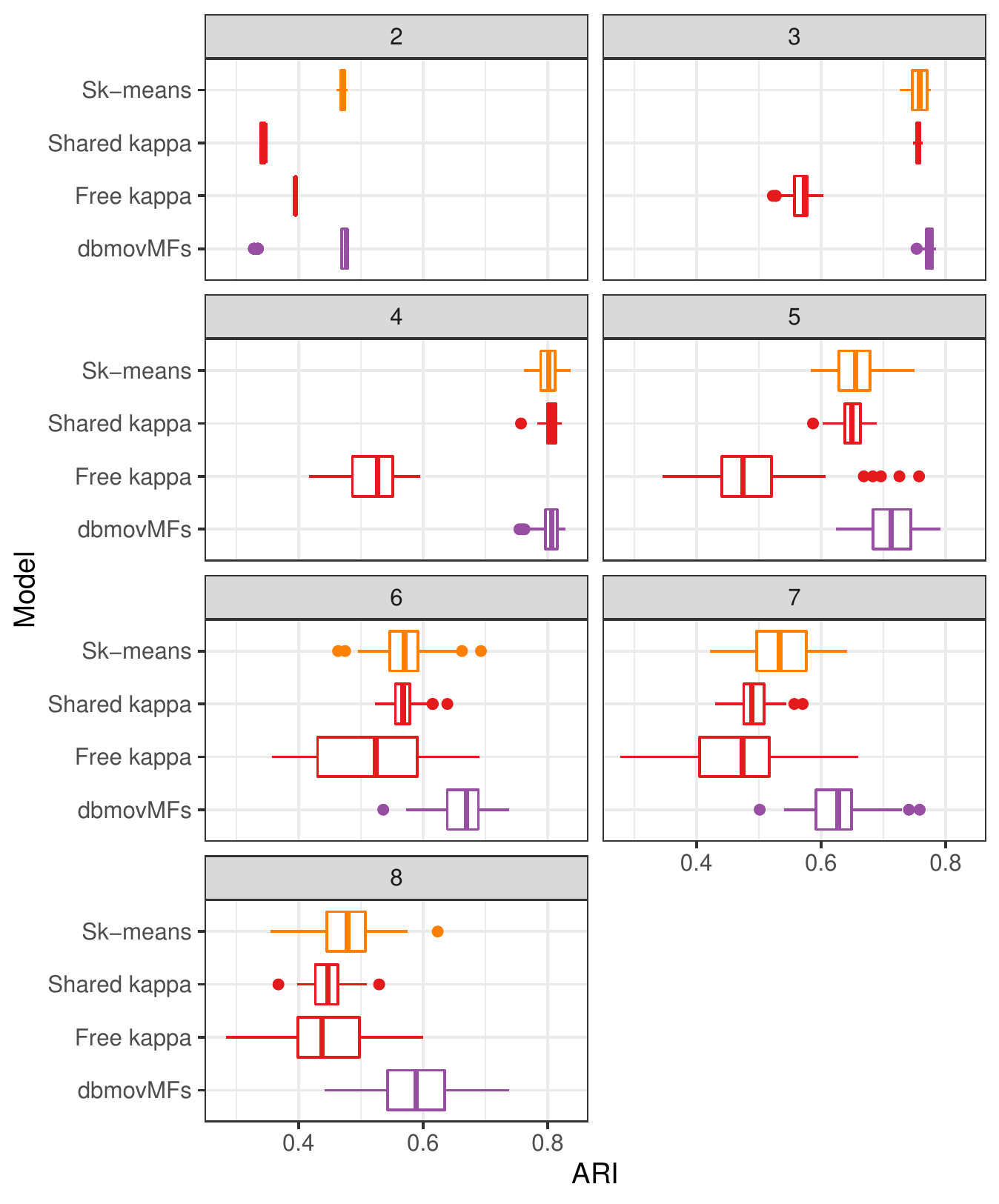}   
  \caption{Adjusted Rand Index between the CSTR classes and the clusters
    obtained by Sk-means, mixture of
    vMF distributions with a common $\kappa$ parameter (shared kappa) and
    mixture of vMF distributions with component specific $\kappa$s (free
    kappa), and dbmovMFs for different values of $K$.}
  \label{fig:cstr:dense}
\end{figure}

Figure \ref{fig:cstr:dense} and Table \ref{tab:cstr:dense:stats} summarize the results obtained by the spherical
k-means, dbmovMFs and the two dense variants of the mixtures of vMF. The mixture with
component specific concentration parameters has by far the largest variability
and the worst results. The adverse effects of a too high value for the
concentration parameter on real world data was already established in
e.g. \cite{hornik2014movmf, pmlr-v51-salah16}. As far as we know, the very
strong sensitivity of the results to the initial configuration is a new result
(as far as we know). Both issues are solved by using a shared concentration parameter. The
variability of the results is then smaller than the one observed for the
spherical k-means and on the optimal configuration with $K=4$, the results are
roughly identical. In particular, a paired t-test does not show significant
differences at a 1\% level between the spherical k-means and the shared
$\kappa$ mixture of vMF for $K\in\{3, 4, 5, 6\}$. 

\begin{table}[htbp]
  \begin{center}
    \footnotesize
\input{results/cstr_dense_stats.tex}    
  \end{center}
  \caption{Adjusted Rand Index between the CSTR classes and the clusters
    obtained by the models under study.}
  \label{tab:cstr:dense:stats}
\end{table}

As shown in \cite{pmlr-v51-salah16}, the block structure enforced by dbmovMFs is also an
efficient way of controlling the adverse effects of the concentration
parameters. While the results for $K=4$ are identical to the ones obtained by
other methods, dbmovMFs is far more robust to a misspecification
of the number of components. Apart for $K=2$ where the spherical k-means
provide the best ARI (significant difference at a 1\% level), in all other
configurations with $K\neq 4$, the ARI obtained by the dbmovMFs is
significantly larger than the ones obtained by other methods.

DbmovMFs appears therefore to provide a more robust solution
than dense models such as classical mixtures of vMF distributions and than spherical k-means,
thanks to its good behavior under mispecification of the number of
clusters. Notice however that it had  extremely poor results on denser data,
as shown on the simulated data. 

\subsubsection{Results for the dense models: model selection}
\begin{figure}[htbp]
  \centering
\includegraphics{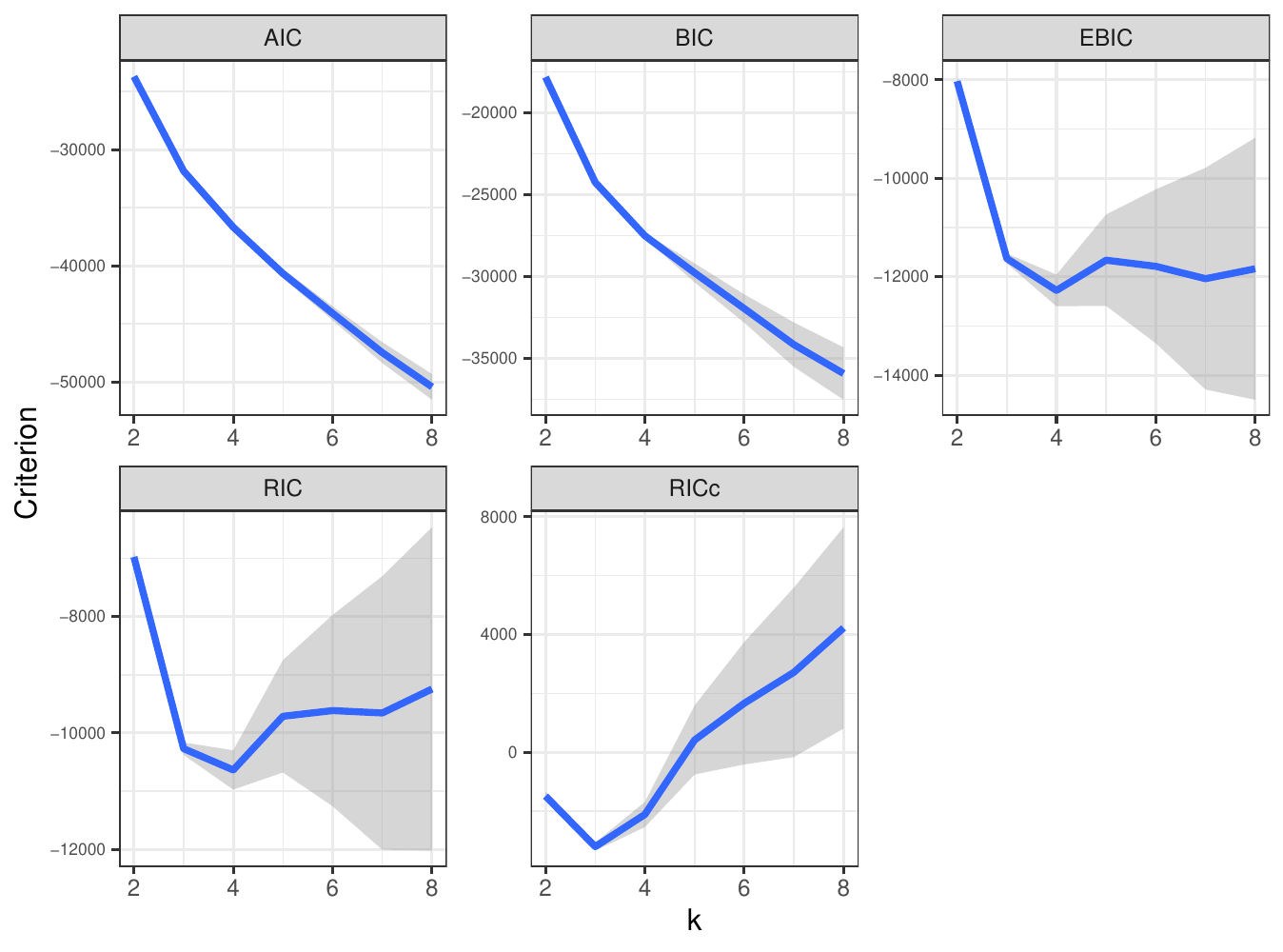}   
  \caption{Model selection criteria for the mixture of
    vMF distributions with a common $\kappa$ parameter: the blue curve is the
    mean value, while the grey envelop displays a 2 standard deviation tube
    around it.}
  \label{fig:cstr:dense:movmf:ic}
\end{figure}

\begin{figure}[htbp]
  \centering
\includegraphics{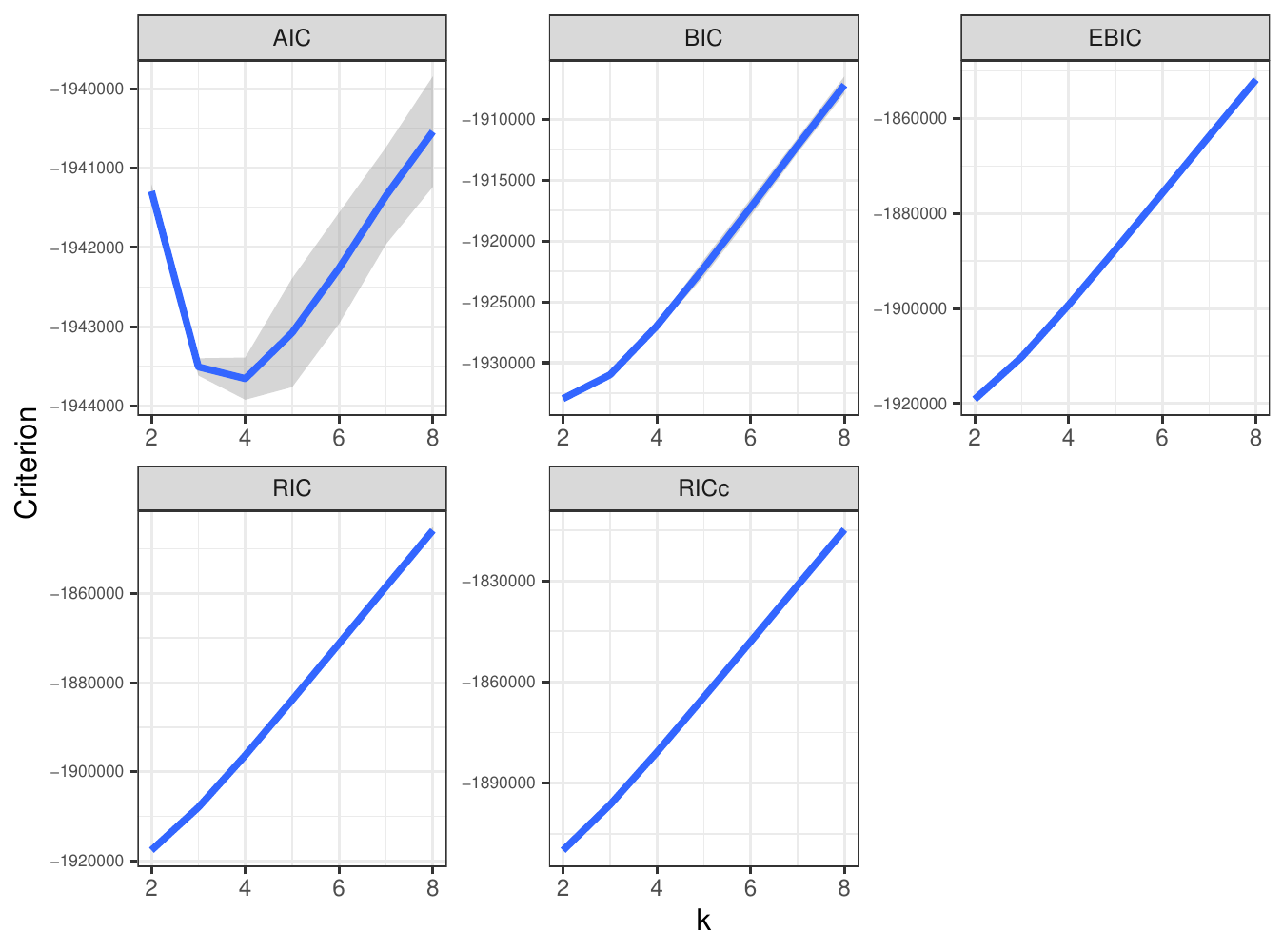}   
  \caption{Model selection criteria for dbmovMFs: the blue curve is the
    mean value, while the grey envelop displays a 2 standard deviation tube
    around it.}
  \label{fig:cstr:dense:db:ic}
\end{figure}

Figures \ref{fig:cstr:dense:movmf:ic} and \ref{fig:cstr:dense:db:ic} display
the behavior of the model selection criteria for the shared $\kappa$ mixture
of vMF and for dbmovMFs. They show quite different
behaviors. For the vMF distribution strongly penalized criteria should be used
to recover the best models, while on the contrary, the small number of
parameters of the
co-clustering approach leads to a better behavior of the AIC. 

Those quite different behaviors do not give a major advantage of one algorithm
over the other on the CSTR data set. We will see in Section
\ref{sec:exper-real-world:WFC} that dbmovMFs is probably overpenalized even by
the AIC for more complex data sets and that vMF mixtures are probably
underpenalized even by e.g. the RIC. This confirms the limitations of
information criterion for this type of unsupervised high dimensional
models. As a consequence we argue that they should be used to guide the
exploration rather than as a proof of existence of a specific number of
clusters. 

\subsubsection{Sparse models}
\begin{figure}[htbp]
  \centering
\includegraphics{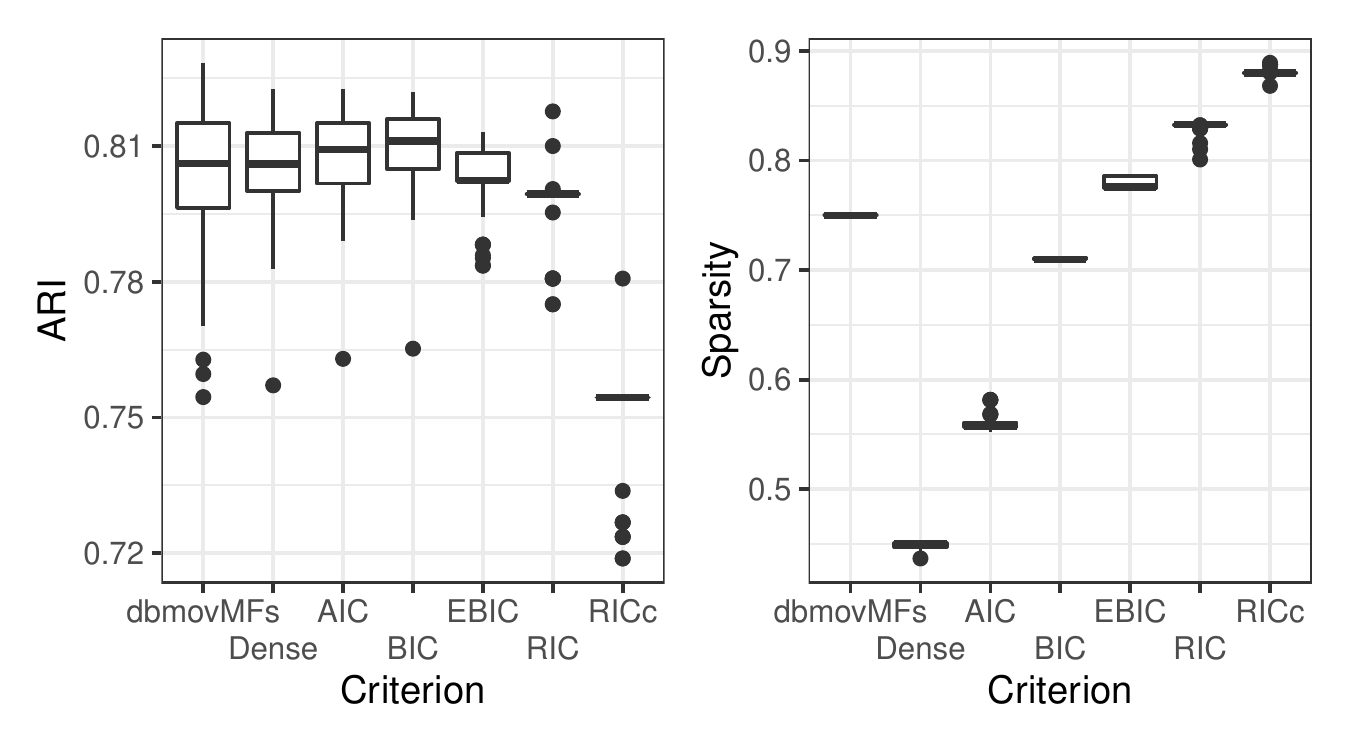}   
  \caption{Adjusted rand index and sparsity for the models selected on the
    $\beta$ path using the different model complexity criteria. The ``dense''
    configuration corresponds to the solution obtained without
    regularization. DbmovMFs results are given for reference.}
  \label{fig:cstr:sparse:ari:sp}
\end{figure}

On a second step, we compute the $\beta$ path for each of the 50 replications of our
procedure, starting each time from the best initialization obtained from the
50 random initial configurations. We restrict ourselves to the shared $\kappa$
model. Figure \ref{fig:cstr:sparse:ari:sp} and Table \ref{tab:cstr:sparse:stats} summarize the results. In terms of
sparse model selection, AIC, BIC and EBIC provide good compromises between the
ARI and the sparsity. Both RIC and RICs select a too sparse model. 

\begin{table}[htbp]
  \begin{center}
    \footnotesize
\input{results/cstr_sparse_stats.tex}    
  \end{center}
  \caption{Adjusted Rand Index between the CSTR classes and the clusters
    obtained by the sparse models under study.}
  \label{tab:cstr:sparse:stats}
\end{table}

A very important point is that none of the criteria is able to provide an
all-in-one selection. Indeed, as shown on Figure
\ref{fig:cstr:dense:movmf:ic}, the number of components should be selected
with EBIC or RIC (and possibly with RICc), as both AIC and BIC are
monotonically decreasing with the number of components. However, if we compute the
$\beta$ path for different number of components and keep as the selected model
the ones that minimize each criteria, this behavior applies to all
criteria. In other words, the regularization is compensating for the increased
number of components. Thus one should first select the number of components
based on EBIC or RIC, and then select the sparsity level with BIC or EBIC,
keeping the number of components fixed. Both sparse models selected by AIC and
BIC are significantly better than the dense model (according to a paired
t-test at a 1\% level). The BIC results are only significantly better than the
dbmovMFs results at a 10\% level.

In summary, using the proposed approach allows to reach similar performances
as dbmovMFs without enforcing a specific sparsity structure. On the contrary,
the sparsity is learned from the data without performance loss. 

\subsubsection{Data exploration}
We use in this section the visualisation method described in Section
\ref{sec:exploratory-use} in order to display the sparsity structure
discovered by the proposed method (we
restrict the illustration to $K=4$). 

\begin{figure}[htbp]
  \centering
\includegraphics{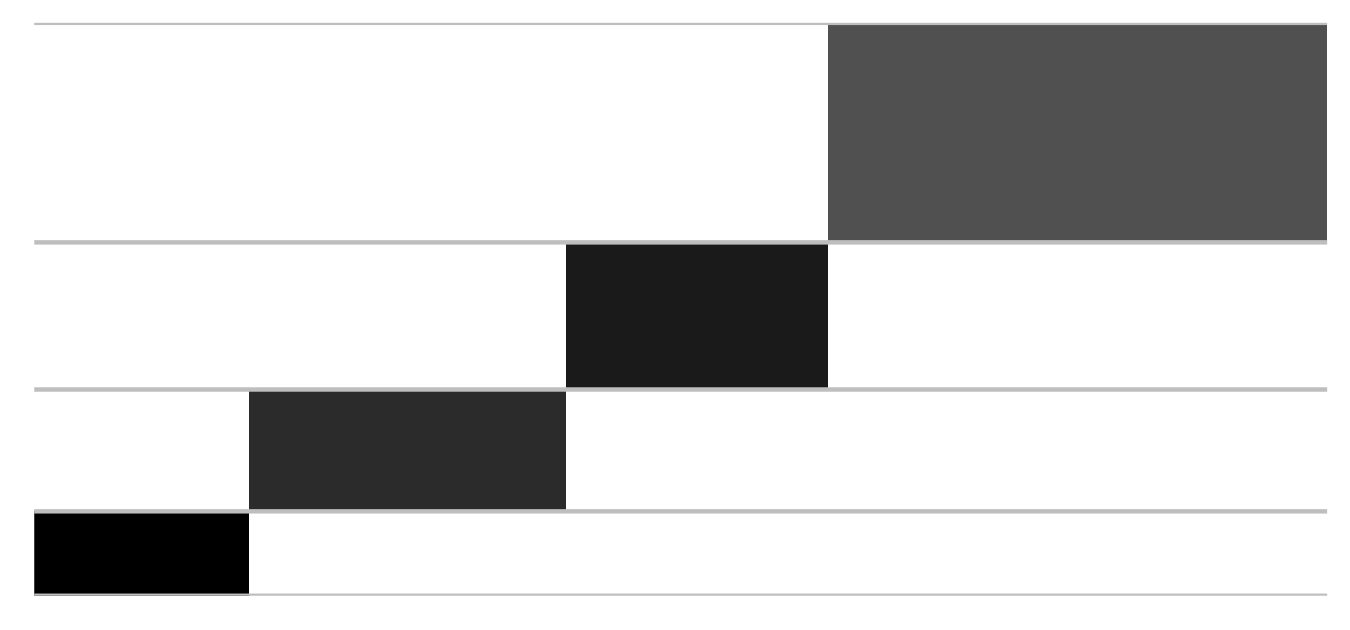}
\caption{Representation of the directional means obtained by dbmovMFs on the CSTR data set.}
  \label{fig:cstr:sparse:proto:db}
\end{figure}

Figure \ref{fig:cstr:sparse:proto:db} represents the block structure obtained
by dbmovMFs algorithm of \cite{pmlr-v51-salah16}. As expected, this
is a very crude model that does favor sparsity over revealing shared coordinates and
finer structure. For the point of view of dbmovMFs, the reports are described
by a collection of specific vocabulary with for instance the largest cluster
(top row) using the largest ``private'' vocabulary (top right
rectangle). 

\begin{figure}[htbp]
  \centering
\includegraphics{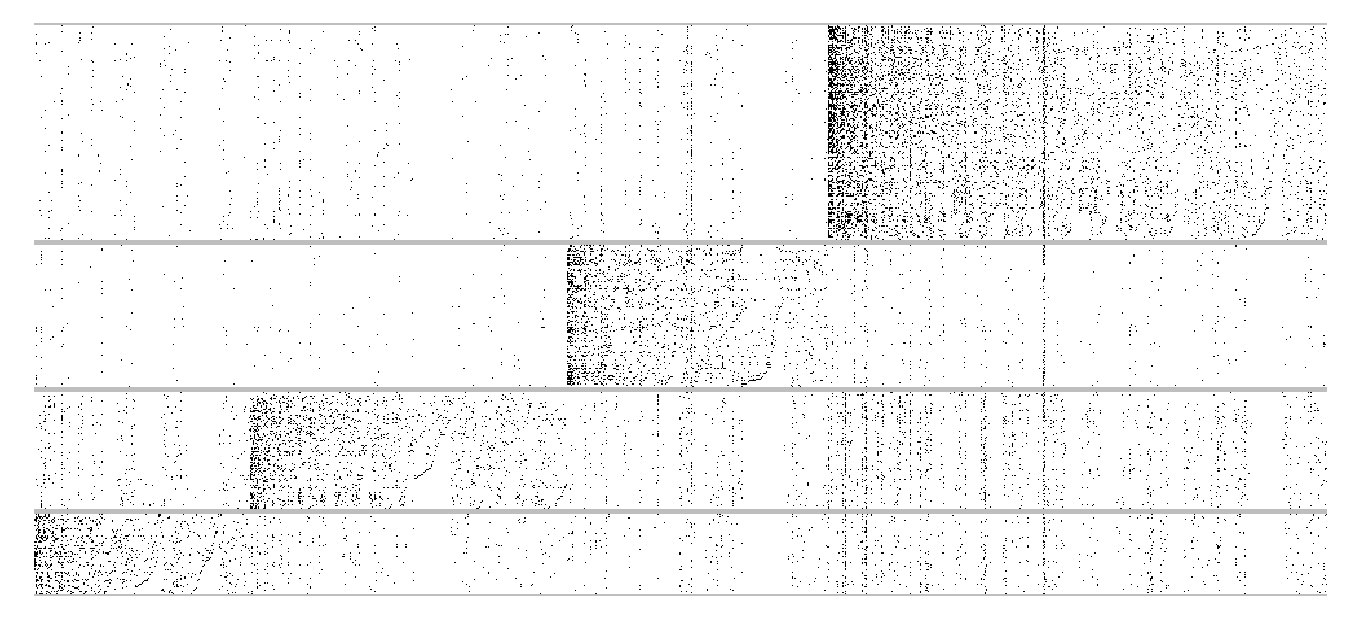}
\caption{Representation of the CSTR data set reorganized as the directional means obtained by dbmovMFs.}
  \label{fig:cstr:sparse:data:db}
\end{figure}

Figure \ref{fig:cstr:sparse:data:db} represents the full data set
using the same ordering: it shows clearly that the coclustering provides only
a crude approximation of actual structure of the data. For instance, the
smallest cluster (bottom row of Figure \ref{fig:cstr:sparse:data:db}) uses all
the words/dimensions that should be specific to the other clusters. Dark
vertical lines on the figure show that some words/dimensions are common to all
clusters. The diagonal structure enforced by dbmovMFs is very useful to bring
stability to the model estimation and to recover the overall clustering
structure, but it appears to be to simplistic to capture the true sparsity
structure. 

\begin{figure}[htbp]
  \centering
\includegraphics{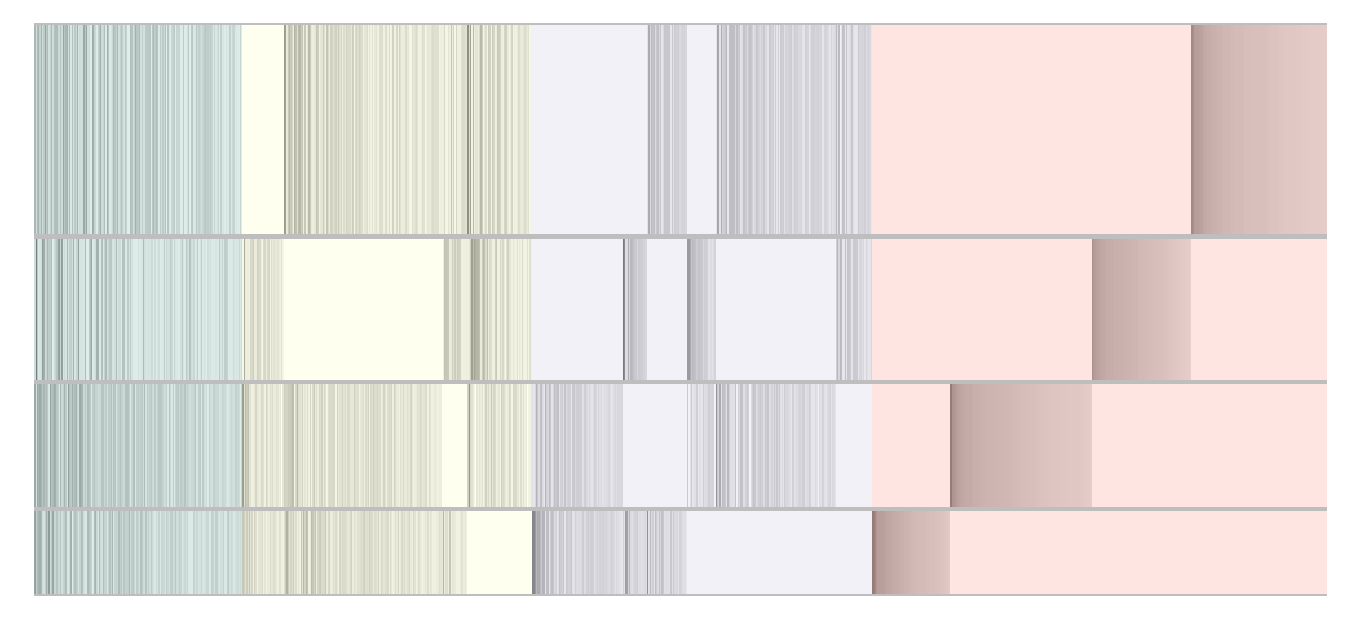}
\caption{Representation of the directional means obtained by the mixture of
  vMF with shared $\kappa$ on the CSTR data set.}
  \label{fig:cstr:sparse:proto:dense}
\end{figure}

Figure \ref{fig:cstr:sparse:proto:dense} shows the structure of the
directional means for the mixture of vMF obtained without
regularization. As shown be the colors, there are four blocks of dimension:
from the block of dimensions/words common to all texts on the left to the
block of cluster specific words. The two intermediate blocks corresponds
respectively to vocabulary shared by 3 clusters and 2 clusters. This
representation confirms that there are indeed specific coordinates but it shows that the
clusters share dimensions in a large proportion, confirming that dbmovMFs
hides most of the structure. 

\begin{figure}[htbp]
  \centering
\includegraphics{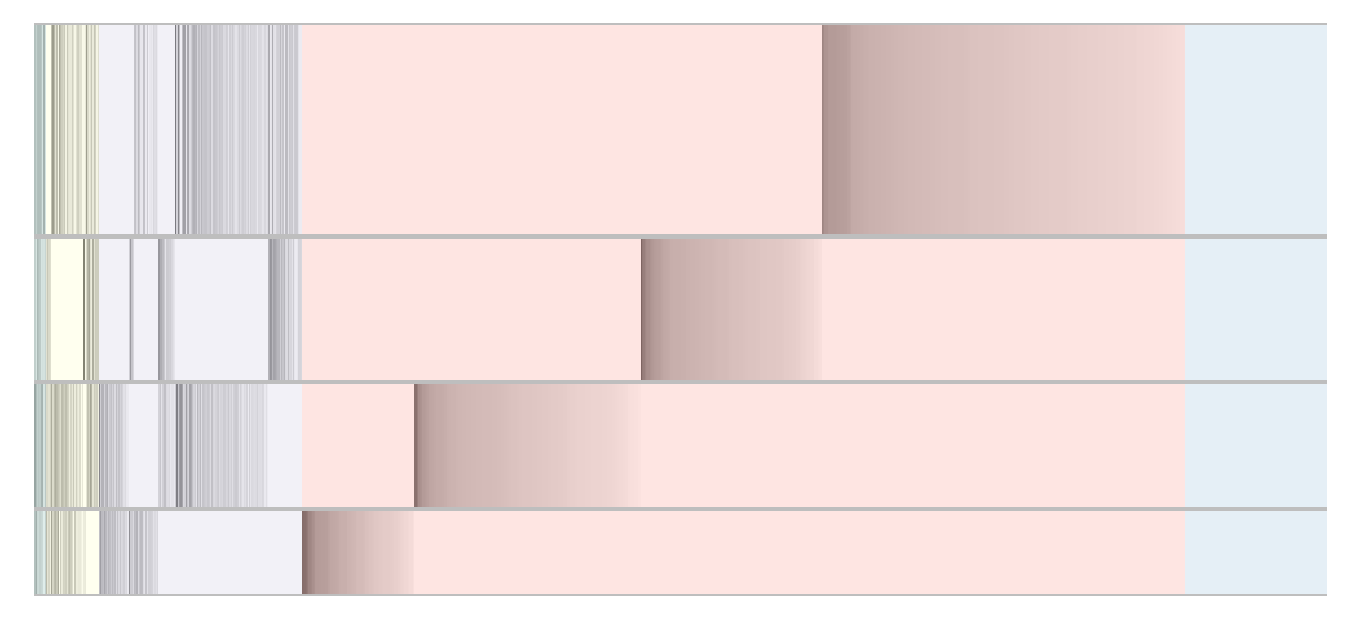}
\caption{Representation of the directional means obtained by the mixture of
  vMF with shared $\kappa$ and regularisation on the CSTR data set.}
  \label{fig:cstr:sparse:proto:bic}
\end{figure}

Figure \ref{fig:cstr:sparse:proto:bic} represents the directional
means obtained by selecting with the BIC the best sparse model along the
$\beta$ path. The result is a compromise between the strictly diagonal
structure obtained by dbmovMFs and the denser solution
obtained without regularisation. It isolate better the specific
dimensions/vocabulary while keeping a smaller subset of shared
dimensions. Notice also that we have now a fifth block of dimensions: those
can be considered as noise dimensions as the corresponding coordinates are
uniformly null in the directional means. 

\begin{figure}[htbp]
  \centering
\includegraphics{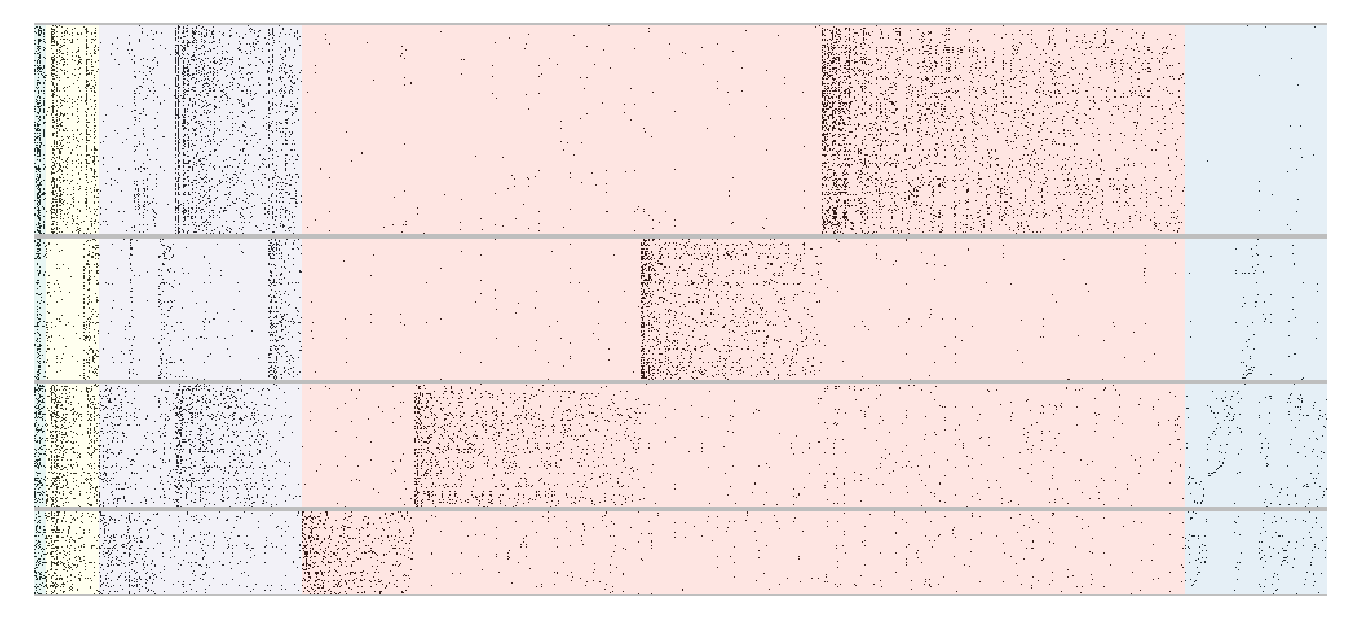}
\caption{Representation of the CSTR data set reorganized as the directional means obtained by the mixture of vMF with shared $\kappa$ and regularisation.}
  \label{fig:cstr:sparse:data:bic}
\end{figure}

This is confirmed by Figure \ref{fig:cstr:sparse:data:bic} that shows the data
set reorder in the same way as the directional means according to the sparse
mixture of vMF. The reordering reveals in a clearer way the underlying
structure of the data. In particular the pink area which corresponds to the
diagonal substructure with ``private'' vocabulary is far less noisy than in
the case of dbmovMFs. 

\subsection{Conclusion}
The comparisons conducted in the Section have shown several important
results. On relatively dense data, the spherical k-means and the mixture of
vMF distribution behave in a similar way. The mixture model is more flexible
in terms of unbalanced between the clusters and recovers them with less data
than the spherical k-means as a consequence of modeling explicitly the
concentration of each cluster. On the contrary, the strong constraints of
dbmovMFs prevents it from inferring a correct structure for relatively dense
data.

On sparse data, dbmovMFs tends to be more stable and more robust against
mispecification than both spherical k-means and vMF distribution mixtures. The
mixture model with component specific concentration parameters should be
avoided when the number of observations is not significantly larger than the
dimensions. A shared concentration parameter is sufficient to bring stability
to the mixture model, but mispecification remains a problem. Overall, the best
results are obtained by the sparse mixture proposed in the paper. In terms of
recovering the clustering structure it obtains results roughly identical to
the ones obtained by the other models, but it reveals patterns in the
directional means that are more consistent with the data than the diagonal
structure imposed by dbmovMFs.

\section{Exploratory analysis on 8-K reports 2015 - 2019 for Wells Fargo}\label{sec:exper-real-world:WFC}

Following \cite{LEE14.1065} and completing the database proposed by \cite{RNTI/papers/1002690}, we create a dataset which focus on 8-K reports. An 8-K is a report of unscheduled material events or corporate changes at a company that could be of importance to the shareholders or the Securities and Exchange Commission (SEC). Also known as a Form 8K, the report notifies the public of events, including acquisitions, bankruptcy, the resignation of directors, or changes in the fiscal year\footnote{A complete list can be found at \url{https://www.sec.gov/fast-answers/answersform8khtm.html}}. We have compiled this dataset, thanks to SEC's EDGAR tool\footnote{\url{https://www.sec.gov/edgar/searchedgar/companysearch.html}}, for the years 2015 - 2019 on all companies from the Standard and Poors 500\footnote{It is a stock market index tracking the performance of 500 large companies listed on stock exchanges in the United States.}.

The corpus contains $37,238$ reports issued by $592$ companies. The texts were pre-processed by applying a classical pipeline:

\begin{itemize}
    \item removal of non-alphanumeric characters;
    \item lemmatisation;
    \item removal of words appearing less than $100$ times and stopwords: we obtain a dictionary of $70223$ distinct roots for the whole corpus.
\end{itemize}

The number of reports produced over the period varies greatly depending on the company concerned. A preliminary analysis shows that the vocabulary of the texts depends heavily on the company, in particular because of the different sectors of activity but above all depending on the context (economic, social, etc.). We therefore carry out the exploration company by company and in particular for this article to focus on Wells Fargo\footnote{ Wells Fargo is an American multinational financial services company.} (WFC) as they published the most during this period.

This company published $672$ reports for the years $2015$ and $2019$ and out of $25$ possible events, only $7$ are represented, with a domination of the event \textit{financial statements and exhibits}, which tends to show that these reports are mainly about the financial state of the company (see table \ref{tab:events:WFC} for event titles and their frequencies). Note that reports can share multiple events. Only $4377$ words (roots) are used in the reports and this dataset is as follow $N=672$ in dimension $d=4377$.


\begin{table}[htbp]
  \centering\small
 \scalebox{0.8}{ \begin{tabular}{cp{8cm}c}
  \noalign{\smallskip}\hline\noalign{\smallskip}
Code&    Type    &Frequencies \\ 
\noalign{\smallskip}\hline\noalign{\smallskip}
1& \emph{\textit{ Financial Statements and Exhibits}} & 658\\
2&\emph{\textit{ Results of Operations and Financial Condition }} &  24\\
3& \emph{\textit{  Amendments to Articles of Incorporation or Bylaws; Change in Fiscal Year }}&19\\
4& \emph{\textit{ Departure of Directors or Certain Officers; Election of Directors; Appointment of Certain Officers: Compensatory Arrangements of Certain Officers}}&27\\
5&  \emph{\textit{ Submission of Matters to a Vote of Security Holders }}&5\\
6&  \emph{\textit{ Other Events}}&36\\
7& \emph{\textit{ Amendments to the Registrant's Code of Ethics, or Waiver of a Provision of the Code of Ethics }} &2\\
\noalign{\smallskip}\hline\noalign{\smallskip}
  \end{tabular}}
  \caption{Wells Fargo Events}
  \label{tab:events:WFC}
\end{table}

In what follows, we will first analyse our dataset with the reference models and then with our own.

\subsection{Reference models}

\begin{figure}[htbp]
  \centering
\includegraphics{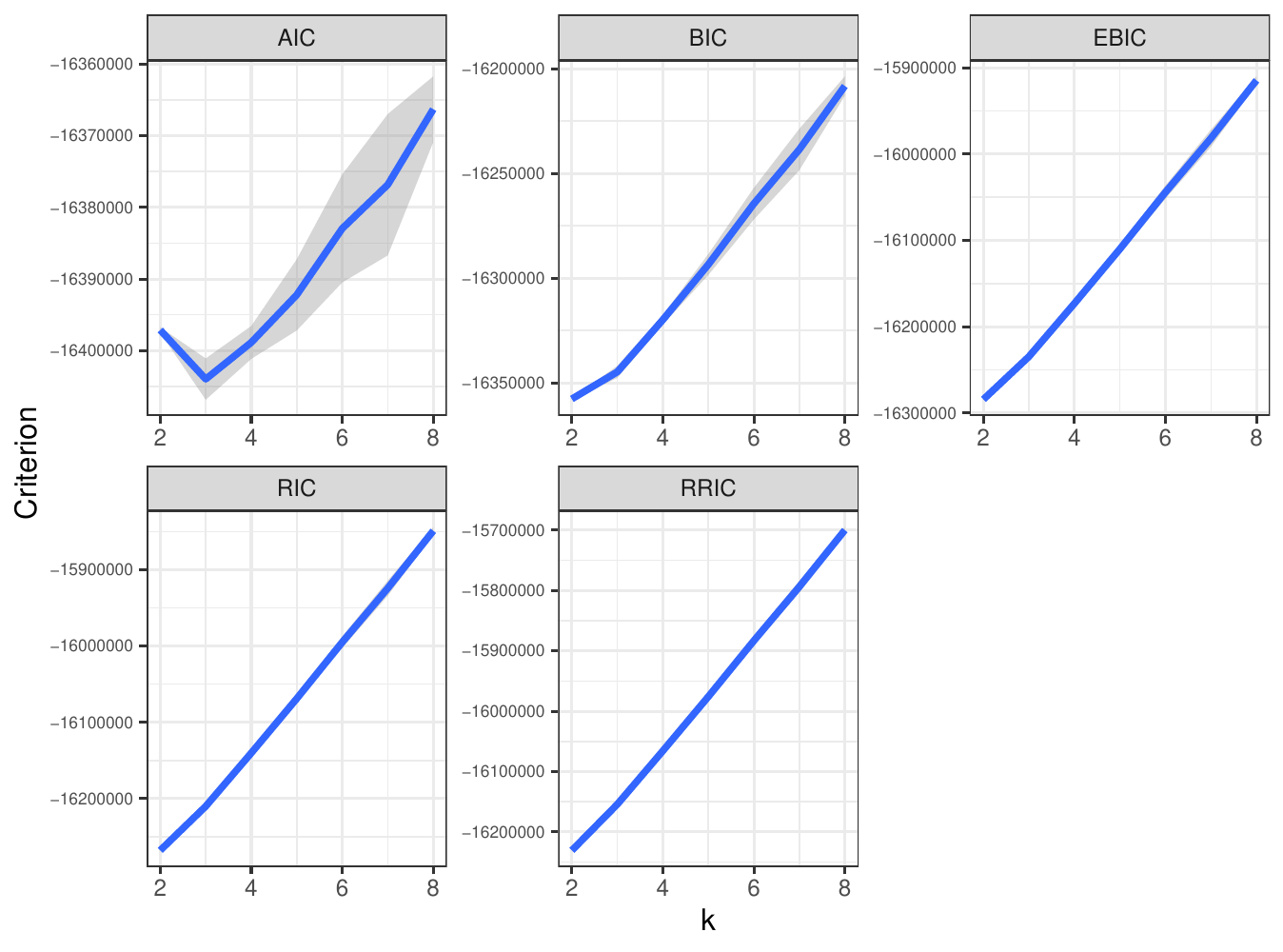}   
  \caption{Model selection criteria for dbmovMFs concerning the analysis of Wells Fargo: the blue curve is the
    mean value, while the grey envelop displays a 2 standard deviation tube
    around it.}
  \label{fig:wfc:dense:db:icnew}
\end{figure}

As the number of clusters $K$ is unknown in this case, we used different methods depending on the reference models. For dbmovMFs, AIC was used and it selected $K=3$, as seen in Figure \ref{fig:wfc:dense:db:icnew}. Whereas for Sk-means, we used the Calinski-Harabasz index \cite{calinskiharabaszindex} and obtained $K=2$.

Table \ref{tab:WFC:dense:db_sk:clust} shows the distribution of reports by cluster obtained by both algorithms. We can note that in both cases, one class is predominant and the second class of Sk-means is dispersed in the three classes of dbmovMFs. Moreover, an ARI of more than $80\%$ shows the similarities between these two clustering.

\begin{table}[ht]
\centering
\begin{tabular}{rrrrrrrrrrrrrrr}
 \hline
 &\multicolumn{3}{c}{Clusters}\\
Algorithms& 1 & 2 & 3  \\ 
 \hline
 Sk-means &    570 & 102 & -  \\ 
dbmovMFs&    22 & 53& 597  \\ 
  \hline
\end{tabular}
\caption{Distribution of reports by cluster obtained by dbmovMFs selectionned by AIC and Sk-means with the Calinski-Harabasz index.}
  \label{tab:WFC:dense:db_sk:clust}
\end{table}

For these reasons, we will now focus on the analysis of the clustering obtained with dbmovMFs.

\begin{figure}[htbp]
  \centering
\includegraphics{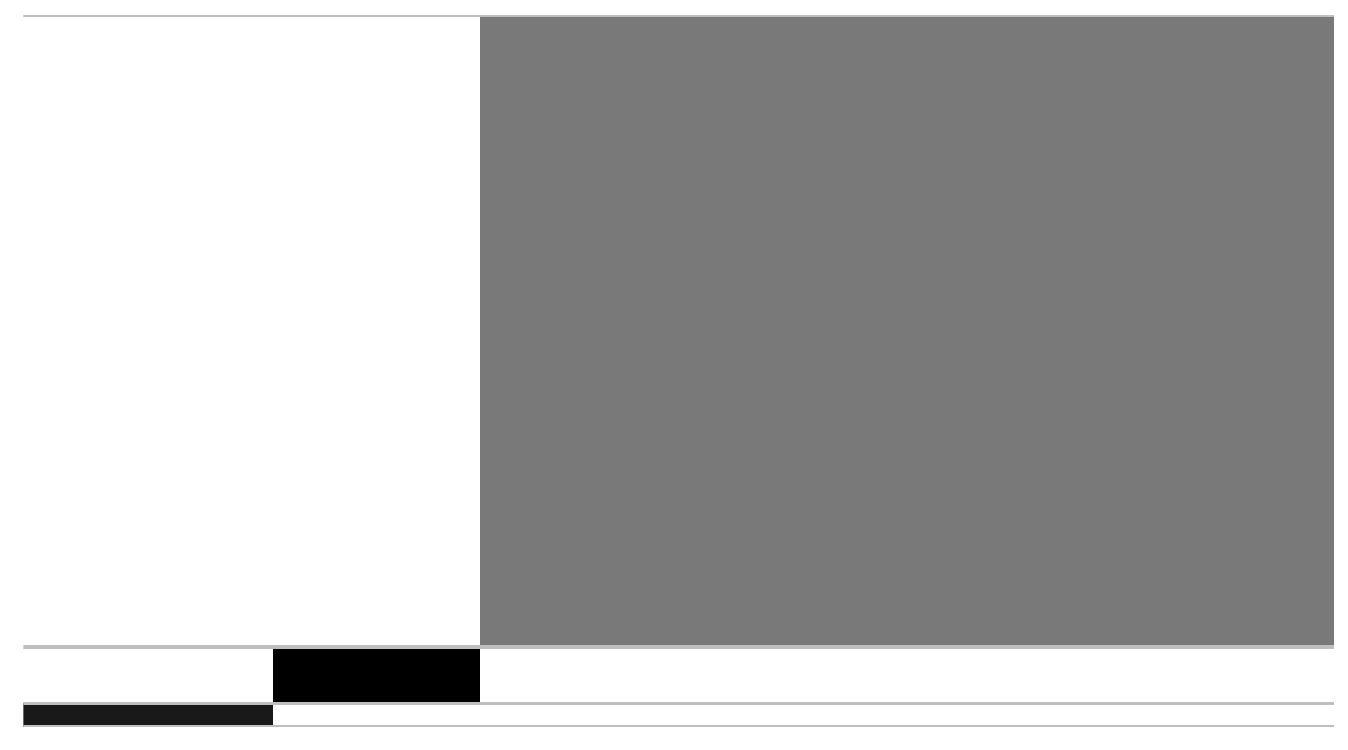}
\caption{Representation of the directional means obtained by dbmovMFs on the Wells Fargo data set.}
  \label{fig:wfc:sparse:proto:db}
\end{figure}

Figure \ref{fig:wfc:sparse:proto:db} represents the block structure obtained
by dbmovMFs. As observed previously, the dbmovMFs solution hides most of the structure and does not facilitate a detailed analysis.

\begin{figure}[htbp]
  \centering
\includegraphics{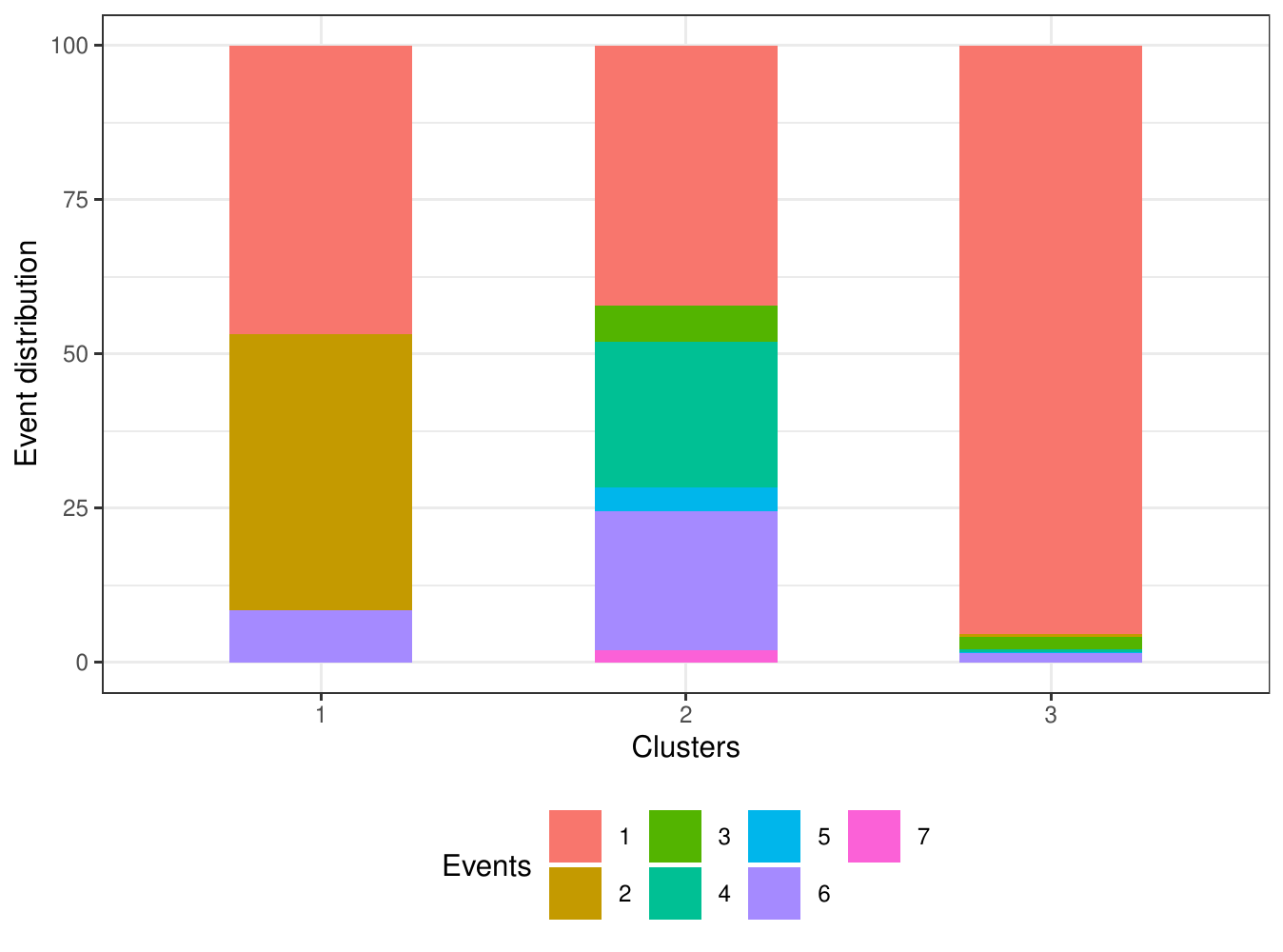}   
  \caption{Distribution of events by cluster in the Wells Fargo dataset obtained with dbmovMFs.}
  \label{fig:WFC:db:event:distribution}
\end{figure}

\begin{figure}[htbp]
  \centering
\includegraphics{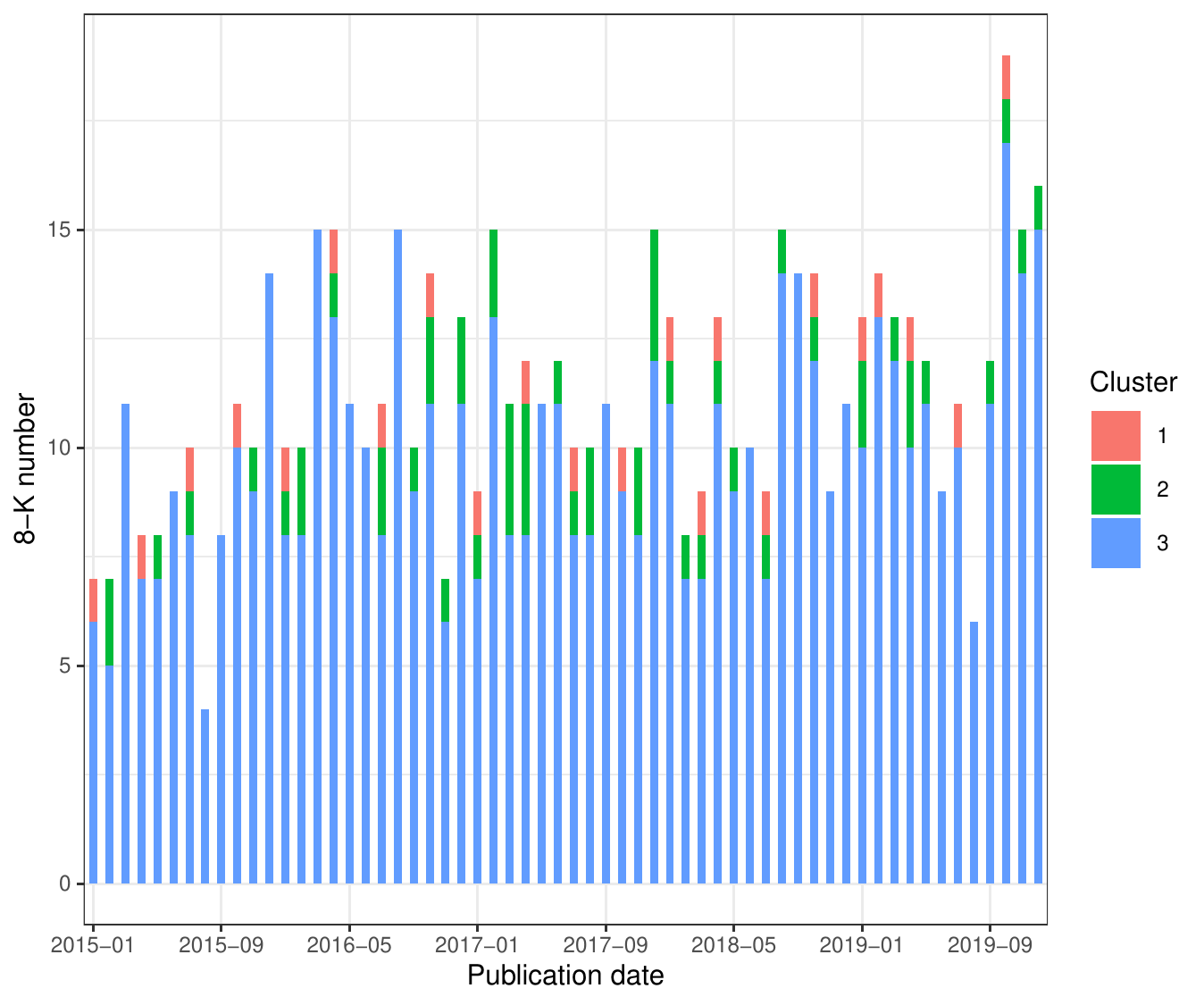}   
  \caption{Distribution of the reports' number per cluster by month in the Wells Fargo dataset obtained with dbmovMFs.}
  \label{fig:WFC:db:cluster:date}
\end{figure}

Figure \ref{fig:WFC:db:event:distribution} shows the distribution of events by cluster. 
It appears that the $1$ is largely composed by financial reports with events as 
\textit{ Financial Statements and Exhibits} and \textit{ Results of Operations and Financial Condition}. From Figure \ref{fig:WFC:db:cluster:date}, we can assert that reports of this class are quarterly reports. Class 2 is mainly concerned by specific events such as \textit{Departure of Directors or Certain Officers; Election of Directors; Appointment of Certain Officers: Compensatory Arrangements of Certain Officers} or \textit{Submission of Matters to a Vote of Security Holders}. Figure \ref{fig:WFC:db:cluster:date} exhibits that this class appears when the company has had to face a negative context and has wanted to reorganise. Class $3$, consisting mainly of the event \textit{Financial Statements and Exhibits}, concerns the company's various financial communications. However, unlike the detailed analysis possible with the mixture of vMF that we develop below, it is very difficult here to see the different aspects of its financial communication and the financial products it issues.

\subsection{Mixture of vMF with a common $\kappa$ parameter}
\begin{figure}[htbp]
  \centering
\includegraphics{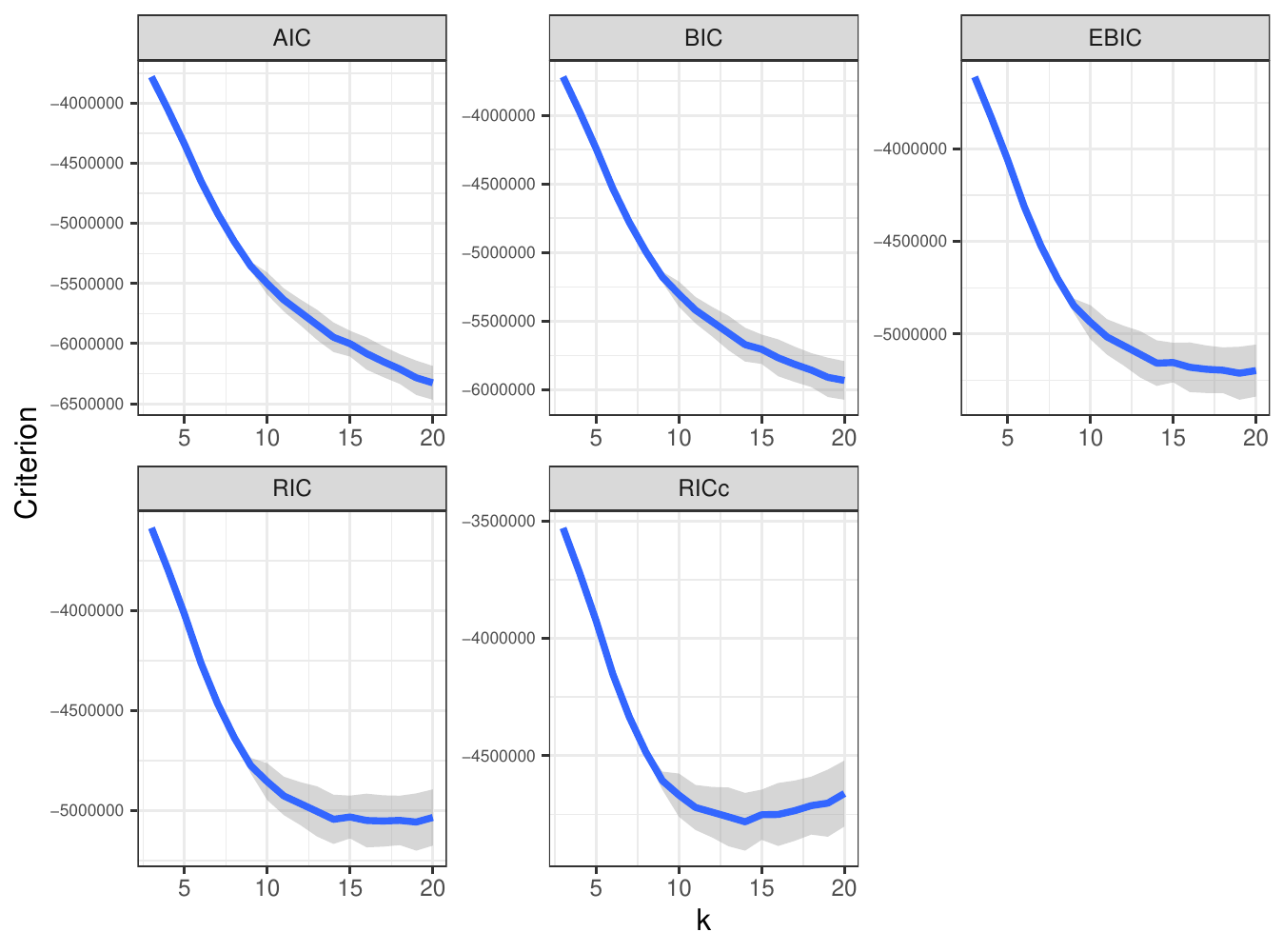}   
  \caption{Model selection criteria for the mixture of
    vMF distributions with a common $\kappa$ parameter concerning the analysis of Wells Fargo: the blue curve is the
    mean value, while the grey envelop displays a 2 standard deviation tube
    around it.}
  \label{fig:WFC:dense:movmf:ic}
\end{figure}

To select the number of clusters $K$, we proceed as exposed previously using the mixture of vMF with a common $\kappa$ parameter. As the first step, we select the number of components thanks to the RICc to obtain $K=14$ as shown in Figure \ref{fig:WFC:dense:movmf:ic}.

\begin{figure}[htbp]
  \centering
\includegraphics{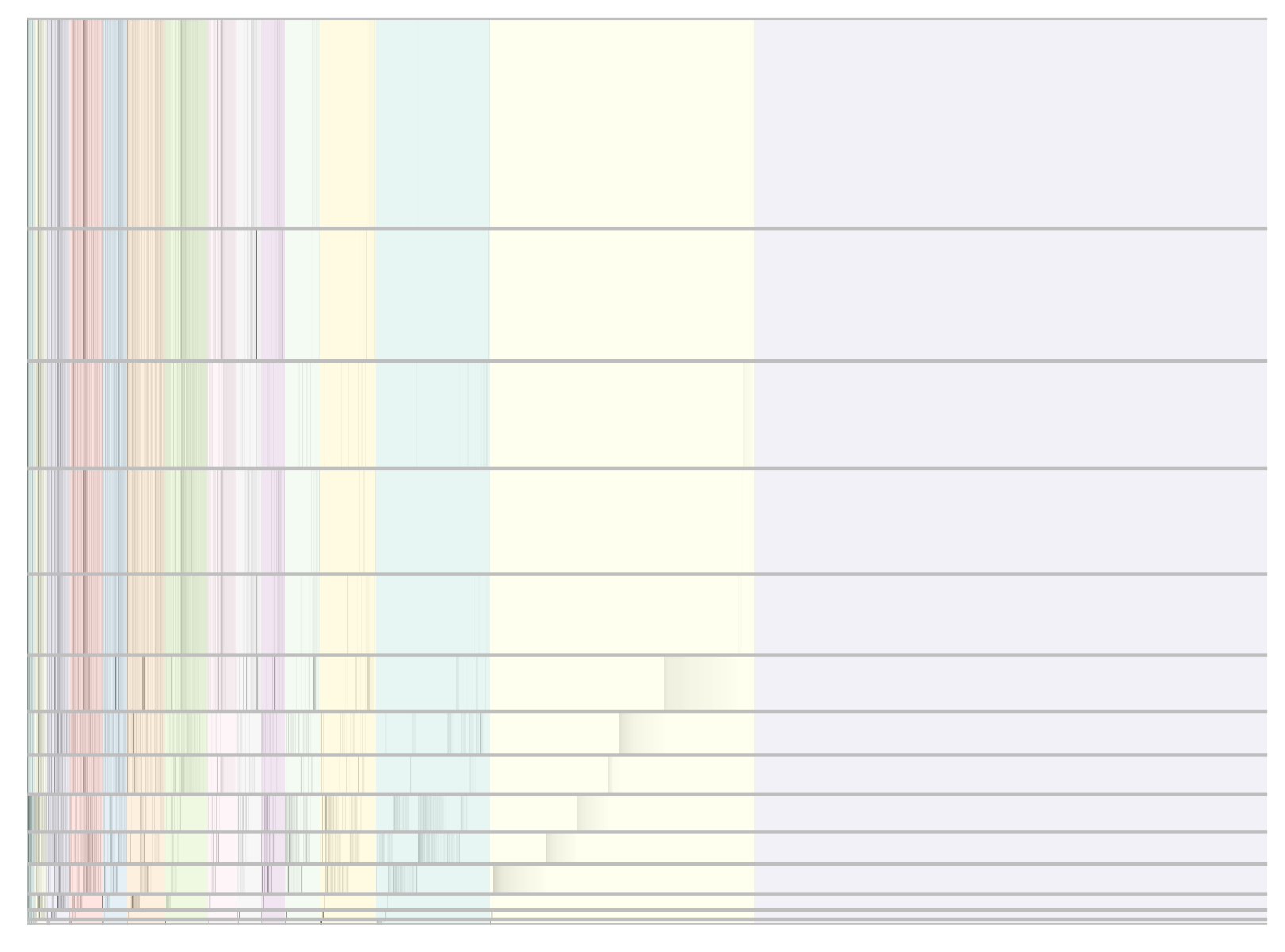}
\caption{Representation of the directional means obtained by the mixture of
  vMF with shared $\kappa$ and regularisation on the Wells Fargo data set.}
  \label{fig:wfc:sparse:data:bic}
\end{figure}

\begin{figure}[htbp]
  \centering
\includegraphics{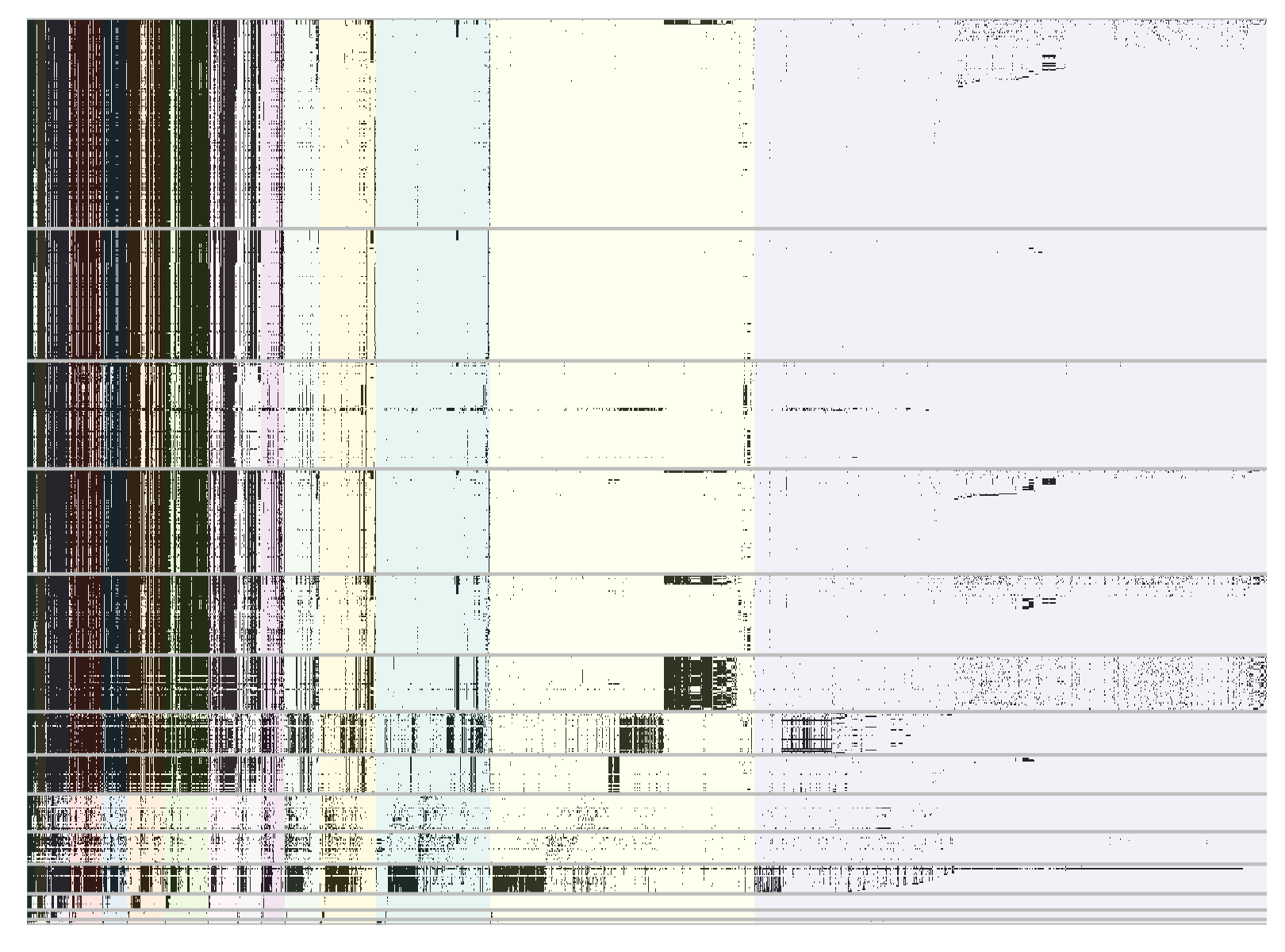}
\caption{Representation of the Wells Fargo data set reorganized as  directional means obtained by the mixture of
  vMF with shared $\kappa$ and regularisation.}
  \label{fig:wfc:sparse:data:full:bic}
\end{figure}

In the second step, the sparsity level was selected using the path following
strategy with a maximum of $1000$ steps and the minimal relative increase
between two values of $\beta$ set to $0.01$. A $\beta$ of $1072.253$ and a sparsity of $82.16\%$
were obtained. Figures \ref{fig:wfc:sparse:data:bic} represents the directional
means. Figure \ref{fig:wfc:sparse:data:full:bic} exhibits the data set reorganized as directional means. It reveals in a clearer way the underlying structure of the data.

Table \ref{tab:WFC:sparse:clust} shows the distribution of reports by cluster obtained. The result is very different from those obtained previously with ARIs below $9\%$ in comparison to clusterings of dbmovMFs and Sk-means. We can note that cluster $3$ is the biggest one with $156$ reports while clusters $1$ and $4$ are composed of very few of them and must be focused on one topic. Moreover, It appears that class $13$ is identical to class $1$ found by dbmovMFs.

\begin{table}[htbp]
\centering
\begin{tabular}{rrrrrrrrrrrrrrr}
 \hline
 &\multicolumn{14}{c}{Clusters}\\
& 1 & 2 & 3 & 4 & 5 & 6 & 7 & 8 & 9 & 10 & 11 & 12 & 13 & 14 \\ 
 \hline
Nb. 8-K &   4 &  80 & 156 &   7 &  12 &  78 &  32 &  60 &  42 &  98 &  24 &  28 &  22 &  29 \\ 
  \hline
\end{tabular}
\caption{Distribution of reports by cluster obtained by the sparse model selectionned by RICc with the path following approach.}
  \label{tab:WFC:sparse:clust}
\end{table}

For its part, table \ref{tab:WFC:sparse:words} shows the unique words by cluster and those in common. The latter makes sense in that they contain generic terms in the company's reports, such as its name or the name of a financial instrument for example. More interesting are the unique words for each cluster as they form coherent subjects. Note the exception for clusters 5 and 10, which share all their representatives' words with at least another cluster.

\begin{figure}[htbp]
  \centering
\includegraphics{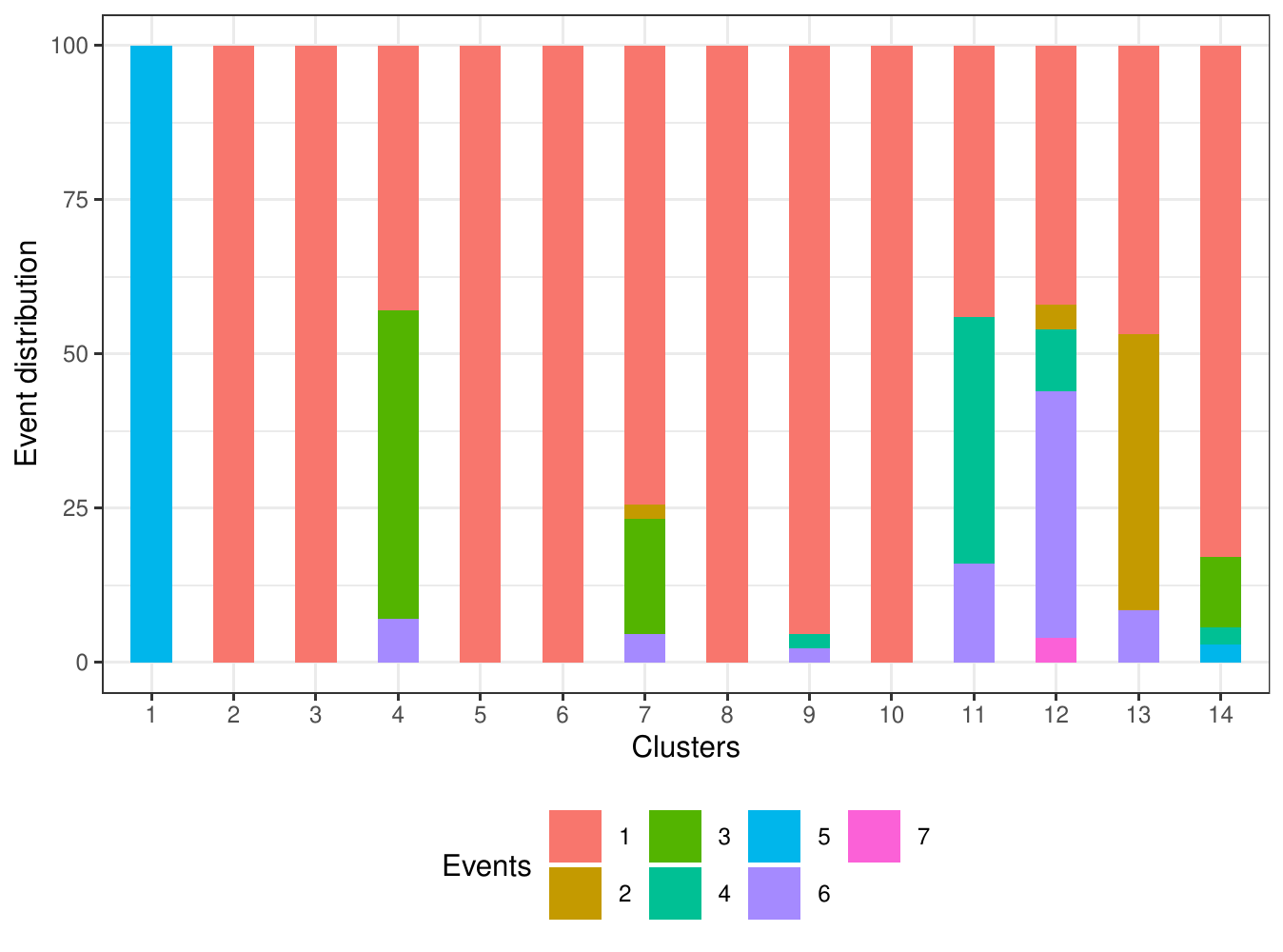}   
  \caption{Distribution of events by cluster in the Wells Fargo dataset with the model obtained by the path following approach.}
  \label{fig:WFC:event:distribution}
\end{figure}

\begin{figure}[htbp]
  \centering
\includegraphics{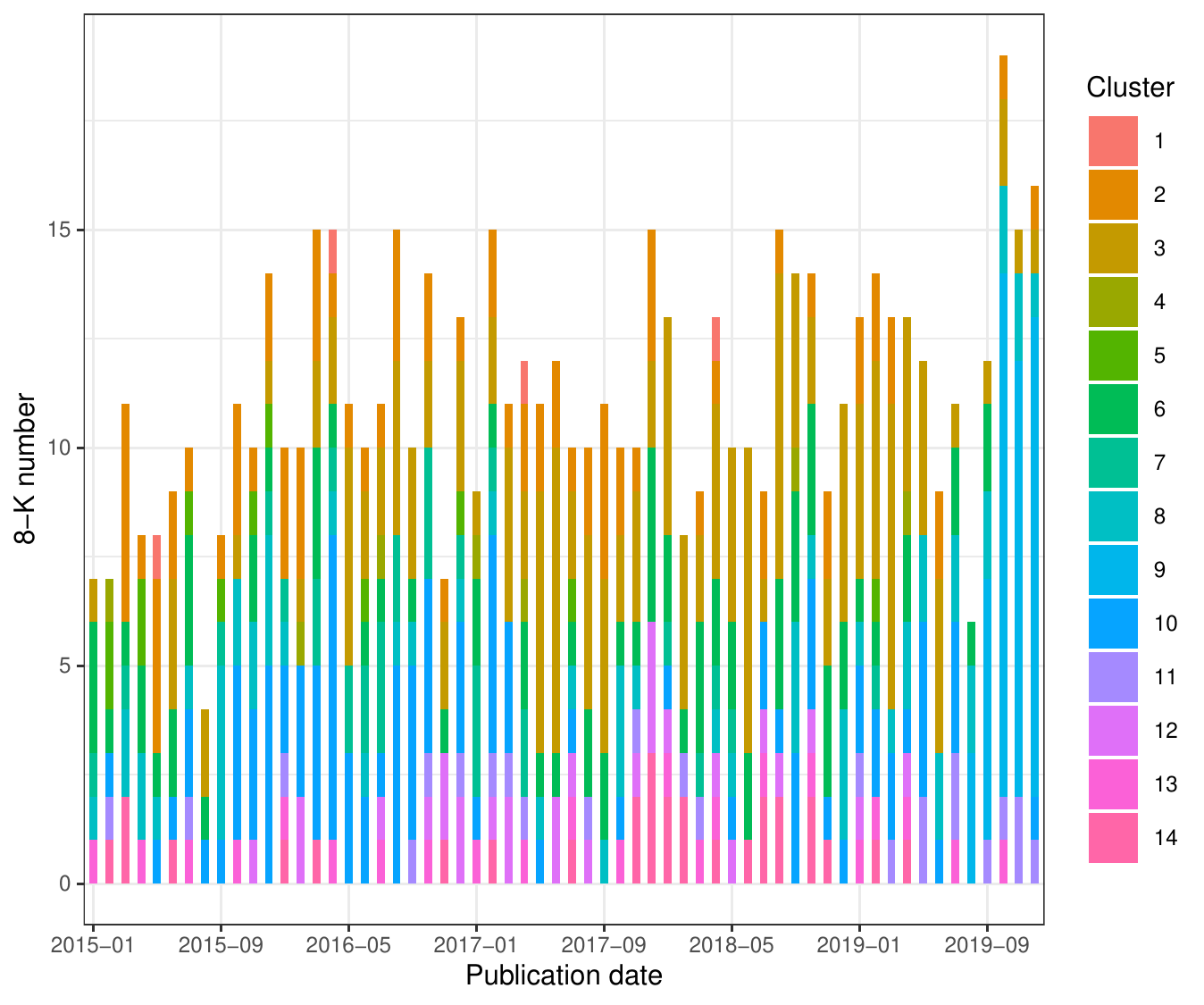}   
  \caption{Distribution of the reports' number per cluster by month in the Wells Fargo dataset with the model obtained by the path following approach.}
  \label{fig:WFC:cluster:date}
\end{figure}

For instance, unique words of cluster 1 - \textit{abstention}, \textit{cast}, \textit{ratify}, \textit{shareowner} - are from the annual meeting lexicon. Figure \ref{fig:WFC:event:distribution} shows that this cluster is entirely composed of the \textit{Submission of Matters to a Vote of Security Holders} event which takes place annually as visible in Figure \ref{fig:WFC:cluster:date}. Figure \ref{WFC:exemple_8k:cluster1} shows an extract of a report from Cluster 1 published by Wells Fargo on 1 May 2015\footnote{The full text is available at \url{https://www.sec.gov/Archives/edgar/data/0000072971/000119312515166149/d920037d8k.htm}}.  Words in \textit{\textcolor{blue}{blue}} represent the common words between all Clusters and in \textit{\textcolor{red}{red}}, the ones specific to this Cluster.

\begin{table}[htbp]
\centering
\scalebox{0.8}{\begin{tabular}{rllllll}
  \hline
  Cluster & 1 & 2 & 3 & 4 & 5 \\ 
  \hline
1   & abstention & cast & ratify & shareowner & -  \\ 
  2   & continuance & bankrupt & insolvent & receiver & annually \\ 
  3   & vme & monthly & shewchuk & sonia & cqr \\ 
  4   & advisable & convene & nonassessable &-  & - \\ 
  5   & - & - & - & - &  \\ 
  6   & sector & bad & homebuilders & gold & miner \\ 
  7   & untrue & omission & canadian & directive & representation \\ 
  8   & adr & absent & determinable & fluctuation & bloomberg \\ 
  9   & domainitemtype & false & thinterestinshareof & shr & text \\ 
  10   & - & - & - & - & - \\ 
  11   & defendant & chair & bonus & rsrs & hear \\ 
  12   & mack & banker & unauthorized & parent & controller \\ 
  13   & portfolio & revenue & offs & sep & jun \\ 
  14   & gics & spin & otc & bulletin & antidilution \\ 
  \hline
  \textit{commun} & security & company & any & well & fargo \\ 
   \hline
\end{tabular}}
\caption{Unique words for each cluster obtained by the sparse model selectionned by RICc with the path following approach. The row \textit{commun} shows words  shared by all clusters.}
  \label{tab:WFC:sparse:words}
\end{table}

\begin{figure}[htbp]
    \begin{quote}
       \begin{description}
    \item[Event :] \emph{Submission of Matters to a Vote of Security Holders.};
    \item[Text :] [\ldots] \textit{\textcolor{blue}{well}}s \textit{\textcolor{blue}{fargo}}  \textit{\textcolor{blue}{company}}  held its annual meeting of stockholders on april 28, 2015. at the meeting, stockholders elected all 16 of the directors nominated by the board of directors as each director received a greater number of votes \textit{\textcolor{red}{cast}}  for  his or her election than votes \textit{\textcolor{red}{cast}} [\ldots] \textit{\textcolor{red}{ratify}} the appointment of kpmg llp as independent registered public accounting firm for 2015  [\ldots].
\end{description}
    \end{quote}
    \caption{\label{WFC:exemple_8k:cluster1}Example of a Cluster 1 8-K report published on 1 May 2015. \textit{\textcolor{blue}{Words}} show the commun ones between all cluster and \textit{\textcolor{red}{words}}, the ones specific to cluster 1.  }
\end{figure}

If we now look at Cluster 11, which appears randomly over time in Figure \ref{fig:WFC:cluster:date}, it is composed of events \textit{Financial Statements and Exhibits}, \textit{Other Events} and especially \textit{Departure of Directors or Certain Officers; Election of Directors; Appointment of Certain Officers: Compensatory Arrangements of Certain Officers}. This cluster focuses on changes in the board and their possible consequences on the company's results. Figure \ref{WFC:exemple_8k:cluster11} shows an extract of a report from Cluster 11 published by Wells Fargo on 12 October 2016\footnote{The full text is available at \url{https://www.sec.gov/Archives/edgar/data/0000072971/000119312516736870/d271369d8k.htm}} notifying the departure of CEO John Stumpf in the wake of numerous scandals\footnote{Example of scandal faced by Wells Fargo \url{https://www.cnbc.com/2016/10/20/wells-fargo-just-lost-its-accreditation-with-the-better-business-bureau.html}.}. It is interesting to note that the unique words of this Cluster express this context. First, the word \textit{chair} refers to a person who sits on the Board of Directors. Second, the term \textit{defendant} implies legal proceedings. Finally, terms \textit{bonus} and \textit{rsrs}\footnote{RSRs is the acronym for Restricted Share Rights.} mention compensation due to the turnover of board members.

\begin{figure}[htbp]
    \begin{quote}
       \begin{description}
    \item[Event :] \emph{Departure of Directors or Certain Officers; Election of Directors; Appointment of Certain Officers: Compensatory Arrangements of Certain Officers \& Financial Statements and Exhibits.};
    \item[Text :] [\ldots]  on october 12, 2016, john g. stumpf notified \textit{\textcolor{blue}{well}}s \textit{\textcolor{blue}{fargo}}  \textit{\textcolor{blue}{company}} ) of his decision to retire as chairman and chief executive officer and a director of the \textit{\textcolor{blue}{company}}, effective immediately. [\ldots]  elected director elizabeth a. duke as the \textit{\textcolor{blue}{company}} s non-executive vice \textit{\textcolor{red}{chair}}.  [\ldots].
\end{description}
    \end{quote}
    \caption{\label{WFC:exemple_8k:cluster11}Example of a Cluster 11 8-K report published on 12 October 2016. \textit{\textcolor{blue}{Words}} show the commun ones between all cluster and \textit{\textcolor{red}{words}}, the ones specific to cluster 11.  }
\end{figure}

Let us now focus on clusters that are made up of the same single event type and do not have unique terms such as clusters 5 and 10, as seen in Figure \ref{fig:WFC:event:distribution}. Figure \ref{fig:WFC:cluster:date} shows that these clusters appear differently over time. Cluster 5 focuses mainly on the period before the resignation of the CEO, i.e. before October 2016, while cluster 10 is found significantly in two periods, i.e. between July 2015 and March 2017 but also between April 2018 and July 2019. These two periods correspond to many legal cases but also to setbacks in business for Wells Fargo. These include high exposure to the fall in oil prices in January 2016 and numerous settlements of fines for fraudulent business practices in April 2018 and concerning the sub-prime crisis in August 2018. An in-depth reading of the texts of these clusters reveals a common subject between them, namely medium-term notes, but of different series and different underlying assets. Cluster 10 is related to medium-term notes, series K, linked to  indexes based on Emerging Markets such as the \textit{iShares MSCI Emerging Markets ETF}\footnote{The iShares MSCI Emerging Markets ETF seeks to track the investment results of an index composed of large- and mid-capitalization emerging market equities.} or developped market as the \textit{MSCI EAFE Index}\footnote{The MSCI EAFE Index is an equity index which captures large and mid cap representation across 21 Developed Markets countries
around the world, excluding the US and Canada.}. Cluster 5 is associated to medium-term notes, series N, linked to reference rates\footnote{More details available at: \url{https://saf.wellsfargoadvisors.com/emx/dctm/Marketing/Marketing_Materials/Fixed_Income_Bonds/e7434.pdf}}. These clusters, therefore, show that the company has issued different types of debt to cope with its context and ensure its financing needs.

Finally, the previous analysis shows the advantages of our method comparing to dbmovMFs for an exploratory analysis. It exhibits the specialisation of each of the clusters which allows an easy understanding of the different events that impact a company over time.  Moreover, when they exist, unique words to each cluster give a precise idea of the main subject of said cluster. For their part, shared terms between all clusters provide an overview of the corpus' subject.

\section{Conclusion}
In this article, we have proposed to estimate a mixture of von Mises-Fisher distributions using
a $l_1$ penalized likelihood. This model attempts to learn sparse directional
means without enforcing a diagonal structure, contrarily to dbmovMFs. Sparse
directional means provide a way to understand the data structure and to
interpret the clustering induced by the mixture model. 

The maximisation of the penalized likelihood is implemented via
expectation-maximization. To avoid estimating parameters from scratch for
different trade-offs between the likelihood and the penalty term, we introduced
a path following approach that detect automatically important change in the
sparsity of the solutions. We showed that selecting the best trade-off can
then be done using the BIC. We also confirmed previous results about the difficulty of
selecting the number of components of the mixture in the high dimensional case
with a relatively low number of observations. Finally, we proposed a pixel
oriented visualisation technique to represent sparse directional means and
provide a first insight on the structure of the data. 

Extensive qualitative and quantitative experiments on differents data sets,
including a new dataset of Wells Fargo 8-K reports, demonstrate the practical interest of
the proposed model. Indeed, the sparsity of the directional means obtained
eases the interpretation of results while achieving similar or better
results in terms of ARI.

However, our results also confirm that dbmovMFs remains more stable than a
mixture of vMF distributions, essentially as a consequence of its low
concentration parameters. As the diagonal structure enforced on the
directional means is very strong, the clusters obtained by dbmovMFs remain
somewhat vague. As shown in our experiments, the directional means obtained by
dbmovMFs are only remotely representative of the true structure of the
data. Using a shared concentration parameter, we managed to bring mixtures of
vMF distributions on par with dbmovMFs when the model is correctly specified
in terms of cluster number. In the future, we will investigate other ways to
constrain the concentration parameters in order to improve the stability of
our model without compromising the quality of the directional means. A
possible solution would be to use a regularisation term on the concentration
parameters, but this introduces at least two difficulties. Firstly the maximisation
phase will be much more complicated, considering that it would introduce a
regularisation term in an already difficult numerical problem (summarized by
equation \eqref{eq:kappa:l1}). Secondly, when all the other parameters are
held constant, the likelihood increases with increasing values of the
concentration parameters. It would therefore be necessary to introduce a way
to define an optimal trade-off between regularizing the concentrations and
maximizing the likelihood. As the regularisation will have to effect on the
number of parameters, information criteria will be of no help in this
setting.

Finally, let us mention that the path-following strategy proposed in this work
could be easily adapted to other penalized models such as the Gaussian mixture
proposed in \cite{JMLR:v8:pan07a}.

\appendix

\section{Derivation of the EM algorithm}\label{appendix:em:derivation}
We derive in this Section the first order optimality conditions of the M phase
of the EM algorithm. 

\subsection{Stationary point equations associated to the
  $\kappa_k$}\label{appendix:kappa}
The Lagrangian \eqref{eq:lagrangien} has partial derivatives with respect to
$\kappa_k$ given by
\begin{equation}
  \label{eq:partialkappa}
   \frac{\partial}{\partial_{\kappa_k}}\mathcal{L}(\boldsymbol{\Theta},\zeta,\boldsymbol{\lambda}|\boldsymbol{\Theta}^{(m)}) =\sum_{i=1}^N\tau^{(m)}_{ik} \left(\frac{c'_d(\kappa_k)}{c_d(\kappa_k)}+ \boldsymbol{\mu}_k^T \boldsymbol{x}_i\right).
 \end{equation}
To simplify this expression, we follow \cite{10.5555/1046920.1088718} and compute
\begin{equation}
  \label{eq:cderiv}
c'_d(\kappa_k)=\frac{1}{(2\pi)^{s+1}I^2_s(\kappa_k)}\left(s\kappa_k^{s-1}I_s(\kappa_k)-\kappa_{k}^sI'_s(\kappa_k)\right),
\end{equation}
where $s=\frac{d}{2}-1$ and $I'_s$ is the derivative of modified Bessel
function of the first kind and order s. As recalled in
\cite{10.5555/1046920.1088718}, this derivative is such that
\begin{equation}
\kappa_kI_{s+1}(\kappa_k)=  \kappa_kI'_{s}(\kappa_k)-sI_{s}(\kappa_k),
\end{equation}
and thus
\begin{equation}
c'_d(\kappa_k)=-\frac{\kappa_k^{s}I_{s+1}(\kappa_k)}{(2\pi)^{s+1}I^2_s(\kappa_k)},
\end{equation}
leading to
\begin{equation}
\frac{c'_d(\kappa_k)}{c_d(\kappa_k)}=-\frac{I_{s+1}(\kappa_k)}{I_{s}(\kappa_k)}.
\end{equation}
Then
$\frac{\partial}{\partial_{\kappa_k}}\mathcal{L}(\boldsymbol{\Theta},\zeta,\boldsymbol{\lambda}|\boldsymbol{\Theta}^{(m)})
=0$ is equivalent to
\begin{equation}
  \frac{I_{d/2}(\kappa_k)}{I_{d/2-1}(\kappa_k)}=\boldsymbol{\mu}_k^T\frac{\sum_{i=1}^n\tau^{(m)}_{ik}
  \boldsymbol{x}_i}{\sum_{i=1}^n\tau^{(m)}_{ik}}.
\end{equation}

\subsection{Stationary point equations associated to the
  $\boldsymbol{\mu}_k$}\label{appendix:mu}
For the directional means, we have to consider the sub-gradient of the
Lagrangian function. We have
\begin{equation}
\partial_{\mu_{kj}} \mathcal{L}(\boldsymbol{\Theta},\zeta,\boldsymbol{\lambda}|\Theta^{(m)})=\kappa_k\left(\sum_{i=1}^n\tau^{(m)}_{ik}x_{ij}\right)-2\lambda_{k}\mu_{kj}-\beta\partial_{\mu_{kj}}\lvert\mu_{kj}\rvert.\label{derivpartielmuttcas}
\end{equation}
Using the well known property of the sub-gradient of the absolute value, we obtain
\begin{eqnarray}
\partial_{\mu_{kj}}\mathcal{L}(\boldsymbol{\Theta},\zeta,\boldsymbol{\lambda}|\Theta^{(m)}) = \left\{
    \begin{array}{ll}
    \{\kappa_kr^{(m)}_{kj}-2\lambda_k\mu_{kj}+\beta\}& \text{when } \mu_{kj}<0,  \\
        \{\kappa_kr^{(m)}_{kj} 
        -\epsilon\beta|\epsilon\in\left[-1; 1\right]\} & \text{when } \mu_{kj}=0, \\
        \{\kappa_kr^{(m)}_{kj}-2\lambda_k\mu_{kj}-\beta\}& \text{when } \mu_{kj}>0, 
    \end{array}
\right
.\label{sousderivpartielmuttcas}
\end{eqnarray}
where
\begin{equation}
\boldsymbol{r}^{(m)}_{k}=\sum_{i}\tau^{(m)}_{ik}\boldsymbol{x}_{i}.  
\end{equation}
The first-order optimality condition is
$0\in\partial_{\mu_{kj}}\mathcal{L}(\boldsymbol{\Theta},\zeta,\boldsymbol{\lambda}|\Theta^{(m)})$,
which leads to the following analysis. 

If we look for a positive solution $\mu_{kj}>0$, the optimality condition is fulfilled when
\begin{equation}
\mu_{kj}=\frac{\kappa_kr^{(m)}_{kj}-\beta}{2\lambda_k}.
\end{equation}
This solution is compatible with $\mu_{kj}>0$ if
$\kappa_kr^{(m)}_{kj}-\beta>0$, that is when
$r^{(m)}_{kj}>\frac{\beta}{\kappa_k}$. In this case we have
also
\begin{equation}
\mu_{kj}=\sign{r^{(m)}_{kj}}\frac{\kappa_k\lvert r^{(m)}_{kj}\rvert-\beta}{2\lambda_k} .
\end{equation}
If we look for a negative solution $\mu_{kj}<0$, then the optimality condition is fulfilled when
\begin{equation}
\mu_{kj}=\frac{\kappa_kr^{(m)}_{kj}+\beta}{2\lambda_k}.
\end{equation}
This is compatible with the hypothesis $\mu_{kj}<0$ if
$\kappa_kr^{(m)}_{kj}+\beta<0$, that is
$r^{(m)}_{kj}<-\frac{\beta}{\kappa_k}$. In this case, we have again
\begin{equation}
\mu_{kj}=\sign{r^{(m)}_{kj}}\frac{\kappa_k\lvert r^{(m)}_{kj}\rvert-\beta}{2\lambda_k} .
\end{equation}
Finally, a zero value, $\mu_{kj}=0$, fulfills the optimality condition if
\begin{equation*}
  0\in\left[\kappa_kr^{(m)}_{kj}+\beta;
  \kappa_kr^{(m)}_{kj}-\beta\right].
\end{equation*}
This is the case when $-\frac{\beta}{\kappa_k}\leq r^{(m)}_{kj} \leq
\frac{\beta}{\kappa_k}$, that is when $\kappa_k\lvert
r^{(m)}_{kj}\rvert-\beta\leq 0$. 

In summary, the first-order optimality condition is fulfilled when
\begin{equation}
\mu_{kj}=\frac{\sign{r^{(m)}_{kj}}}{2\lambda_k}\max(\kappa_k\lvert
r^{(m)}_{kj}\rvert-\beta,0). 
\end{equation}
The Lagrange multipliers are computed using the equality constraints
$\left\lVert\boldsymbol{\mu_k}\right\rVert^2_2=1$. This gives
\begin{align*}
\left\lVert \sum_{j=1}^d\frac{\sign{r^{(m)}_{kj}}}{2\lambda_k}\max(\kappa_k\lvert
  r^{(m)}_{kj}\rvert-\beta,0)\right\rVert_2^2&=1,  \\
\frac{1}{4\lambda_k^2}\sum_{j=1}^d  (\max(\kappa_k\lvert
  r^{(m)}_{kj}\rvert-\beta,0))^2&=1,
\end{align*}
and thus
\begin{equation}
\lambda_k=\frac{1}{2} \sqrt{\sum_{j=1}^d (\max(\kappa_k\lvert  r^{(m)}_{kj}\rvert-\beta,0))^2} .
\end{equation}

\section{Implementation details}\label{appendix:implementation}
We discuss in this Section important technical details about the concrete
implementation of Algorithm \ref{code:EM}. 

Firstly, it is well known that initialisation plays an important part in EM
algorithms. In our case, a simple strategy was sufficient to obtain
satisfactory results. We proceed by selecting uniformly at random without
replacement $K$ observations in the data set $\boldsymbol{X}$ which serve as
initial values for the $(\boldsymbol{\mu}_k)_{1\leq k\leq K}$. Then we perform
crisp assignments of all the observations to their closest directional mean
(with respect to the inner product, i.e. the cosine similarity). This enables
us to compute initial values of $\boldsymbol{\alpha}$ as the ratio of
observations assigned to each prototype. Finally, we compute initial values of
$\boldsymbol{\kappa}$ using the EM estimator, i.e. solving equation
\eqref{eq:kappa:l1} using for the $\tau_{ik}$ the crisp assignment
matrix. Algorithm \ref{code:EM:init} summarizes the process. Notice that the
final estimation can fail and the full process may have to be repeated several
time in order to produce a proper initial configuration (see below for
details).

\begin{algorithm}[htb]
  \caption{EM initialisation}\label{code:EM:init}
  \begin{algorithmic}
    \STATE{Select uniformly at random $(\boldsymbol{\mu}_k)_{1\leq k\leq K}$
      among the rows of $\boldsymbol{X}$ without replacement}
    \STATE{$c_i\leftarrow \arg\max_{1\leq k\leq K}\boldsymbol{\mu}_k^T\boldsymbol{x}_i$}
    \STATE{$\tau_{ik}\leftarrow \mathbb{I}_{k=c_i}$}
    \STATE{$\alpha_k=\frac{1}{n}\sum_{i=1}^n\tau_{ik}$}
    \STATE{set $\kappa_k$ to the solution of
      \begin{equation*}
 \frac{I_{d/2}(\kappa_k)}{I_{d/2-1}(\kappa_k)}=\boldsymbol{\mu}_k^T\frac{\sum_{i=1}^n\tau_{ik}
  \boldsymbol{x}_i}{\sum_{i=1}^n\tau_{ik}}.       
      \end{equation*}
    }
  \end{algorithmic}
\end{algorithm}

Secondly, mixture models can fall into problematic local configurations. As
pointed out in \cite{10.5555/1046920.1088718}, $\kappa_k$ can
become unbounded if the corresponding component focuses on a single
observation, in a similar behavior as the one observed for mixture of
Gaussian distributions when the standard deviation of the component
vanishes. As in \cite{10.5555/1046920.1088718}, we prevent this issue by
capping $\kappa_k$ to a large value ($10^6$ in our
experiments).

On the contrary, a component of the mixture can also become
useless when $\kappa_k\rightarrow 0$. This corresponds to the component
converging to a uniform distribution. This behavior is easily detected as it
manifests by having the right hand side of equation \eqref{eq:kappa:l1} taking
a value larger or equal to 1. We monitor this quantity and interrupt the
algorithm when such a situation is encountered. We report in this case a
convergence issue. Notice that the initialisation process described above can
also fail for this reason.

Finally, when $\beta>0$, equation \eqref{eq:mugene} can produce a zero
``directional mean'': this means in practice that the M step has failed. When
we detect this issue, we stop the algorithm and report a convergence issue.

\bibliographystyle{elsarticle-harv} 
\bibliography{cas-refs}

\end{document}

%% file: results/cstr_dense_stats.tex
\begin{tabular}{rcccccccc}
  \toprule
  &\multicolumn{2}{c}{SK-means}
  & \multicolumn{2}{c}{Shared $\kappa$}
  &\multicolumn{2}{c}{Free $\kappa$}
  &\multicolumn{2}{c}{dbmovMFs} \\
K & mean & sd & mean & sd & mean & sd & mean & sd \\ 
  \midrule
2 & 0.471 & $4.52 \, 10^{-3}$ & 0.344 & $2.95 \, 10^{-3}$ & 0.395 & $3.28 \, 10^{-4}$ & 0.442 & $6.28 \, 10^{-2}$ \\ 
  3 & 0.757 & $1.31 \, 10^{-2}$ & 0.756 & $3.83 \, 10^{-3}$ & 0.567 & $1.89 \, 10^{-2}$ & 0.772 & $7.35 \, 10^{-3}$ \\ 
  4 & 0.802 & $1.77 \, 10^{-2}$ & 0.804 & $1.22 \, 10^{-2}$ & 0.519 & $4.48 \, 10^{-2}$ & 0.803 & $1.72 \, 10^{-2}$ \\ 
  5 & 0.659 & $4.05 \, 10^{-2}$ & 0.650 & $2.11 \, 10^{-2}$ & 0.497 & $8.78 \, 10^{-2}$ & 0.716 & $4.06 \, 10^{-2}$ \\ 
  6 & 0.572 & $4.59 \, 10^{-2}$ & 0.569 & $2.24 \, 10^{-2}$ & 0.520 & $9.51 \, 10^{-2}$ & 0.663 & $4.15 \, 10^{-2}$ \\ 
  7 & 0.535 & $5.21 \, 10^{-2}$ & 0.493 & $2.82 \, 10^{-2}$ & 0.463 & $8.28 \, 10^{-2}$ & 0.625 & $5.18 \, 10^{-2}$ \\ 
  8 & 0.481 & $4.99 \, 10^{-2}$ & 0.448 & $3.13 \, 10^{-2}$ & 0.441 & $7.37 \, 10^{-2}$ & 0.588 & $6.16 \, 10^{-2}$ \\ 
   \bottomrule
\end{tabular}

%% file: results/cstr_sparse_stats.tex
\begin{tabular}{lrr}
  \toprule
Criterion/model & mean & sd \\ 
  \midrule
dbmovMFs & 0.803 & $1.723 \times 10^{-2}$ \\ 
  Dense & 0.804 & $1.217 \times 10^{-2}$ \\ 
  AIC & 0.807 & $1.083 \times 10^{-2}$ \\ 
  BIC & 0.808 & $9.483 \times 10^{-3}$ \\ 
  EBIC & 0.803 & $7.687 \times 10^{-3}$ \\ 
  RIC & 0.797 & $7.991 \times 10^{-3}$ \\ 
  RICc & 0.750 & $1.248 \times 10^{-2}$ \\ 
   \bottomrule
\end{tabular}